\documentclass{article}
\usepackage[utf8]{inputenc}
\usepackage{multirow}
\usepackage{xcolor}         %

\usepackage{algorithm}
\usepackage{algorithmic}
\usepackage[algo2e, noend, noline, ruled, linesnumbered]{algorithm2e}

\usepackage[authoryear]{natbib} 
\usepackage{mathtools}
\mathtoolsset{showonlyrefs=true}

\input{mlstyle-article.sty}

\title{Learning in Mean Field Games: A Survey}

\author{Mathieu Lauri{\`e}re$^{1,*}$, Sarah Perrin$^{2,*}$, Julien P\'erolat$^{2}$, Sertan Girgin$^{2}$, Paul Muller$^{2}$, \\
Romuald \'Elie$^{2}$, Matthieu Geist$^{\dagger,3}$, Olivier Pietquin$^{\dagger,3}$\thanks{$^1$ NYU Shanghai, $^2$ Google Deepmind, $^3$ Cohere. $^*$equal contributions, $^\dagger$ equal contributions}}

\renewcommand\footnotemark{} %

\begin{document}

\maketitle

\begin{abstract}

Non-cooperative and cooperative games with a very large number of players have many applications but remain generally intractable when the number of players increases. Introduced by \cite{lasrylions-japanMR2295621} and \cite{huangmalhamecaines2006large}, Mean Field Games (MFGs) rely on a mean-field approximation to allow the number of players to grow to infinity. Traditional methods for solving these games generally rely on solving partial or stochastic differential equations with a full knowledge of the model. Recently, Reinforcement Learning (RL) has appeared promising to solve complex problems at scale. The combination of RL and MFGs is promising to solve games at a very large scale both in terms of population size and environment complexity. 
In this survey, we review the quickly growing recent literature on RL methods to learn equilibria and social optima in MFGs. 
We first identify the most common settings (static, stationary, and evolutive) of MFGs. We then present a general framework for classical iterative methods (based on best-response computation or policy evaluation) to solve MFGs in an exact way. Building on these algorithms and the connection with Markov Decision Processes, we explain how RL can be used to learn MFG solutions in a model-free way. Last, we present numerical illustrations on a benchmark problem, and conclude with some perspectives. %
\newline

\end{abstract}

\tableofcontents

\section{Introduction}

Since their introduction by~\cite{lasrylions-japanMR2295621} and~\cite{huangmalhamecaines2006large}, mean field games (MFGs for short) have gained momentum as a powerful paradigm to study large populations of strategic agents. The main idea, borrowed from statistical physics, is to use the mean field distribution corresponding to the limiting mean field situation with an infinite number of players. All the individual interactions can then be replaced by the interaction between a representative player and the mean field distribution, which considerably simplifies the model and the analysis. This approximation relies on the assumption that the population is homogeneous and that the interactions are symmetric in the sense that each player interacts only with the empirical distribution of the other players. The solution to the MFG provides an $\epsilon$-Nash equilibrium for the corresponding $N$-player game, with $\epsilon$ going to $0$ as $N$ goes to infinity. Furthermore, under suitable assumptions, $N$-player Nash equilibria or social optima converge to the corresponding mean field equilibrium or social optimum. Such results build on the idea of propagation of chaos~\citep{sznitman1991topics} but are more subtle since the players are not simple particles but rational agents making optimal decisions and reacting to other players' decisions~\citep{lacker2017limitmkv,cecchin2019convergence,lacker2020convergenceclosedloopmfg,cardaliaguet2019master}.

\subsection{Mean field games}

\paragraph{General intuition. }
We start this survey by defining at a high level what a Mean Field Game (MFG) is. Intuitively, an MFG is a game with an infinite number of identical anonymous players. All players have a similar behavior, i.e. they are symmetric: we do not need to retain the identity of a player as part of its state. Furthermore, as we have an infinite number of anonymous players, we can replace all the atomic players by their distribution 
over the state (and sometimes action) space. The population's distribution enables to focus only on the interaction between a so-called representative player, which is sampled from the population's distribution, and the population's distribution itself. Our ultimate goal is to compute a Nash equilibrium, which corresponds to the situation where no player has an interest in deviating from its current behavior, provided that the other players do not deviate either. Looking for a Nash equilibrium makes the assumption that the players are all rational, i.e their goal is to maximize their own reward (or minimize their cost).

 Most of the literature focuses on two types of problems: Nash equilibria or social optimum. These two settings are typically referred to respectively as MFG and mean field control (MFC) - or control of McKean-Vlasov dynamics -  \citep{bensoussanfrehseyam2013mean,carmona2018probabilistic}. In both cases, the solutions are typically characterized through optimality conditions taking the form of a coupled system of forward-backward equations. The forward equation describes the evolution of the population distribution while the backward equation represents the evolution of the value function (i.e. the utility of its behaviour) for a representative player. In the continuous time and continuous space setting, the equations can be partial differential equations (PDEs)~\citep{lasrylions-japanMR2295621} or stochastic differential equations (SDEs) of McKean-Vlasov type~\citep{carmona2013meanfbsde} depending on whether one relies on the analytical or the probabilistic approach. We refer to \textit{e.g.} \citet{bensoussanfrehseyam2013mean,carmona2018probabilistic,carmona2018probabilistic2,achdou2020meancetraro} for more details. 
In this survey, we focus on the discrete time case, which is closer to the framework of Markov Decision Processes~\citep{bertsekasshreve1996stochastic,puterman2014markovbook}. For the sake of simplicity, most of the presentation assumes finite state and action spaces, although many ideas can be extended to continuous spaces.

\paragraph{Example. } As a typical example, we can consider crowd motion. Each player is an agent represented by their position and is able to control their velocity so as to reach a target destination while minimizing the effort made to move. Typically, passing through a crowded region, \textit{i.e.} a region with a high density of players, requires extra efforts or reduces the velocity, thus creating some congestion. If we assume that the number of agents is extremely large and that these agents are homogeneous and have symmetric interactions, then we can approximate the empirical distribution of positions by the mean field distribution corresponding to the limiting regime with an infinite population. This allows to simplify tremendously the computation of a Nash equilibrium because we only need to compute the optimal policy of the representative player.

\begin{remark}[On MFGs and non-atomic anonymous games]
\label{rem:MFG-nonatomic-anonymous}
    Games modeling infinite populations of agents have also been studied in the framework of non-atomic anonymous games, which have founds applications particularly in economics, see e.g.~\citet{aumann1964markets,schmeidler1973equilibrium,aumannshapley2015values}. In such games, there is typically a continuum of players, indexed by, say, real numbers in $I=[0,1]$ and the population is represented by a non-atomic measure on $I$. Each player perceives the other players through some aggregate quantity. Although this is very similar to the MFG framework, the key difference is that the MFG approach completely avoids representing the continuum of players. The main idea is to exploit the homogeneity of the population and the symmetry of interactions to simplify the characterization of an equilibrium: it is sufficient to understand the behavior of a single representative player facing a distribution representing the aggregate information available to this player. The analysis is greatly simplified, particularly when it comes to stochastic games. Defining rigorously a continuum of random variables with nice measurability properties is not trivial, as noticed for instance by~\cite{duffie2012exact,sun2006exact} who used the concept of rich Fubini extension to develop an exact law of large numbers. The MFG framework allows to carry out the mathematical analysis of Nash equilibria without requiring such sophistication.
\end{remark}

\paragraph{Some applications.} MFGs have found various applications such as population dynamics~\citep{gueantlasrylions2011mean,achdou2017mean,cardaliaguet2016segregation}, crowd motion modeling~\citep{achdoulasry2019meancrowd, burger2013mean,djehiche2017mean,aurell2019modeling,achdou2016mean,chevalier2015micro}, flocking~\citep{nourian2010synthesisflocking,nourian2011meanflocking,grover2018meanhomogeneousflocking,perrin2021mfgflockrl}, opinion dynamics and consensus formation \citep{stella2013opinion,bauso2016opinion,parise2023graphon}, autonomous vehicles \citep{huang2019game,shiri2019massive,chen2023hybrid}, epidemics control \citep{laguzet2015individualsir,hubert2018nash,elie2020contact,lee2021controllingepidemics,aurell2022optimalincentives,doncel2022meansir}, macro-economic models \citep{achdou2017income,elie2019tale,achdou2017income,achdou2014pde,chan2015bertrand,gomes2014socio, djehiche2016mean}, finance \citep{lachapelle2016efficiency,cardaliaguetlehalle2018mean,lasrylions-japanMR2295621,lachapelle2016efficiency,gomes2020mean,carmona2020applicationsmfgfinanceeconsurvey,bernasconi2023dealer}, energy production and management \citep{alasseur2020extended,couillet2012electrical,elie2019mean, bagagiolo2014mean,kizilkale2019integral,li2016mean, gueantlasrylions2011mean,achdou2016long,chan2017fracking,graber2018existence}, security and communication \citep{meriaux2012mean,samarakoon2015energy,hamidouche2016mean,yang2017mean, kolokoltsov2016mean,kolokoltsov2018corruption}, traffic modeling \citep{bauso2016densitynetwork,salhab2018meanroute,huang2019stabilizing,tanaka2020linearly,cabannes2021solving}, or engineering \citep{djehiche2016mean,zhou2021reinforcementdecentralized,zhou2021decentralizedtracking,zhou2022decentralizedpursuite}.

\paragraph{Numerical methods. }
Most existing numerical methods for MFGs and MFC problems are based on the aforementioned optimality conditions phrased in terms of PDEs or SDEs. 
 Such approaches typically rely suitable discretizations, \textit{e.g.} by finite differences~\citep{MR2679575,achdou2012mean,briceno2018proximal,BricenoAriasetalCEMRACS2017}, semi-Lagrangian schemes~\citep{carlini2014fully,MR3392626}, or probabilistic approaches~\citep{chassagneux2019numerical,angiuli2019cemracs}. To solve the discrete systems, several approaches have been proposed, such as Newton iterations~\cite{achdou2012mean,camilli2023convergence}, Picard iterations~\cite{carlini2014fully} or methods based on the reformulation of the problem as an optimization problem~\cite{benamou2015augmented,achdou2016mean,briceno2018proximal,guo2024mf}. We refer to  \textit{e.g.}~\citet{achdoulauriere2020mfgnumerical,lauriere2021numericalams} for recent surveys on these methods. A major advantage of these methods is that they are well understood and very successful in small dimension. However, they are in general not suitable to tackle MFGs with high dimensional states or controls due to the curse of dimensionality~\citep{bellman1957markovian}. To address this limitation, stochastic methods based on approximations by neural networks have recently been introduced by~\citet{carmonalauriere2021convergenceergodic,carmona2022convergence,fouque2019deep,germain2019numerical} using optimality conditions for general mean field games, by~\citet{ruthotto2020machine} for MFGs which can be written as a control problem, and by~\citet{cao2024connecting,lin2020apac} for variational MFGs in connection with generative adversarial networks (GANs)~\citep{goodfellow2014generative}. We refer to~\citet{hulauriere2022recentdevml} for a recent survey on machine learning methods for control and games, with applications to MFGs and MFC problems. However, these methods focus on computing equilibria or social optima when the model is known, and strongly rely 
 on exact computations of gradients by exploiting the full knowledge of the model. Some methods are combining PDE techniques with approximate dynamic programming, see e.g.~\citep{yin2013learning}. The learning methods we focus on in this survey aim instead at solving MFGs and MFC in a model-free RL fashion to foster the scalability of numerical methods for these problems.

\subsection{Learning}

\paragraph{Two notions of learning. } As this survey lies at the intersection of several scientific fields and communities, the second question that we need to address before diving more deeply into the topic of interest is 
what learning means in our context. There are mainly two interpretations of learning. The first one comes from game theory and economics and, according to \cite{fudenberglevine1998theory}, refers to \emph{“The theory of learning in games [\dots] examines how, which, and what kind of equilibrium might arise as a consequence of along-run non equilibrium process of learning, adaptation, and/or imitation.”} From this point of view, the main focus is on how the players iteratively adjust their behavior until convergence to an equilibrium. 
The second interpretation of learning is mainly used in Machine Learning and in Reinforcement Learning (RL). As \cite{mitchell1997machine} puts it, \emph{"a computer program is said to learn from experience E with respect to some class of tasks T and performance measure P if its performance at tasks in T, as measured by P, improves with experience E.”} In this concept, the concept of learning is very related to the idea of improving one's performance by using data or samples. 
In this survey, we are interested in delineating these two notions of learning while explaining how they can be combined.

\paragraph{Learning in games. } 

The literature on learning in games goes back at least to the works of~\citet{Shan50} and \citet{samuel1959some}. Since then, a lot of progress has been made but remains mostly limited to games with a small number of players. Many of the recent breakthrough results have been obtained using a combination of RL~\citep{Suttonbarto2018} and deep neural networks~\citep{goodfellow2016deeplearning}, see 
\textit{e.g.} Go~\citep{silver2016mastering,silver2017mastering, Silver18AlphaZero}, Chess~\citep{campbell2002deep}, Checkers~\citep{schaeffer2007checkers, samuel1959some}, Hex~\citep{Anthony17ExIT}, Starcraft II~\citep{vinyals2019grandmaster}, poker games~\citep{Brown17Libratus, Brown19Pluribus, moravvcik2017deepstack, bowling2015heads}, Stratego \citep{mcaleer2020pipeline, perolat2022mastering} or Diplomacy \citep{meta2022human}.

\paragraph{Learning in mean field games. } The goal of this survey is to review the quickly growing literature at the intersection of learning and MFGs. Combining mean field approximations, which allow to tackle large population games, and RL, which allows to handle highly complex environments, is a promising avenue 
to solve multi-agent systems at a very large scale, both in terms of population size and model complexity.

\subsection{Outline of the survey}

In the rest of this section, we introduce a few useful notations. In Section~\ref{sec:settings}, we present several settings of MFG and MFC problems that have appeared in the literature. We stress their similarities and differences, both in terms of definitions and solutions. In Section~\ref{sec:iterative-methods}, we present algorithms to compute MFG and MFC solutions. We start by recalling some classical methods to solve MDPs and then describe mainly two classes of algorithms to compute
equilibria in MFGs. These algorithms are based on iteratively updating the mean field and the policy, so we refer to them as iterative methods. Building on these methods and the connection between MDPs and RL, we explain in  Section~\ref{sec:rl-algorithms} how RL and deep RL methods can be adapted to solve MFGs and MFC problems. Section~\ref{sec:numerical-experiments} discusses metrics that can be used to assess the numerical convergence of algorithms and illustrate some of the methods on a representative MFG example. Finally, we conclude in Section~\ref{sec:conclusion} with a short summary and some perspectives.

\section{Definition of the problems}
\label{sec:settings}

\begin{figure}[tbh]
    \centering
    \begin{minipage}{\linewidth}
    \centering
    \includegraphics[width=.6\linewidth]{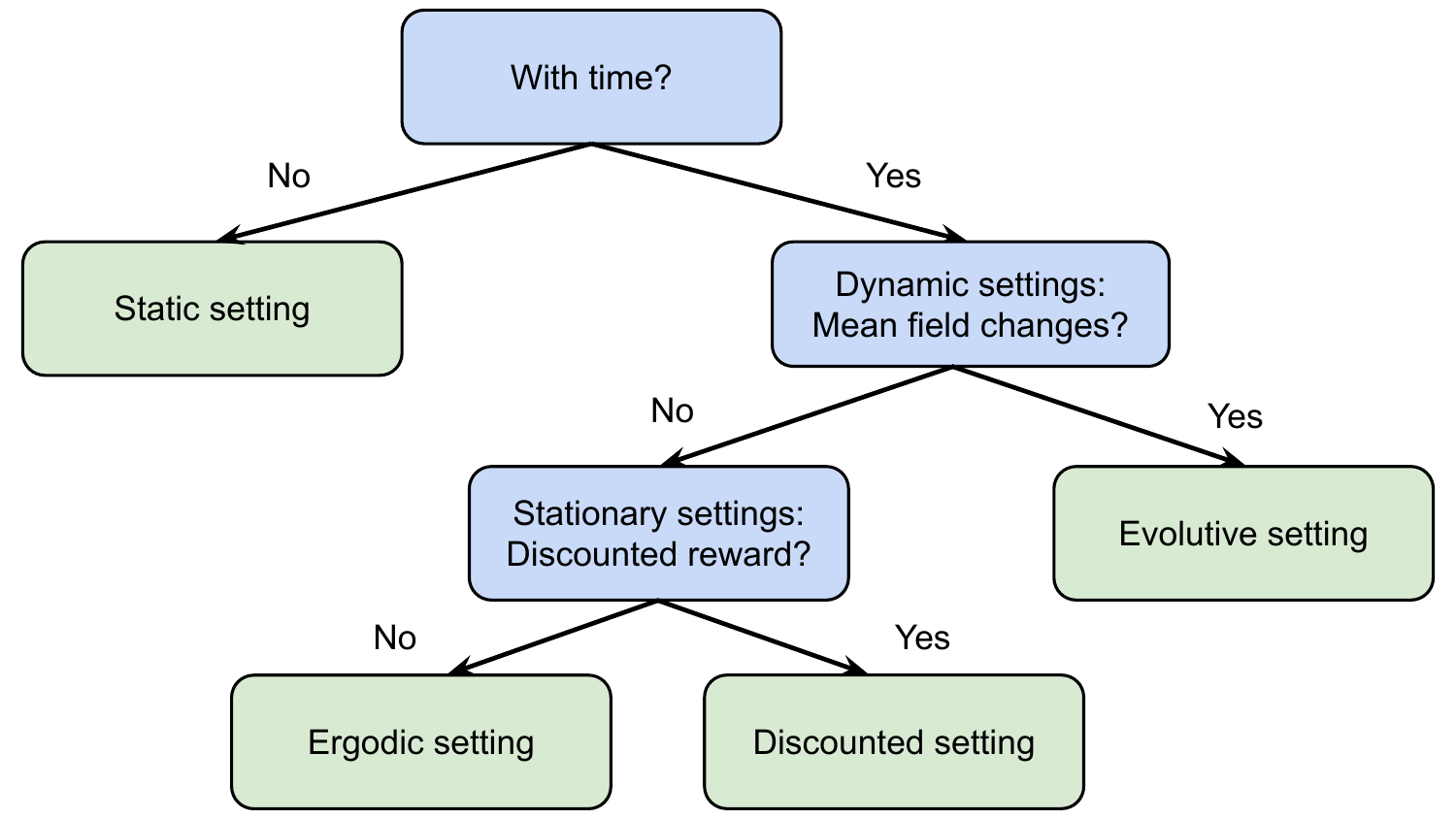}
    \end{minipage}
    \caption{Organization of the MFG settings, depending on whether there is a time component and, if yes, whether the mean field is stationary or not.}
    \label{fig:diagram-settings}
\end{figure}

In this section, we present a short introduction to several settings of MFGs that have been studied in the literature. Here we focus on the definition of the settings and we provide some intuition behind these models. For theoretical results related to existence of Nash equilibria and their properties, we provide references to the literature. 
Figure~\ref{fig:diagram-settings} summarizes the organization we follow for the presentation. The first distinction is whether time is involved or not. The second distinction is whether the mean field distribution through which interactions occur is evolving or stationary. 
These settings correspond to different applications and different notions of Nash equilibrium. We present five settings, that can be summarized as follows. We start with games in which the agents take a single decision. There is no notion of time intrinsic to the game so we call them \defi{static MFGs}. We then turn to games in which there is a dynamical component.
In such games, each agent has a state that evolves along time, and they can act on this evolution. At the level of the population, in some situations, we can expect the distribution of states to be in a stationary regime, in which the population is stable at a macroscopic level, even though each agent's state is possibly changing. We refer to this setting as a \defi{stationary MFG}. Another setting which shares the property of modeling interactions through a stationary mean field distribution is the \defi{ergodic MFG} setting, which considers a time averaged reward. In some other cases, one wants to understand not only the stationary regime, but how the population evolves, starting from an initial configuration. This setting is relevant for applications where the agents' behaviors change along time, for instance because there is a finite horizon at which the game stops. We call such games \defi{evolutive MFGs}. This setting comes at the expense of having policies and mean-field terms that depend on time and are thus harder to compute. To mitigate this complexity while not falling completely into the stationary regime, an intermediate model has been introduced. The idea is to try to keep the best of the stationary and evolutive settings by considering a proxy for the whole evolution of the distribution. We call this setting \defi{discounted distribution MFGs}. In the rest of this section, we define each setting as well as the corresponding notion of Nash equilibrium, along with relevant concepts. 

\paragraph{Useful notations. } 
We introduce notations used throughout the text. 
    $\states$ and $\actions$ denote a finite state set and a finite action set respectively.
    If $E$ is a finite set, $|E|$ denotes the cardinality of $E$.
    $2^E$ denotes the set of subsets of a set $E$.
    $\Delta_E$ is the set of probability distributions on a set $E$; when $E$ is finite (which is generally the case for us), it is also identified with the corresponding simplex in the Euclidean space of dimension $|E|$, and we view probability distributions as (normalized) vectors in $\RR^{|E|}$.
    $\argmax$ is understood as the set of all maximizers.
    $\PP$ and $\EE$ denote respectively probability and expectation.
    Given a probability distribution $p$ on a set $\cX$ and a function $\varphi: \cX \to \RR$, we denote $\EE_{x \sim p}[\varphi(x)] = \EE[\varphi(x) \,|\, x \sim p] = \sum_{x \in \cX} p(x) \varphi(x)$.

\subsection{Static MFG}
\label{sec:static-mfne}

\looseness=-1
\paragraph{Notations. } Let $\actions$ be a finite action space. The behavior of one player, called a \defi{strategy}, is an element of $\Delta_\actions$, that is a distribution over the action set. The behavior of the population is also an element of $\Delta_\actions$. We denote a generic element of $\actions$ by $a$, and we denote a generic strategy (individual behavior) by $\pi$.

\paragraph{Finite population game. } Consider first an $N$-player game of the following form. Every agent chooses an action and gets a reward which depends on her own action and the empirical distribution of actions taken by all the agents. To be specific, denote by $a_i$ the action taken by agent $i \in \{1,\dots,N\}$. This agent gets the reward: $r(a^i, \xi^N)$, where $\xi^N = \frac{1}{N}\sum_{j=1}^N \delta_{a^j}$ is the empirical distribution of actions. If players use random actions following policies, say strategy $\pi^i \in \Delta_\actions$ for player $i$, then we can consider the average reward for player $i$, namely:
\begin{equation}
    \label{eq:total-reward-static}
    J^N_{\mathrm{static}}(\pi^i; \pi^{-i})
    = \EE_{a^j \sim \pi^j,j=1,\dots,N}\left[ r(a^i, \xi^N)\right],
\end{equation}
where $\pi^{-i} = (\pi^1,\dots,\pi^{i-1}, \pi^{i+1}, \dots, \pi^N)$. We assume that all each player chooses their strategy before seeing the other player's strategies or actions. We can look for a Nash equilibrium, which in this context is defined as follows,
\begin{definition}[Static finite-population Nash equilibrium]
\label{def:static-NE}
For $\epsilon \ge 0$, an \defi{$\epsilon-$Nash equilibrium} is a tuple of strategies $(\hat\pi^1,\dots,\hat\pi^N)$ such that for every $i=1,\dots,N$, for every $\pi^i \in \Delta_\actions$, 
$$
    J^N_{\mathrm{static}}(\hat\pi^i; \hat\pi^{-i})
    \ge 
    J^N_{\mathrm{static}}(\pi^i; \hat\pi^{-i}) - \epsilon.
$$
A \defi{Nash equilibrium} is an $\epsilon-$Nash equilibrium with $\epsilon=0$.
\end{definition}

Note that the game is \defi{symmetric} or \defi{anonymous} in the sense that player $i$'s reward depends on player $j$'s action only through the empirical distribution of actions $\xi^N$, so from the point of view of player $i$, all the other players are exchangeable. Furthermore, the population is \defi{homogeneous} in the sense that all the players have the same reward function $J^N_{\mathrm{static}}$ (as a function of their strategy and the rest of the population's strategies). Based on this, we can (at least formally) pass to the limit and consider a game with an infinite population.

\paragraph{Mean field game. } We now turn our attention to the MFG. We denote a generic population behavior by $\xi \in \Delta_\actions$. The policy of a representative player is denoted by $\pi \in \Delta_\actions$. Although, in this static setting, they are both elements of $\Delta_\actions$, we use different notations to stress that the interpretation is different. The model is completely defined by the reward function $r: \actions \times \Delta_\actions \to \RR$, where the first input is the player's action and the second input is the population's action distribution. To summarize, we have the following definition.
\begin{definition}[Static MFG setting]
    A \defi{static MFG setting} is defined by the tuple $(\actions, r)$ where $\actions$ is the action space and $r: \actions \times \Delta_\actions \to \RR$ is the reward function. 
\end{definition}

Given a population behavior $\xi \in \Delta_\actions$, the total reward of a player using $\pi \in \Delta_\actions$ is defined as the expected reward when sampling an action according to $\pi$:
    $$
        J_{\mathrm{static}}(\pi; \xi) = \EE_{a \sim \pi}\left[ r(a, \xi)\right].
    $$
The reward function $r$ can for instance encode crowd aversion or attraction towards a population action of interest. Note that in contrast with the $N$-player game's reward~\eqref{eq:total-reward-static}, here the population is represented by $\xi$. In fact, we implicitly assume that all the other players use the same strategy $\xi$. 

\begin{example}
One of the first examples in the MFG literature is the problem of choosing a starting time for a meeting, introduced and solved explicitly by~\citet{gueantlasrylions2011mean}. In this problem, the players choose at what time they want to arrive to the meeting room so that they are neither too late nor too early. The global outcome is the time at which the meeting actually starts, which is not known in advance and depends on the everyone's arrival time. Despite its name, there is no dynamic aspect in the original formulation of the example. Another popular example is the problem in which each agent chooses a location on a beach. They want to put their towel close to an ice cream stall but not in a very crowded area. The global outcome is the distribution of towels on the beach. To be specific, a simple model can be as follows: $\actions = [0,1]$, which represents possible positions on the beach, $a_{\mathrm{stall}}\in\actions$ is the position of the stall, and the reward is $r(a,\xi) = -|a - a_{\mathrm{stall}}| - \ln(\xi(a))$, where the first term penalizes the distance to the stall and the second term penalizes choosing a location $a$ at which the density $\xi(a)$ of people is high. 
The towel on the beach problem has inspired many examples in the MFG literature, see e.g.~\citet{perrin2020fictitious} for a dynamic version of a similar game and~\citet{carmona2022stochasticgraphonstatic} for a static game with non-homogeneous population.
\end{example}

\begin{example}
    A classical example of static game in game theory is the {\bf prisoner's dilemma}. One can define a mean field version of this game as follows~\citep{muller2022learningmfce}:     Consider the action set $\actions=\{C,D\},$  where the choices are cooperate or defect, and reward function
    $
        r(C, \mu) = 3 \mu(C) - \mu(D)\;,$ $ r(D, \mu) = 4 \mu(C) - 0 \mu(D). 
    $
    A representative player gets a reward which depends on the proportion of cooperating and defecting players in the population. Here, playing $D$ is a dominating strategy. 
    Another classical example in game theory is the rock-paper-scissors game, for which an MFG generalization can be defined as follows~\citep{muller2022learningmfce}: the action set  is $\actions\{R,P,S\}$ and the reward is:
    $
        r(R, \mu) = \mu(P) - \mu(S)\;,$ $
        r(P, \mu) = \mu(S) - \mu(R)\;,$ $
        r(S, \mu) = \mu(R) - \mu(P)\;.
    $
    We refer to \citep{muller2022learningmfce} for more details on these two MFGs, as well as numerical illustration of learning algorithms. 
\end{example}

A central concept is the notion of best response. Let us define the (set-valued) \defi{best response map} by:
$$
    \BR_{\mathrm{static}}: \Delta_\actions \to 2^{\Delta_\actions}, \xi \mapsto \BR_{\mathrm{static}}(\xi) := \argmax_{\pi \in \Delta_\actions} J_{\mathrm{\mathrm{static}}}(\pi; \xi).
$$
\begin{definition}[Static mean field Nash equilibrium]\label{def:static-MFNE}
$\hat\pi \in \Delta_\actions$ is a \defi{static mean field Nash equilibrium} (static MFNE) if $\hat\pi \in \BR_{\mathrm{static}}(\hat\pi).$
\end{definition}
The above definition has the advantage to clearly show that the equilibrium is a fixed point of $ \BR_{\mathrm{static}}$.

Another point of view, which seems a bit circumvoluted in the static setting but will be useful in the dynamic settings presented in the sequel, consists in saying that the equilibrium is given by a pair of a representative agent's behavior and the population's behavior. Here, it means that the equilibrium is a pair $(\hat\pi, \hat\xi) \in \Delta_\actions \times \Delta_\actions$ such that:
\begin{enumerate}
    \item Optimality: $\hat\pi$ is optimal for the representative agent facing $\hat\xi$, i.e., $\hat\pi \in \BR_{\mathrm{static}}(\hat\xi)$,
    \item Consistency: $\hat\xi$ corresponds to the population behavior when all every agent uses $\hat\pi$, i.e., $\hat\xi = \hat\pi$.
\end{enumerate}
The second point represents the fact that all the agents are ``rational in the same way'' and hence, at equilibrium, adopt the same behavior.  This viewpoint is unnecessarily complicated in this setting as $\hat\pi$ alone is enough to define the MFNE, but will be useful in dynamic settings.

Yet another way to rewrite the above definition of MFNE is through the notion of \defi{exploitability}, see Section~\ref{sec:metrics} below. A policy is a Nash equilibrium if and only if its exploitability is zero.

\begin{remark} 
    \looseness=-1
    Consistently with the literature on normal-form games~\citep{fudenbergtirole1991gametheory}, each player chooses a distribution over actions \emph{without} seeing the strategy chosen by other players and the resulting distribution at the population level. Each agent thus tries to anticipate, in a rational way, the distribution generated by other players' actions. %
\end{remark}

A Nash equilibrium corresponds to a situation in which no selfish player has any incentive to deviate unilaterally. However, it is not necessarily a situation that is optimal from the point of view of the whole population. The notion of social optimum is discussed below in Section~\ref{sec:setting-soc-opt}.

\begin{remark}
    Although we provided an intuitive explanation for $\xi$ in terms of a continuum of players, we want to stress that in the definition of an MFG equilibrium, we actually do not need to introduce mathematically a continuum of indices for the players. As already pointed out in Remark~\ref{rem:MFG-nonatomic-anonymous}, this shortcut is one of the main advantages of the MFG paradigm compared with non-atomic anonymous games.%
\end{remark}

\paragraph{Connection between finite population and mean field games. } The connection between a finite population game and the corresponding mean field games can be viewed in two different ways. First, one can consider a sequence of $N$-player games with increasing $N$, and try to show that the corresponding sequence of equilibrium policies converges to the MFG equilibrium policy. Second, one can see whether the MFG equilibrium policy is an equilibrium policy in an $N$-player game; in general this is not the case, but it provides an approximate equilibrium whose approximation quality improves when $N$ is larger. In general, the second direction is easier to establish, and it does not even require the existence of a Nash equilibrium for the $N$-player game. In the static setting, it would take the following form: under suitable assumptions, if $\hat\pi \in \Delta_\actions$ is a static MFNE (see Definition~\ref{def:static-MFNE}), then it is an $\epsilon(N)-$Nash equilibrium for the corresponding $N$-player game (see Definition~\ref{def:static-NE}), and $\epsilon(N) \to 0$ as $N \to +\infty$. This is a fundamental motivation to study and solve MFGs.
In the context of continuous space models, we refer to e.g.~\cite{lackerramanan2019rare} and~\cite[Section 4]{lacker2018mean-lecturenotes} for more details.

\subsection{Background on MDPs}
\label{sec:background-mdp}

We recall a few important concepts pertaining to optimal control in discrete time for a single agent. We will only review the main ideas and we refer to \textit{e.g.} \citet{bertsekasshreve1996stochastic,puterman2014markovbook} for more details. The notion of Markov decision processes will play a key role in the description of dynamic MFGs.

\subsubsection{Notations for the dynamic setting}

In contrast with the static case, in the dynamic setting, each agent has a state which evolves in time. The agent can influence the evolution of their own state using actions. The population's state is the distribution of the agents' states, the joint distribution of their states and actions. This is what constitutes the mean field, with which every agent interacts through its dynamics and its rewards.  

As far as the population distribution is concerned, we will consider mainly two types of settings: one in which the population distribution is fixed, and one in which it can also evolve. Typically, the former is conceptually simpler and easier to compute but the latter is more realistic since many real world applications involve a population evolving in time. In each cases, several variants can be considered. For the sake of brevity, we shall focus only on the main ideas.

We consider discrete time models, with time denoted by $n \in \NN = \{0,1,2,\dots\}$. We will denote by $N_T$ the \defi{time horizon}, which can be finite or infinite, we will use the notation $N_T$. We denote by $\actions$ a finite action space as before. We let $\states$ be a finite \defi{state space}. A \defi{stationary policy} is an element of $\sPol := (\Delta_\actions)^\states$, i.e., a function from states to probability distributions over actions.

Depending on the setting, we might consider policies that are stationary or that depend on time.

\begin{remark}
    To alleviate the presentation, we choose to focus on finite state and action spaces and discrete time. As mentioned earlier, in some applications, models with continuous time and space might be more relevant. Such models are typically analyzed using partial differential equations or stochastic differential equations. Suitable discretizations can lead to (possibly non-trivial) approximations of these continuous models with discrete models. For instance, ~\citet{hadikhanloosilva2019finite,bertucci2022mean} analyzed the convergence of a finite MFG to a continuous one. We do not discuss in detail the continuous settings here and we refer the interested reader to the literature, e.g.,~\citet{huangmalhamecaines2006large,lasrylions-japanMR2295621,bensoussanfrehseyam2013mean,carmona2018probabilistic,carmona2018probabilistic2}.
\end{remark}

\subsubsection{Stationary MDP}
\label{sec:background-statio-mdp}
A  \defi{stationary Markov decision process} (MDP) is a tuple $(\states, \actions, p, r, \gamma)$ where $\states$ is a state space, $\actions$ is an action space, $p: \states \times \actions \to \Delta_\states$ is a transition kernel, $r:\states \times \actions \to \RR $ is a reward function and $\gamma\in(0,1)$ is a discount factor. Using action $a$ when the current state is $x$ leads to a new state distributed according to $p(\cdot|x,a) \in \Delta_\states$ 
and produces a reward  $r(x,a)$. The reward could be stochastic but to simplify the presentation, we consider that $r$ is a deterministic function of the state and the action. A \defi{stationary policy} $\pi: \states \to \Delta_\actions$, $x \mapsto \pi(\cdot | x)$ provides a distribution over actions for each state. We denote by $\Pi$ the set of such policies. 
The goal of the MDP is to find a policy $\pi^*$ which maximizes the total return defined as the expected (discounted) sum of future rewards:
\begin{align*}
    J(\pi) = \EE\Big[\sum_{n \ge 0} \gamma^n  r(x_n,a_n) \Big],
\end{align*}
subject to:
\begin{equation*}
    \left\{\begin{split}
    &x_0 \sim m_0, 
    \\
    &a_n \sim \pi(\cdot|x_n), \quad x_{n+1} \sim p(\cdot|x_n, a_n), \quad n \ge 0,
    \end{split}
    \right.
\end{equation*}
where $m_0$ is an initial distribution whose choice does not influence the set of optimal policies.

Assuming the model is fully known to the agent, the problem can be solved using for instance dynamic programming. The \defi{state-action value function under policy $\pi$} is defined as:
\begin{equation}
    \label{eq:statio-bellman-Qfct}
    Q^{\pi}(x,a) = \EE\left[ \sum_{n \ge 0} \gamma^n r(x_n,a_n) \Big| x_0 = x, a_0 = a, a_n \sim \pi(\cdot|x_n), x_{n+1} \sim p(\cdot|x_n,a_n) \right].
\end{equation}
By dynamic programming, it satisfies the following fixed point equation with unknown $Q:\states\times\actions\to\RR$:
\begin{equation*}
    Q = B^{\pi} Q,
\end{equation*}
where $B^{\pi}$ denotes the \defi{Bellman operator} associated to $\pi$:
\begin{equation}
    \label{eq:mdp-statio-bellman-op}
    (B^{\pi} Q)(x,a) = r(x,a) + \gamma \mathbb{E}_{x' \sim p(\cdot|x,a), a' \sim \pi(\cdot|x')}[Q(x',a')].
\end{equation}
We recall that the expectation is to be understood as:
\begin{equation}
    \label{eq:mdp-statio-expectation}
    \mathbb{E}_{x' \sim p(\cdot|x,a), a' \sim \pi(\cdot|x')}[Q(x',a')] = \sum_{x'} p(x'|x,a) \sum_{a'} \pi(a'|x') Q(x',a').
\end{equation}
The \defi{optimal state-action value function} is defined as:
\begin{equation}
    \label{eq:statio-bellman-Qfct-opt}
    Q^{*}(x,a) = \sup_{\pi} Q^{\pi}(x,a).
\end{equation}
It satisfies the fixed point equation:
\begin{equation}
    \label{eq:statio-mdp-opt-bellman-fixed-point}
    Q = B^{*} Q,
\end{equation}
where $B^{*}$ denotes the \defi{optimal Bellman operator}:
\begin{equation}
    \label{eq:mdp-statio-opt-bellman-op}
    (B^{*} Q)(x,a) = r(x,a) + \gamma \mathbb{E}_{x' \sim p(\cdot|x,a)}[ \max_{a'} Q(x',a')],
\end{equation}
with
\begin{equation}
    \label{eq:mdp-statio-opt-expectation}
    \mathbb{E}_{x' \sim p(\cdot|x,a)}[ \max_{a'} Q(x',a')] = \sum_{x'} p(x'|x,a)  \max_{a'}  Q(x',a').
\end{equation}
The (state only) \defi{value function} associated to a policy $V^\pi: x \mapsto \EE_{a \sim \pi(\cdot|x)}[Q^\pi(x,a)]$ and the (state only) \defi{optimal value function} $V^*: x \mapsto \EE_{a \sim \pi^*(\cdot|x)}[Q^*(x,a)]$, where $\pi^*$ is an optimal policy. These value functions can also be characterized as fixed points of two Bellman operators. 
Note that these objects are all independent of time, as we search for a stationary solution.

\subsubsection{Finite Horizon MDP}
\label{sec:background-finitehorizon-mdp}
One can also consider problems set with a finite time horizon. A  \defi{finite-horizon Markov decision process} (MDP) is a tuple $(\states, \actions, p, r, N_T)$ 
where $\states$ is a state space, $\actions$ is an action space, $N_T$ is a time horizon, $p: \{0,\dots,N_T-1\} \times \states \times \actions \to \cP(\states)$ is a transition kernel, and $r: \{0,\dots,N_T\} \times \states \times \actions \to \RR $ is a reward function. At time $n$, using action $a$ when the current state is $x$ leads to a new state distributed according to $p_n(\cdot|x,a) \in \Delta_\states$ and produces a reward  $r_n(x,a) \in \RR$. A {\bf (non-stationary) policy} $\bspi: \{0,\dots,N_T-1\} \times \states \to \cP(\actions)$, $(n,x) \mapsto \pi_n(\cdot | x)$ provides a distribution over actions for each state at time $n$. So the set of such policies is $\Pi^{N_T}$. 
The goal of the MDP is to find a policy $\bspi^*$ which maximizes the total return defined as the expected (discounted) sum of future rewards:
\begin{equation*}
    J(\bspi) = \EE\Big[\sum_{n = 0}^{N_T}   r_n(x_n,a_n) \Big],
\end{equation*}
subject to:
\begin{equation*}
    \left\{\begin{split}
    &x_0 \sim m_0, 
    \\
    &a_n \sim \pi_n(\cdot|x_n), \quad x_{n+1} \sim p_n(\cdot|x_n, a_n), \quad n = 0, \dots, N_T,
    \end{split}
    \right.
\end{equation*}
where $m_0$ is an initial distribution whose choice does not influence the set of optimal policies.

Here again, assuming the model is known to the agent, the problem can be solved using for instance dynamic programming. The \defi{state-action value function under policy $\bspi$} is defined as: 
$$
\left\{
\begin{split}
    &\bfQ^{\bspi}_{N_T}(x,a) = r_{N_T}(x,a)
    \\
    &\bfQ^{\bspi}_n(x,a) = \EE\left[ \sum_{n' \ge n} r_{n'}(x_{n'},a_{n'}) \Big| x_n = x, a_n = a, a_{n'} \sim \pi_{n'}(\cdot|x_{n'}), x_{n'+1} \sim p_{n'}(\cdot|x_{n'},a_{n'}) \right], 
    \\
    & \qquad\qquad\qquad n = N_T-1,\dots, 0.
\end{split}
\right.
$$
The \defi{optimal state-action value function} is defined as:
\begin{equation}
    \label{eq:mdp-optQ}
    \bfQ^{*}(x,a) = \sup_{\bspi} \bfQ^{\bspi}(x,a).
\end{equation}
Here again, we can introduce the \defi{value function} associated to a policy: $\bfV^{\bspi}_{n}(x) = \EE_{a \sim \pi_n(\cdot|x)}[\bfQ^{\bspi}_{n}(x,a)]$, and the \defi{optimal value function}: $\bfV^{*}_{n}(x) = \EE_{a \sim \pi^*_n(\cdot|x)}[\bfQ^{*}_{n}(x,a)]$, $\bspi^*$ is an optimal policy.

Formally, the finite-horizon MDP can be restated as a stationary MDP by incorporating the time $n$ in the state. However, it can be simpler to directly tackle this MDP using techniques that are specific to the finite-horizon setting. In particular we stress that, in contrast with the stationary setting presented above, the value functions are here characterized by equations which are not fixed point equations but backward equations. They can be solved by backward induction, as we will discuss in the sequel (see Section~\ref{sec:iterative-methods}). For more details on finite-horizon MDP we refer to \textit{e.g.}~\citep{puterman2014markovbook}.

Next, we present settings which involve states and time steps. 

\subsection{Dynamic MFG setting with evolving mean field}

\subsubsection{Motivation from finite population games}
\label{sec:dynamic-finite-pop}
When several agents are interacting, the above MDP settings need to be modified. Typically, each agent wants to solve an MDP, but the states or actions of other agents impact her transitions or rewards so the notion of optimality of a policy is relative to other agents' behavior. It becomes necessary to anticipate the behavior of the rest of the population. We will, as in the static setting, rely on the notion of Nash equilibrium. 

To fix the ideas, consider a system with $N$ agents. 
Let $x^i_n$ and $a^i_n$ denote respectively the state and the action of player $i$ at time $n$. Let $\mu^N_n = \frac{1}{N} \sum_{j-1}^N \delta_{x^j_n}$ denote the empirical distribution of states at time $n$. 

Assume that at time $n$, player $i$'s state evolves according to:
$
    x^i_{n+1} \sim p_n(\cdot|x^i_n, a^i_n, \mu^N_n), 
$
and player $i$ gets the one-step reward:
$
    r_n\left(x^i_n, a^i_n, \mu^N_n\right),
$
where $p: \NN \times \states \times \actions \times \Delta_\states \to \Delta_\states$ and $r: \NN \times \states \times \actions \times \Delta_\states \to \RR$ are a transition kernel and a one-step reward function.

Player $i$ takes her action according a policy. The choice of the class of policies is determined by the information structure, i.e., it depends on what the players can base their decisions on. For instance, we can assume that each player can see the states of all the players at the current time. In this case, a policy for player $i$ is a function $\pi^i: \NN \times (\states)^N \to \Delta_\actions$ which, for each time step and each vector of states gives a distribution over actions, from which player $i$ can sample an action. So at time $n$, we have:
$
    a^i_n \sim \pi^i_n(\cdot| x^1_n,\dots,x^N_n). 
$
Alternatively, since the transitions and the rewards of player $i$ depends on other players only through the empirical distribution, we can expect that the equilibrium policies will have a similar dependence. 

It is in general difficult to characterize and learn policies of this type, due to the number of players. A much more tractable class of policies is so-called \defi{decentralized} policies, depending only on the player's own state (i.e., $\pi^i_n(\cdot|x^i_n)$). But this class is more restrictive:  in general, it yields different Nash equilibria.

However, it is possible to find $\epsilon(N)$-Nash equilibria that are decentralized policies by using MFGs, with $\epsilon(N) \to 0$ as $N \to +\infty$. This relies on the fact that, here the model is \defi{symmetric} (the interactions occur only through $\mu^N$) and \defi{homogeneous} (all the players share the same $p$ and $r$ functions). %

\subsubsection{Evolutive setting}
\label{sec:evol-mfg-setting}

\paragraph{Setting definition. }  We next turn our attention to a model in which not only the agents' state can evolve, but the population's distribution too. In this case, the mean field is not stationary. At each time step, the transition and the reward of every agent depends on the \emph{current} distribution instead of the stationary one. The terminology we use is inspired by the literature on MFG PDE systems, see e.g.~\citep{bardi2019non,cardaliaguet2019long}. The model is defined as follows.  
\begin{definition}[Evolutive MFG setting] 
An \defi{evolutive MFG setting} is defined by a tuple $(\states, \actions, m_0, N_T, p, r)$ consisting of:
\begin{itemize}
    \item a state space $\states$ and an action space $\actions$,
    \item an initial distribution  $m_0 \in \Delta_\states$,
    \item a time horizon  $N_T \in \NN \cup \{+\infty\}$,
    \item a sequence of one-step transition probability kernels
$p_n: \states \times \actions \times \Delta_\states \to \Delta_\states$, $0 \le n \le N_T$,
    \item a sequence of one-step reward functions $r_n: \states \times \actions \times \Delta_\states \to \RR$, $0 \le n \le N_T$.
\end{itemize}
\end{definition}

\begin{remark}[Finite and infinite horizon discounted settings]
    Our notation covers two very common settings:
    The finite horizon, in which  $N_T < +\infty$, and the infinite horizon setting, in which $N_T = +\infty$. In the latter case, it is common to assume that $p$ is independent of time, and that $r$ is of the form $r_n(x, a, \mu) = \gamma^n \tilde r(x, a, \mu)$ where $\tilde r$ is independent of time and bounded. 
\end{remark}

\paragraph{Motivation in finite population games. } 

We provide some intuition for this setting by building on the framework presented in Section~\ref{sec:dynamic-finite-pop}. But in contrast with the stationary setting setting, here  
the total reward of player $i$ is defined as:
$$
    J^N_{\mathrm{evol}}(\bspi^i; \bspi^{-i}) = \EE\left[ \sum_{n=0}^{N_T} r_n(x^i_n, a^i_n, \mu^N_n)\right].
$$
We can (at least formally) pass to the limit as $N \to +\infty$ to derive an MFG corresponding to this setting.

\paragraph{Mean field game. } We now turn our attention to the mean field game. 
In this context, a population behavior is a \defi{mean field sequence}, generally denoted by $\bsmu$, which is an element of $(\Delta_\states)^{N_T+1}$. A \defi{policy} $\bspi$ is an element of $\strategies^{N_T}$. We use bold letter to stress that these are sequences, which can also be viewed as functions of time.

The total reward is:
\begin{equation}
\label{eq:evol-J}
    J_{\mathrm{evol}}(\bspi; \bsmu) = \EE\left[ \sum_{n=0}^{N_T} r_n(x_n, a_n, \mu_n)\right],
\end{equation}
subject to the following evolution of the agent's state:
\begin{equation}
    \label{eq:evol-evol-player}
    \left\{
    \begin{split}
        &x_0 \sim m_0, 
        \\
        &x_{n+1} \sim p_n(\cdot|x_n, a_n, \mu_n), \quad a_n \sim \pi_n(\cdot|x_n), \quad n \ge 0.
    \end{split}
    \right.
\end{equation}
This evolution is analogous to~\eqref{eq:statio-evol-player} except that what used to be the stationary mean-field state is replaced by the current mean-field state at time $n$.

\begin{remark}[Non-stationarity]
\label{rem:evol-finite-infinite}
    Note that even in the infinite horizon setting and even if $p$ and $r$ are stationary (constant with respect to the time parameter $n$), in general the optimal policy still depends on time. This is in contrast with the stationary setting (Section~\ref{sec:statio-mfg-setting}) and is due to the fact that the population distribution starts from $m_0$ and potentially changes. The player needs to take that into account in her decisions. To be specific, even if $r_n(x,a,\mu) = \tilde r(x,a,\mu)$ is independent of time, the reward associated to a fixed state-action pair $(x,a)$ is $\tilde r(x,a, \mu_n)$ at time $n$ and $\tilde r(x,a, \mu_{n'})$ at time $n'$. Unless the mean-field state is stationary, in general the two reward values will be different. 
\end{remark}

\begin{example}[Crowd motion] This setting is probably the most commonly studied one in the MFG literature. As a typical example, we can think of a model for crowd motion in the spirit of e.g.~\citet{achdoulasry2019meancrowd}: the agents start from an initial position and want to reach a point of interest while avoiding crowded areas. Because the population distribution changes as the agents move, looking for a stationary solution is not satisfactory if we want to compute the evolution of the whole population. This is because a stationary solution would only give the optimal policy (from the Nash equilibrium perspective) against the stationary distribution, and would not be able to recover the full evolution of the agents. In contrast, a time-dependent policy in the evolutive setup is able to adjust the agents' behavior step by step. 
\end{example}

Several examples on grid-world state spaces will be provided in Section~\ref{sec:numerical-experiments}, with numerical illustrations. 

\begin{remark}
    Discrete time finite state mean field games have been introduced by~\citet{gomes2010discrete}. In the model analyzed therein, the players can directly control their transition probabilities. 
    Note that the model we consider here is a bit more general since the transition probabilities are functions of the actions, but they are not necessarily chosen freely by the players. 
\end{remark}

We define the (set-valued) \defi{best response map} by:
$$
    \BR_{\mathrm{evol},m_0,N_T}(\bsmu) := \argmax_{\bspi \in \sPol^{N_T}} J_{\mathrm{evol}}(\bspi; \bsmu) \subseteq \sPol^{N_T},
$$
and the \defi{population behavior map} $ \MF_{\mathrm{evol},m_0,N_T}: \sPol^{N_T} \to (\Delta_\states)^{N_T+1}$ by:
\begin{equation}
\label{eq:defi-MFevolNT}
    \MF_{\mathrm{evol},m_0,N_T}(\bspi) := \hbox{ mean field sequence when using $\bspi$ and starting from $m_0$},
\end{equation}
where this sequence is defined by:
    \begin{equation}
    \label{eq:evol-mu}
        \begin{cases}
            \mu_0=m_0, 
            \\
            \mu_{n+1} = P_{n,\mu_n,\pi_n}^\top \mu_n, \quad n \ge 0.
        \end{cases}
    \end{equation}
    When the context is clear, we will use the notation $\bsmu^{m_0,\bspi} = \MF_{\mathrm{evol},m_0,N_T}(\bspi)$. 
    
\begin{definition}[Evolutive MFG Nash Equilibrium]
A pair $(\hat\bspi,\hat\bsmu) \in \sPol^{N_T} \times \Delta_\states^{N_T+1}$ is an \defi{evolutive mean field Nash equilibrium} (evolutive MFNE) if it satisfies the following two conditions:
\begin{itemize}
    \item Best response: $\hat\bspi \in \BR_{\mathrm{evol},m_0,N_T}(\hat\bsmu)$;
    \item Mean field sequence: $\hat\bsmu = \MF_{\mathrm{evol},m_0,N_T}(\hat\bspi)$.
\end{itemize}
\end{definition}            
Alternatively, an evolutive MFNE can be defined as a fixed point: $\hat\bspi$ is an \defi{evolutive MFNE policy} if it is a fixed point of $\BR_{\mathrm{evol},m_0,N_T} \circ \MF_{\mathrm{evol},m_0,N_T}$, and $\hat\bsmu$ is an \defi{evolutive MFNE sequence} if it is the mean field sequence generated by an evolutive MFNE policy.

\paragraph{Value functions. }
The state-action value function associated to a policy $\bspi$ and the optimal state-action value function are defined analogously to standard MDP but parameterized by $\bsmu$. We denote them respectively by $\bfQ^{\bspi,\bsmu}$ and $\bfQ^{*,\bsmu}$. 

For the sake of completeness, let us provide more details in the finite horizon setting. Assume $N_T<+\infty$. %
The \defi{state-action value function} associated to a policy $\bspi$ for a given distribution $\bsmu$ is defined as: %
$$
\left\{
\begin{split}
    &\bfQ^{\bspi,\bsmu}_{N_T}(x,a) = r_{N_T}(x,a,\mu_{N_T})
    \\
    &\bfQ^{\bspi,\bsmu}_n(x,a) = \EE\left[ \sum_{n'=n}^{N_T} r_{n'}(x_{n'},a_{n'},\mu_{n'}) \Big| x_n = x, a_n = a, x_{n'+1} \sim p_{n'}(\cdot|x_{n'},a_{n'},\mu_{n'}), a_{n'} \sim \pi_{n'}(\cdot|x_{n'}) \right], 
    \\
    & \qquad\qquad\qquad n = N_T-1,\dots, 0.
\end{split}
\right.
$$
The \defi{optimal state-action value function} is defined as:
$$
    \bfQ^{*,\bsmu}(x,a) = \sup_{\bspi} \bfQ^{\bspi,\bsmu}(x,a).
$$
We can also define the (state only) value function, as in MDP, see Section~\ref{sec:background-finitehorizon-mdp}.

\subsection{Dynamic MFG settings with stationary mean field}

\subsubsection{Stationary setting}
\label{sec:statio-mfg-setting}

\paragraph{Setting definition. } Here we consider an infinite horizon model, meaning that there is no terminal time. We assume that the players interact through a \emph{stationary} distribution, which represents a steady state of the population.

\begin{definition}[Stationary MFG setting] A \defi{stationary MFG setting} is defined by a tuple $(\states, \actions, p, r, \gamma)$ consisting of:
\begin{itemize}
    \item a state space $\states$ and an action space $\actions$,
    \item a one-step transition probability kernel  
$p: \states \times \actions \times \Delta_\states \to \Delta_\states$, 
    \item a one-step reward function $r: \states \times \actions \times \Delta_\states \to \RR$, 
    \item and a discount factor $\gamma \in [0,1]$.
\end{itemize}
\end{definition}
The main difference with standard MDPs as recalled in Section~\ref{sec:background-mdp} is the presence of a third input for $p$ and $r$, which is an element of the mean field state space $\Delta_\states$. It plays the role of the population distribution, which influences the dynamics and the rewards. We will sometimes refer to the population distribution as the \defi{mean field state}.

\paragraph{Mean field game. } We start by defining a few notations and then the notion of Nash equilibrium.

Recall that in this setting, we focus on a situation in which the population distribution is stationary. However, each agent's state might be evolving. Assume the state of the population is given by  $\mu \in \Delta_\states$ and consider a representative player using policy $\pi \in \sPol$. 
The total, discounted reward of this player is given by: 
\begin{equation}
\label{eq:def-Jstatio}
    J_{\mathrm{statio},\gamma}(\pi; \mu) = \EE\left[ \sum_{n=0}^{\infty} \gamma^n r(x_n, a_n, \mu)\right],
\end{equation}
where the state of the agent evolves according to:
\begin{equation}
    \label{eq:statio-evol-player}
    \left\{\begin{split}
    &x_0 \sim \mu, 
    \\
    &x_{n+1} \sim p(\cdot|x_n, a_n, \mu), \quad a_n \sim \pi(\cdot|x_n), \quad n \ge 0.
    \end{split}
    \right.
\end{equation}
We stress that, in this setting, the initial distribution for the representative player's state is $\mu$. It is not fixed exogenously. Instead, it will be determined as part of the solution.

This stationary MFG setting has been studied for instance by~\citet{subramanian2019reinforcement} with applications to malware spread and investments in product quality,  by~\citet{guo2019learningneurips,guo2023general} with applications to auctions and by~\citet{angiulifouquelauriere2022unified} in the context of linear-quadratic MFGs.

The evolution of the population is given by a transition matrix defined by: for all $\tilde\mu\in\Delta_\states,$ $\pi \in \sPol,$ $\mu \in \Delta_\states$ and $x \in \states$,
\begin{equation}
\label{eq:transition-matrix-mf}
    (P_{\tilde\mu,\pi}^\top \mu)(y) = \sum_{x}\mu(x) \sum_{a} \pi(a|x) p(y | x, a, \tilde\mu).
\end{equation}
In words, $P_{\tilde\mu,\pi}^\top \mu$ is the next state distribution for a representative agent starting with state distribution $\mu$ and using policy $\pi$ while the population has state distribution $\tilde\mu$.%

Given a population state, the goal for a representative agent, is to find the best reaction, i.e., a policy that maximizes their total reward. We define the (set-valued) \defi{best response map} by:
$$
    \BR_{\mathrm{statio}, \gamma}: \Delta_\states \to 2^{\strategies}, 
    \quad \mu \mapsto \BR_{\mathrm{statio}, \gamma}(\mu) := \argmax_{\pi \in \strategies} J_{\mathrm{statio},\gamma}(\pi; \mu) \subseteq \strategies,
$$
and the (set-valued) \defi{population behavior map} by:%

\begin{equation}
    \label{eq:def-Mstatio}
    \MF_{\mathrm{statio}}: \sPol \to 2^{\Delta_\states}, \quad  \pi \mapsto \MF_{\mathrm{statio}}(\pi) := \{ \mu \in \Delta_\states  \,|\, \mu = P_{\mu,\pi}^\top \mu\},
\end{equation}
which is the \defi{stationary distribution} obtained when using $\pi$ (that we assume to be unique). 
Note that $\mu$ is involved in the transition matrix $P_{\mu,\pi}$, which makes the computation of a stationary distribution non-trivial in general. We come back to this point in Section~\ref{sec:mfg-br-based}. 

\begin{definition}[Stationary MF Nash Equilibrium]
\label{def:statioMFNE}
A pair $(\hat\pi, \hat\mu) \in \strategies \times \Delta_\states$ is a \defi{stationary mean field Nash equilibrium} (stationary MFNE) if it satisfies the following two conditions: \begin{itemize}
    \item $\hat\pi \in \BR_{\mathrm{statio}, \gamma}(\hat\mu)$;
    \item $\hat\mu \in \MF_{\mathrm{statio}}(\hat\pi)$.
\end{itemize}
\looseness=-1
\end{definition}
An equilibrium is thus a fixed point: $\hat\pi$ is a \defi{stationary MFNE policy} if it is a fixed point of $\BR_{\mathrm{statio}, \gamma} \circ \MF_{\mathrm{statio}}$, and $\hat\mu$ is a \defi{stationary MFNE distribution} if it is the stationary distribution of a stationary MFNE policy.

\begin{example}
    As discussed in Section~\ref{sec:evol-mfg-setting}, some applications of MFGs naturally depend on a time-evolving mean field. In some other cases, it can be meaningful to look at a situation in which the mean field is in a stationary regime. This point of view has been used for instance in MFGs for macroeconomic models~\citep{achdou2014pde,achdou2017income}, where the state of the economy can be considered as stationary, at least as an approximation of the real situation. Another example is flocking, in which we can imagine that the density representing the flock of birds is stationary, although each bird in the flock is moving. We refer to~\cite{nourian2010synthesisflocking} in continuous time (who actually used an ergodic setting -- see Section~\ref{sec:ergodic-setting}) and~\cite{perrin2021mfgflockrl} in discrete time. 
\end{example}

\paragraph{Value functions. } To characterize the best response against a given mean field state, we can adapt the notions developed for standard MDP and recalled above. 
In this setting, the \defi{state-action value function} associated to a stationary policy $\pi$ for a given distribution $\mu$ is defined as:
$$
    Q^{\pi,\mu}(x,a) = \EE\left[ \sum_{n \ge 0} \gamma^n r(x_n,a_n,\mu) \Big| x_0 = x, a_0 = a, x_{n+1} \sim p(\cdot|x_n,a_n,\mu), a_n \sim \pi(\cdot|x_n) \right].
$$
The problem then reduces to a standard stationary MDP, parameterized by $\mu$. %
By dynamic programming, $Q^{\pi,\mu}$ satisfies the fixed point equation:
\begin{equation}
    \label{eq:statio-bellman-fixed-point}
    Q = B^{\pi,\mu} Q,
\end{equation}
where $B^{\pi,\mu}$ denotes the \defi{Bellman operator} associated to $\pi$ and $\mu$:
\begin{equation}
    \label{eq:statio-bellman-op}
    (B^{\pi,\mu} Q)(x,a) = r(x,a,\mu) + \gamma \mathbb{E}_{x' \sim p(\cdot|x,a,\mu), a' \sim \pi(\cdot|x')}[Q(x',a')],
\end{equation}
where
\begin{equation}
    \label{eq:statio-expectation}
    \mathbb{E}_{x' \sim p(\cdot|x,a,\mu), a' \sim \pi(\cdot|x')}[Q(x',a')] = \sum_{x'} p(x'|x,a,\mu) \sum_{a'} \pi(a'|x') Q(x',a').
\end{equation}
The \defi{optimal state-action value function} is defined as:
$$
    Q^{*,\mu}(x,a) = \sup_{\pi} Q^{\pi,\mu}(x,a).
$$
It satisfies the fixed point equation:
\begin{equation}
    \label{eq:statio-opt-bellman-fixed-point}
    Q = B^{*,\mu} Q,
\end{equation}
where $B^{*,\mu}$ denotes the \defi{optimal Bellman operator} associated to $\mu$:
\begin{equation}
    \label{eq:statio-opt-bellman-op}
    (B^{*,\mu} Q)(x,a) = r(x,a,\mu) + \gamma \mathbb{E}_{x' \sim p(\cdot|x,a,\mu)}[ \max_{a'} Q(x',a')],
\end{equation}
with
\begin{equation}
    \label{eq:statio-opt-expectation}
    \mathbb{E}_{x' \sim p(\cdot|x,a,\mu)}[ \max_{a'} Q(x',a')] = \sum_{x'} p(x'|x,a,\mu)  \max_{a'}  Q(x',a').
\end{equation}
Note that the functions $Q^{\pi,\mu}$ and $Q^{*,\mu}$, and the operators $B^{\pi,\mu}$ and $B^{*,\mu}$ are all independent of time, as we search for stationary equilibria. 

We can also define the (state only) value function, as in MDPs, see Section~\ref{sec:background-statio-mdp}.

\subsubsection{Ergodic setting}
\label{sec:ergodic-setting}

Another setting in which the distribution of interest is stationary while the agents can move is the so-called \defi{ergodic setting}. The terminology comes from the fact that the total reward is (the limit of) a long time average. This terminology is commonly used in the PDE literature on MFGs, see for instance \citep{feleqi2013derivationmulti,bardi2014linear,arapostathis2017solutions}. 

\begin{definition}[Ergodic MFG setting]
An \defi{ergodic MFG setting} is defined by a tuple $(\states, \actions, p, r)$ where:
\begin{itemize}
    \item a state space $\states$ and an action space $\actions$,
    \item a one-step transition probability kernel
$p: \states \times \actions \times \Delta_\states \to \Delta_\states$,
    \item a one-step reward function $r: \states \times \actions \times \Delta_\states \to \RR$.
\end{itemize}
\end{definition}
Note that the model does not prescribe any initial distribution $m_0$ and there is no discount factor $\gamma$. Given a mean field $\mu \in \Delta_\states$, the total reward of a representative player using policy $\pi \in \Pi$ is defined as:
$$
    J_{\mathrm{ergo}}(\pi; \mu) = \lim_{N_T \to +\infty} \frac{1}{N_T+1} \EE\left[ \sum_{n=0}^{N_T} r(x_n, a_n, \mu)\right],
$$
subject to the  evolution~\eqref{eq:statio-evol-player} of the agent's state. This evolution is the same as in the stationary setting. The only difference is in the total reward (see~\eqref{eq:def-Jstatio}). The notion of Nash equilibrium can be defined analogously to the stationary setting (Definition~\ref{def:statioMFNE}) and we omit the details for the sake of brevity. We refer to e.g.~\cite{anahtarci2020valueiteration,anahtarci2019learningdiscountedaveragecost,wikecek2020discrete} for more details on this setting, such as the existence of Nash equilibria and the connection with finite population games.

\begin{remark}[Connection between stationary and ergodic settings]
We expect that the Nash equilibrium of a stationary setting with discount $\gamma \in (0,1)$ converges to the stationary of the corresponding ergodic setting as $\gamma \to 1$. The connection between the discounted, infinite horizon MFGs and long-time average MFGs has been studied e.g. by~\cite{bardi2012explicit,cardaliaguet2019long,gomes2020selection,cao2023stationary} for continuous models.
\end{remark}

\subsubsection{Discounted distribution setting}
\label{sec:discounted-distrib-setting}

We now present a setting that is somehow between the stationary and the evolutive ones. Note that in the stationary setting, we focus on the stationary distribution of the population while in the evolutive setting, we care about the entire sequence starting from $m_0$. In the first case, we can restrict our attention to stationary policies, whereas this is not possible in the second case, as highlighted in Remark~\ref{rem:evol-finite-infinite}. An intermediate approach consists in replacing the distribution $\mu_n^{m_0,\bspi}$ at time $n$ by an aggregate which keeps some memory of $m_0$, instead of the stationary distribution.

\begin{definition}[Discounted distribution MFG setting]
A \defi{discounted distribution MFG setting} is defined by a tuple $(\states, \actions, m_0, p, r, \gamma)$ consisting of:
\begin{itemize}
    \item a state space $\states$ and an action space $\actions$,
    \item an initial distribution  $m_0 \in \Delta_\states$,
    \item a one-step transition probability kernel
$p: \states \times \actions \times \Delta_\states \to \Delta_\states$,
    \item a one-step reward function $r: \states \times \actions \times \Delta_\states \to \RR$,
    \item a discount factor $\gamma \in [0,1)$.
\end{itemize}
\end{definition}

For the sake of brevity, we directly present the MFG model. We omit the $N$-player interpretation, which is similar to the other dynamic settings, although we come back to the interpretation of the individual agent's dynamics in Remark~\ref{rem:intepretation-discounted-distrib} below. 
The total reward is defined as in the stationary setting, i.e., when the mean field is $\mu \in \Delta_\actions$ and the player uses policy $\pi$, her reward is $J_{\mathrm{statio},\gamma}(\pi; \mu)$ as defined in~\eqref{eq:def-Jstatio}. The only difference with the stationary setting will come from the difference in the mean field term through which the interactions occur. In the present setting, instead of using the stationary distribution induced by the policy used by the population, we will use a $\gamma-$discounted distribution.

We define the \defi{discounted distribution} induced by a policy $\pi$ as:
$$
    \MF_{\mathrm{statio}, \gamma, m_0}(\pi) := \mu^{m_0,\pi}_\gamma := (1-\gamma) \sum_{n \ge 0} \gamma^n \mu^{m_0,\pi}_n \in \Delta_\states,
$$
where $\bsmu^{m_0,\pi}$ follows the dynamics~\eqref{eq:evol-mu} but with $\pi_n = \pi$ for all $n \ge 0$ after starting from $m_0$ at time $0$, and with the mean-field term replaced by $\mu^{m_0,\pi}_\gamma$, i.e.,
    \begin{equation*}
        \begin{cases}
            \mu^{m_0,\pi}_{0}=m_0, 
            \\
            \mu^{m_0,\pi}_{n+1} = P_{\mu^{m_0,\pi}_\gamma,\pi}^\top \mu^{m_0,\pi}_n, \quad n \ge 0.
        \end{cases}
    \end{equation*}

This formulation allows us to work with a single distribution, which plays the role of a summary of what happens along the mean field sequence. In contrast with the stationary MFG setting, here the initial distribution $m_0$ still influences the mean field term through which the iterations occur, namely, $\mu^{m_0,\pi}_\gamma$. 
However, we can restrict our attention to stationary policies just as in the stationary MFG setting.

\begin{example}[Exploration]
    In~\citep{perrin2020fictitious,geist2022concave}, this discounted distribution setting has been used for an MFG in which the agents explore the spatial domain. From the point of view of the population, it amounts to maximizing the entropy of the distribution. The discounted stationary distribution can be used as a proxy to evaluate with a single distribution how well the population explores the state space. This example will be described in detail and illustrated numerically in Appendix~\ref{sec:experiments-four-rooms-exploration} of the long version~\citep{lauriere2022learning}. 
\end{example}

\begin{remark}
\label{rem:intepretation-discounted-distrib}
    The discounted distribution can be interpreted as the stationary distribution of an agent who starts with distribution $m_0$, uses policy $\pi$ but has a probability to stop at any time step. To be specific, let $\tau$ be a random variable with geometric distribution on $\{0,1,2\dots,\}$ with parameter $(1-\gamma)$. Let us denote by $\mu_n^{\gamma, m_0,\pi}$ is the distribution of $x_n$, where:
    \begin{equation*}
    \left\{
    \begin{split}
        &x_0 \sim m_0, 
        \\
        &x_{n+1} \sim p(\cdot|x_n, a_n, \mu^{m_0,\pi}_\gamma ), \quad a_n \sim \pi(\cdot|x_n), \quad 0 \le n \le \tau
        \\
        &x_{n+1} = x_{n}, \quad \tau \le n
    \end{split}
    \right.
    \end{equation*}
    Then we have: for every $x \in \states$,
    \begin{align*}
        \PP(x_n = x) 
        &= \sum_{k \le n} \PP(\tau=k) \PP(x_k = x|\tau=k) + \PP(\tau > n) \PP(x_n = x|\tau > n)
        \\
        &= (1-\gamma) \sum_{k \le n} \gamma^{k}\mu_k^{\gamma, m_0,\pi}(x) + \PP(\tau > n)\PP(x_n = x|\tau > n).
    \end{align*}
    When $n \to +\infty$, $\PP(\tau > n) \to 0$, so we (formally) obtain that:
    $$
        \mu^{m_0,\pi}_\gamma(x)
        = \lim_{n \to +\infty} \PP(x_n = x) 
        =  (1-\gamma) \sum_{k} \gamma^{k}\mu_k^{\gamma, m_0,\pi}(x).
    $$
\end{remark}

\begin{definition}[Discounted distribution MFG Nash Equilibrium]
A pair $(\hat\pi, \hat\mu) \in \strategies \times \Delta_\states$ is a \defi{discounted distribution mean field Nash equilibrium} (discounted distribution MFNE) if it satisfies the following two conditions: 
\begin{itemize}
    \item $\hat\pi \in \BR_{\mathrm{statio}, \gamma}(\hat\mu)$;
    \item $\hat\mu = \MF_{\mathrm{statio}, \gamma, m_0}(\hat\pi)$.
\end{itemize}
\end{definition}
The above definition can be restated in terms of the stationary reward function and the discounted distribution: $\hat\pi$ maximizes the function $\pi \mapsto J_{\mathrm{statio},\gamma}(\pi; \hat\mu)$ and $\hat\mu = \mu_\gamma^{m_0,\hat\pi}$. Alternatively, $\hat\mu$ is the mean field of a discounted distribution MFNE if it is a fixed point of: $\MF_{\mathrm{statio}, \gamma, m_0} \circ \BR_{\mathrm{statio}, \gamma}$.

\subsection{Other solution concepts}

In this section, we present two other notions of solutions, which can be studied instead of Nash equilibria: social optimal and (coarse) correlated equilibria.

\subsubsection{Social optimum and Mean Field Control} 
\label{sec:setting-soc-opt}

The notions of MFNE discussed above correspond to non-cooperative games, in which each player maximizes their own reward while trying to anticipate the behavior of other selfish agents. A different question consists in considering that the agents are cooperative and try to maximizer a social welfare criterion by choosing together a suitable policy. This situation can also be interpreted as an optimization problem from the point of view of a social planner, who tries to figure out which behavior is optimal from the society standpoint. For the sake of brevity, we do not discuss here the connection with finite population problems and we refer the interested reader to e.g.~\citep{motte2023quantitative,carmonalaurieretan2019linearpg} for theoretical results connecting discrete time MFC problems to finite population problems.

\paragraph{Static setting.} The social welfare function is defined as the reward obtained on average by the agents:
$$
    \pi \mapsto J^{\mathrm{social}}_{\mathrm{static}}(\pi) := J_{\mathrm{static}}(\pi; \pi).
$$

\begin{definition}[Static social optimum]
A \defi{social optimum} is a strategy $\pi^*$ is  that maximizes the social welfare function $J^{\mathrm{social}}_{\mathrm{static}}$. \end{definition}

\paragraph{Stationary case. } The total, discounted social welfare associated to a policy $\pi$ is:
$$
    J^{\mathrm{social}}_{\mathrm{statio},\gamma}(\pi)
    = J_{\mathrm{statio},\gamma}(\pi; \mu^\pi)
    = \EE\left[ \sum_{n=0}^{\infty} \gamma^n r(x_n, a_n, \mu^\pi)\right]
$$
subject to the following evolution of the agent's state:
\begin{equation*}
    \left\{\begin{split}
    &x_0 \sim \mu^\pi, 
    \\
    &x_{n+1} \sim p(\cdot|x_n, a_n, \mu^\pi), \quad a_n \sim \pi(\cdot|x_n), \quad n \ge 0,
    \end{split}
    \right.
\end{equation*}
where $\mu^\pi$ is the stationary distribution induced by $\pi$. Here we see that perturbing $\pi$ changes $\mu^\pi$, which is reflected in the third argument of the transition function and the reward function. This type of problem is also referred to as \defi{mean field control (MFC)}. 
\begin{definition}[Stationary social optimum]
A \defi{stationary social optimum}, also called a \defi{solution to the stationary  MFC problem}, is a policy $\pi^* \in \sPol$ which maximizes the social welfare function $J^{\mathrm{social}}_{\mathrm{statio}}$.
\end{definition}
This setting has been considered for instance by~\citet{subramanian2019reinforcement} or by \citet{angiulifouquelauriere2022unified}.

\paragraph{Evolutive case. } The total social welfare is:
\begin{equation}
\label{eq:def-J-social-evol}
    J^{\mathrm{social}}_{\mathrm{evol}}(\bspi)
    = J_{\mathrm{evol}}(\bspi; \bsmu^{m_0,\bspi})
    = \EE\left[ \sum_{n=0}^{N_T-1} r_n(x_n, a_n, \mu^{m_0,\bspi}_n)\right],
\end{equation}
subject to the following evolution of the agent's state:
\begin{equation*}
    \left\{
    \begin{split}
        &x_0 \sim m_0, 
        \\
        &x_{n+1} \sim p_n(\cdot|x_n, a_n, \mu^{m_0,\bspi}_n), \quad a_n \sim \pi_n(\cdot|x_n), \quad n \ge 0.
    \end{split}
    \right.
\end{equation*}
\begin{definition}[Evolutive social optimum]
An \defi{evolutive social optimum}, also called a \defi{solution to the evolutive  MFC problem}, is a policy $\bspi^* \in \sPol^{N_T}$ which maximizes the social welfare function $J^{\mathrm{social}}_{\mathrm{evol}}$.
\end{definition}

The social welfare can be defined analogously in other settings.

\begin{remark}[Terminology]
The literature refers to this type of problems mean field social optimum or MFC, since it is an optimal control problem of mean-field type, in the sense that it involves mean field interactions. See~\cite{bensoussanfrehseyam2013mean}. Another terminology commonly found in the literare is \defi{optimal control of McKean-Vlasov}, since, in continuous time and spaces, the controlled dynamics is of McKean-Vlasov type because it involves the law of the process itself. We refer to~\cite{carmona2018probabilistic,carmona2018probabilistic2} and the references therein for more details.
\end{remark}

\begin{example}
    This type of problems typically has two interpretations. We can think of it as a group of players trying to cooperate in order to maximize the social welfare. This viewpoint has applications for instance in macroeconomics (see e.g.~\citep{nuno2018social}) or cybersecurity (see e.g.~\citep[Example 1]{carmona2023model}). It is particularly relevant when it is compared with Nash equilibria -- we come back to this point below, when discussing the Price of Anarchy. Another interpretation is to imagine that a central planner is trying to design an optimal control for a large group of agents, e.g. robots. An example could be to find an optimal way to make agents move from one population distribution to another, which can be called ``distribution planning''; see e.g.~\citep[Example 2]{carmona2023model}.
\end{example}

\paragraph{Price of anarchy.} For a given model, one can compare the situation where all players use a Nash equilibrium policy with the situation where all players use a socially optimal policy. 
The expected reward obtained by a representative player can only be higher in a mean field social optimum than in a MFNE, by the very definition of a social optimum. The discrepancy between the two situations is quantified by the following notion the \defi{price of anarchy} (PoA). If there are multiple NE, the PoA is defined by considering the worst equilibrium. In the static setting, it is defined as:
    \begin{equation}
        \label{eq:def-poa}
        \PoA_{\mathrm{static}} = \frac{\sup_{\pi} J^{\mathrm{social}}_{\mathrm{static}}(\pi)}
        {\inf_{\hat\pi \in \Nashs_{\mathrm{static}}} J^{\mathrm{social}}_{\mathrm{static}}(\hat\pi)}.
    \end{equation}
    \looseness=-1
    In the denominator, $\Nashs_{\mathrm{static}}$ denotes the set of static MFNE. 
The PoA can be defined analogously in the other settings.

The term ``Price of Anarchy'' has been coined by~\citet{koutsoupiaspapadimitriou1999worst}. Since then, this notion has been widely studied in game theory and can be viewed as a way to measure the inefficiency of Nash equilibria~\citep{roughgardentardos2007introduction}. In the context of MFGs, it has been studied e.g. by~\citet{lackerramanan2019rare} in a static setting, and by~\citet{carmonagravestan2019price} in a dynamic setting. Typically, we expect the price of anarchy to increase when the interactions strength increase. We refer to e.g.~\cite[Sections 2.6 and 4.4]{lauriere2021numericalams} for illustrations in an LQ model and a model of crowd motion with congestion.

\subsubsection{Mean-Field (Coarse) Correlated Equilibria}
We have discussed so far either fully decentralized decision makings, via Nash equilibria, or fully centralized control, via Mean-Field Control.  \emph{Nash equilibria} represent situations where agents do not have any incentive to unilaterally change behavior while \emph{social optima} implies a central coordination of agents. In general,  Mean Field Nash equilibria are not unique and, in such situation, proper coordination of the agents becomes crucial for selecting a relevant equilibrium. For this purpose, we now turn towards the in-between notions of equilibria with some form of centralized coordination.

Correlated equilibria and coarse correlated equilibria are widely studied in the game theory literature \cite{BlumInternalExternalRegret, morrill2021hindsight, morrill2021efficient, Aumann1987CorrelatedEA, Neumann1928,Neumann1944} and have recently been introduced in MFG settings by \citet{campi2022correlated,muller2022learningpsro, DeglInnocenti-phdthesis,muller2022learningmfce,bonesini2022correlated} (see also~\cite{campi2023coarse-open,campi2023coarse-emission} for continuous time models). They exactly fit the intuition above: a centralized instance provides a \defi{population recommendation}, which is a distribution over policies according to which the population will play. For each player, a policy is sampled from this population recommendation and sent to said player. Players only observe their policies, but they know the probabilities that the central instance assigns to each \emph{population recommendation}. From here, two deviation types that players may consider:
\begin{itemize}
    \item Deviate given a policy. Each player considers the question: ``Given that I have been tasked to play policy $\pi$, should I play something else?'' Equilibria for which there is no incentive to deviate in this way are called \defi{correlated equilibria}.
    \item Deviate in general. Each player considers the question: ``Given what typically happens, should I play $\pi'$ all the time instead of listening to recommendations?'' Equilibria for which there is no incentive to deviate this way are called \defi{coarse-correlated equilibria}.
\end{itemize}

More formally, consider for simplicity a finite set of policies, denoted by $\Pi$ (for example in the static setting, to fix the ideas).  The sets of relevant policy deviations are $\mathcal{U}_{CE} = \{ u \mid u : \Pi \rightarrow \Pi \} $ and $\mathcal{U}_{CCE} = \{ u \mid u : \Pi \rightarrow \Pi, u \text{ constant} \} $. We also write, given a \emph{population recommendation} $\nu \in \Delta_\Pi$, $\mu(\nu)$ the mean-field state distribution when the population's policies are distributed according to $\nu$. 
We denote by $\Delta_\Pi$ the set of probability distributions over policies, and by $\mathcal{P}(\Delta_\Pi)$ the set of probability distributions over $\Delta_\Pi$. 

\begin{definition}[MF  (Coarse) Correlated  Equilibrium]\label{def:mfcce}
A distribution $\rho \in \mathcal{P}(\Delta_\Pi)$ is a \textbf{mean field correlated equilibrium} if
\[
    \mathbb{E}_{\nu \sim \rho, \pi \sim \nu} \left[ J(u(\pi), \mu(\nu)) - J(\pi, \mu(\nu))  \right] \leq 0 \qquad \forall u \in \mathcal{U}_{CE}\;.
\]

Similarly, $\rho \in \mathcal{P}(\Delta_\Pi)$ is a \textbf{mean field coarse-correlated equilibrium} if
\[
    \mathbb{E}_{\nu \sim \rho, \pi \sim \nu} \left[ J(u(\pi), \mu(\nu)) - J(\pi, \mu(\nu))  \right] \leq 0 \qquad \forall u \in \mathcal{U}_{CCE}\;.
\]
\end{definition}

As detailed in \citep{muller2022learningmfce}, similarly to Nash equilibria, using a mean-field (coarse) correlated equilibrium in an $N$-player game yields an $\epsilon(N)-$(coarse) correlated equilibrium, with $\epsilon(N) = \mathcal{O} \left( \frac{1}{\sqrt{N}} \right)$. As detailed in Section~\ref{subsec:iterative-solving-mfgs} below, one motivation to study (coarse) correlated equilibrium stems from the dynamics of learning algorithms: in some situations (e.g., lack of monotonicity), it can be shown that learning algorithms such as Fictitious Play converge to (coarse) correlated equilibrium even when the Nash equilibrium is not unique. 

Given that the set of correlated equilibria strictly includes the set of Nash equilibria, its price of anarchy (see~\eqref{eq:def-poa}) will be at least as high. However, its \defi{Price of Stability} -- the ratio between the social welfare of the \emph{best} equilibrium and the optimal social welfare -- will typically be lower than with Nash equilibria, thanks to the possibility of coordinating strategies.

\subsection{Extensions} 
\label{sec:mfg-settings-extensions}
We conclude this section by mentioning a few extensions. For the sake of readability, we use the basic settings described above in the sequel. However, we conclude this section by mentioning several variants that have been considered in the literature.

\paragraph{State-action distribution. } 
    In the models with state and action spaces, we have presented the settings with interactions through the state distribution. One can also consider interactions through the distribution of actions, or even through the joint distribution of states and actions. Let us consider the stationary setting for instance. In this case, assume the state of the population is given by  $\xi \in \Delta_{\states \times \actions}$ and consider a representative agent using policy $\pi \in \sPol$. The total, discounted reward of a representative player is given by: 
$$
    J_{\mathrm{statio},\gamma}(\pi; \xi) = \EE\left[ \sum_{n=0}^{\infty} \gamma^n r(x_n, a_n, \xi)\right],
$$
where the state of the agent evolves according to:
\begin{equation*}
    \left\{\begin{split}
    &x_0 \sim \mu = \xi_1, 
    \\
    &x_{n+1} \sim p(\cdot|x_n, a_n, \xi), \quad a_n \sim \pi(\cdot|x_n), \quad n \ge 0.
    \end{split}
    \right.
\end{equation*}
with $\mu = \xi_1 \in \Delta_\states$ denoting the first marginal of $\xi$. This setting is considered for instance by \citet{guo2019learningneurips,guo2023general}. MFG with interactions through state-action distributions have first been studied  by \citet{gomes2014existence,gomesvoskanyan2016extended} and are sometimes referred to as \defi{extended MFGs} or \defi{MFGs of controls}, see~\citet{cardaliaguetlehalle2018mean,kobeissi2022classical}. Let us stress that a state-action distribution is not always a product distribution, meaning that for some $\xi \in \Delta_{\states \times \actions}$ there is no $\mu \in\Delta_\states$ and $\nu \in \Delta_\actions$ such that $\xi = \mu \otimes \nu$. In fact, in general the actions of a player are given by a function of the player's states, and hence the joint distribution cannot be written as a product. To simplify the presentation, we restrict our attention to interactions through state-only distributions but most of the ideas can be extended to state-action distributions. The interested reader is referred to~\citet[Section 4.6]{carmona2018probabilistic} and the references therein. 

\paragraph{Multiple populations. } Mean field theory allows us to approximate a homogeneous population of individuals by the limiting probability distribution. In multi-population MFGs, there is a finite number of sub-populations, each of them representing a homogeneous group of agents. The transition function and the reward function are the same for all the agents of one sub-population, but may be different from one group to the other. In this way, the MFG paradigm can still be used. We refer for instance to~\citet{huangmalhamecaines2006large,feleqi2013derivationmulti,cirant2015multi,bensoussan2018meanmulti} for an analytical approach and to~\citet[Section 7.1.1]{carmona2018probabilistic} for a probabilistic formulation. In the context of reinforcement learning, multi-population MFGs have been studied e.g. by~\citet{subramanian2020multi}. 
 A further extension, which converse multi-population MFGs as a special case, is called graphon games; see~\cite{parise2023graphon,carmona2022stochasticgraphonstatic} in the static case and e.g.~\cite{caines2019graphon,caines2021graphon,aurell2022stochastic} in the dynamic case.

\paragraph{Influencial players. } Another way to incorporate diversity in the types of players is by consider one (or several) player who has more influence than other players, who form a mean field of infinitesimal players. There are at least two different ways to include such ``influential'' players in a MFG. First, one can consider a Nash equilibrium between the major player and the mean field of minor players. Such games are generally referred to as \defi{major-minor MFGs}; see~\cite{nourian2013epsilon,carmona2016probabilistic,carmona2016finite} for a probabilistic viewpoint and~\cite{lasry2018meanmajor,cardaliaguet2020remarks} for a PDE viewpoint. A different way to include an influential player is to consider a \defi{Stackelberg}, or \defi{leader-follower}, structure, as e.g. in~\citep{bensoussan2016mean,elie2019tale,dayanikli2023machine}.

\paragraph{Mean field type games. } We discussed above MFGs and MFC problems. \defi{Mean field type games} correspond to finite-population games in which each player solves an MFC problem. We refer to~\cite{djehiche2016mean,tembine2017mean,barreiro2021mean} for more details and examples of applications. The limit when the number of players tend to infinite gives rise to a type of MFG which combines the MFG and the MFC frameworks~\cite{angiuli2022reinforcementmixedmfcg,carmona2023nash}.

\paragraph{Population-dependent policies. }
In all the previous settings, the policies are \emph{independent} of the population distribution. This aspect is classical in the MFG and MFC literature because, if a player anticipates correctly the policy used by the rest of the population, they can anticipate the whole population behavior without uncertainty. As a consequence, the distribution needs not be an input to the agent's policy. However, this aspect might be counter-intuitive from a learning perspective, because it means that the agents react optimally only to the equilibrium population behavior but they cannot adjust their behavior if the distribution deviates from this equilibrium.

Population-dependent policies are tightly connected with population-dependent value functions, and the so-called \defi{Master equation} in MFGs. This equation has been introduced by P.-L. Lions in the continuous setting (continuous time, state and action)~\citep{lionsCDF}. There, it is a partial differential equation (PDE) which corresponds to the limit of systems of Hamilton-Jacobi-Bellman PDEs characterizing Nash equilibria in symmetric $N$-player games. For more details in the continuous setting, we refer the interested reader to~\citet{BENSOUSSAN20151441,cardaliaguet2019master}. In the discrete time and space setting, population-dependent value functions and policies have been studied by~\citet{mishra2020model} and by~\citet{perrin2022generalization,wu2024population}, where a deep RL method to learn such policies is developed.

\paragraph{Common noise. } Besides idiosyncratic noise affecting the evolution of each agent independently, it is possible to consider macroscopic shocks affecting the whole population. This is referred to as \defi{common noise} in the MFG literature. Because the whole population's evolution is stochastic, using policies functions of the player's state only is in general suboptimal. This is because even if the player knows the policy used by all the other players, she cannot predict with certainty the evolution of the distribution. In this case, it is more efficient to use master (population-dependent) policies. We refer to~\citet{carmonadelaruelacker2016meancommonnoise} and to~\citet{cardaliaguet2019master} for respectively a probabilistic and an analytical treatment of MFGs with common noise.

\section{Iterative methods}
\label{sec:iterative-methods}

We now turn our attention to the question of computing mean field Nash equilibria in the settings presented above. The goal is to compute a pair consisting of a policy and a mean field which form a fixed point. A simple strategy is, starting with some initial pair, to update alternatively the policy and the mean field until convergence to an equilibrium. In this section, we assume that the model is completely known. We call the algorithms presented here \defi{iterative methods} for the sake of convenience and to distinguish them from the RL algorithms discussed later on. As we will discuss in the sequel, these methods rely on fixed point-type iterations. In contrast with the MDP setting, the underlying operator for these iterations is not always contractive, which triggers the introduction of variants to help ensuring convergence.

\subsection{Overview of the methods}
\label{sec:iterative-methods-overview}
As explained above, the main idea is to alternate an update of the population distribution and an update of the representative agent's policy, which can be represented as:
\begin{equation} 
\label{eq:iterative-methods-scheme}
    \dots \to
    \mu^{\ell} 
    \xrightarrow[]{\substack{\hbox{\,\,\,policy update\,\,\,}}} %
    \pi^{\ell+1}
    \xrightarrow[]{\substack{\hbox{\,\,\,mean field update\,\,\,}}}
    \mu^{\ell+1}
    \to \dots
\end{equation}
At a high level, we expect the scheme described in~\eqref{eq:iterative-methods-scheme} to converge towards a fixed point $(\hat\mu,\hat\pi)$ which is a Nash equilibrium.

The \defi{mean field update} is done using the population distribution or the sequence of distributions induced by the current policy. Notice that, since the dynamics is known, it is straightforward to compute the mean field induced by a given policy. The converse is more challenging: except in some special cases, given a mean field, it is hard to find which policy generated it as many policies can generate the same mean field. Thus, computing not only the mean field but also an equilibrium policy is a crucial point.

The \defi{policy update} can typically be done in two different ways. In the first family of methods, the policy is updated by computing a best response against the mean field. In the second family, the policy is updated based on the evaluation of the previous policy. We call these two families \defi{best-response based} and \defi{policy-evaluation based} respectively. In fact, this distinction stems from an analogous distinction between two families of methods to solve standard MDPs, respectively value iteration and policy iteration. 

In the rest of this section, we first recall value iteration and policy iteration methods in standard MDPs. We then explain how these methods are adapted in the MFG setting. For the sake of simplicity, we focus on two settings: the stationary setting and the finite horizon evolutive setting. We stress the main similarities and differences between the methods to solve these types of MFGs. The methods in these two settings can be adapted to tackle the static setting and the $\gamma$-discounted setting, which are thus omitted for the sake of brevity.

\subsection{Solving standard MDPs}
\label{sec:iterative-sol-statio-mdp}

We recall here two fundamental families of methods to compute optimal policies: value iteration and policy iteration. For more details on these methods, we refer to e.g.~\citep{Suttonbarto2018,bertsekasshreve1996stochastic,bertsekas2012dynamicbook,puterman2014markovbook,meyn2022controlrlbook}.

\subsubsection{Value iteration}
\label{sec:mdp-value-iteration}
One way to obtain an optimal policy is to first compute the optimal value function by using the optimal Bellman operator, and then consider a greedy policy with respect to this optimal value function. Since we are motivated by applications to RL algorithms, we focus on the state-action value function.

\paragraph{Stationary MDP. } In a stationary MDP, the \defi{value iteration} method can be expressed as follows: $Q^0$ is given, and for $k=0,\dots,K-1$, 
\begin{equation}
\label{eq:mdp-statio-value-iter}
        Q^{k+1} = B^* Q^k. %
\end{equation}
At the end we use the following policy as an approximation of the optimal policy:
$$
    \pi^K \in \Greedy Q^{K},
$$
where $\Greedy$ denotes the greedy policy operator defined by:
\begin{equation}
    \label{eq:greedy-statio-pol-op}
    \Greedy Q = \left\{\pi \,:\, \forall x \in \states, \, \sum_{a \in \actions} \pi(a|x) Q(x,a) = \argmax_{a} Q(x,a) \right\}.
\end{equation}

Thanks to the fact that the Bellman operator is a $\gamma$-contraction, when $K \to +\infty$, $\pi^K \to \pi^*$ under suitable conditions on the MDP. 
Equivalently, the above iterations can also be written as follows, by introducing the greedy policy at every iteration: $Q^0$ is given, and for $k=0,\dots,K-1$,
\begin{equation*}
\left\{ 
\begin{split}
            &\pi^{k} = \Greedy Q^k,
            \\
            &Q^{k+1} = B^{\pi^k} Q^k, %
\end{split}
\right.
\end{equation*}
where the Bellman operator $B^{\pi^k}$ associated to the current policy $\pi^k$ is defined in~\eqref{eq:mdp-statio-bellman-op}.

\paragraph{Finite horizon MDP. } In a finite horizon MDP, the optimal value function $\bfQ^{*}$ can be computed by dynamic programming since it satisfies the \defi{optimal Bellman equation}:
\begin{equation}
\label{eq:finite-horizon-opt-Bellman}
\left\{
\begin{split}
    &\bfQ^{*}_{N_T}(x,a) = r_{N_T}(x,a), \qquad (x,a) \in \states \times \actions,
    \\
    &\bfQ^{*}_n(x,a) = r_n(x,a) +  \gamma \EE\left[ \max_{a \in \actions} \bfQ^{*}_{n+1}(x_{n+1},a) \Big| x_{n+1} \sim p_n(\cdot|x,a)\right], 
    \\
    &\qquad\qquad (x,a) \in \states \times \actions, n = N_T-1,\dots, 0.
\end{split}
\right.
\end{equation}
Computing $\bfQ^{*}$ using the above equation is called \defi{backward induction}. Once it has been computed, an optimal policy can be found by taking the greedy policy at each step, i.e.:
$$
    \bspi^* = \GreedyT \bfQ^*,
$$
where $\GreedyT$ is the finite-horizon greedy policy operator defined as:
\begin{equation}
\label{eq:greedy-time-pol-op}
    (\GreedyT \bfQ)_n = \Greedy \bfQ_n, \qquad n=0,\dots,N_T-1.
\end{equation}

\begin{remark} \looseness=-1
    Notice that the Bellman equation~\eqref{eq:finite-horizon-opt-Bellman} is a backward equation and not a fixed-point equation, contrary to Eq.~\eqref{eq:statio-mdp-opt-bellman-fixed-point} characterizing the optimal value function in the stationary case. Since the horizon is finite, the optimal value function is computed with a finite number of steps, which is an important difference with the stationary MDP setting. 
\end{remark}

\subsubsection{Policy iteration}
\label{sec:mdp-policy-iteration}

The optimal policy can also be computed by successive improvements of a policy. Starting from an initial policy, at each iteration, we evaluate the performance of this policy by computing the associated value function, and then we take a greedy step to improve the policy. These two steps are called \defi{policy evaluation} and \defi{policy improvement}, and the overall algorithm is called \defi{policy iteration}.

\paragraph{Stationary MDP. } In a stationary MDP, the method consists in applying the Bellman operator $B^{\pi^k}$ associated to the current policy $\pi^k$ (see~\eqref{eq:mdp-statio-bellman-op}) and then applying the greedy policy operator defined in~\eqref{eq:greedy-statio-pol-op}. Thus, this method takes the following form: $\pi^0$ is given, and for $k=0,\dots,K-1$:
\begin{equation*}
\left\{ 
\begin{split}
        &Q^{k+1} = Q^{\pi^k},
        \\
        &\pi^{k+1}\in \Greedy Q^{k+1}. %
\end{split}
\right.
\end{equation*}
At the end, we return $\pi^K$ and use it as an approximation of $\pi^*$. 
As $K \to +\infty$, we have $\pi^K \to \pi^*$ under suitable assumptions on the MDP. 

At iteration $k$, the value function $Q^{\pi^k}$ can be computed by applying repeatedly the Bellman operator $B^{\pi^k}$ until convergence to its fixed point, or until an approximation of $Q^{\pi^k}$ is obtained with a finite number of iterations: with $Q^{k,0}$ given, we repeat for $m = 0,\dots,M-1$,
\begin{equation}
\label{eq:mdp-statio-policy-iter-evalQ}
    Q^{k,m+1} = B^{\pi^k} Q^{k,m},
\end{equation}
and we use $Q^{k,M}$ as an approximation of $Q^{\pi^k}$.

\paragraph{Finite horizon MDP. } In a finite horizon, we can define the following method by analogy with the stationary case: $\bspi^0$ is given, and for $k=0,\dots,K-1$:
\begin{equation*}
\left\{ 
\begin{split}
        &\bfQ^{k+1} = \bfQ^{\bspi^k},
        \\
        &\bspi^{k+1}\in \GreedyT \bfQ^{k+1}. %
\end{split}
\right.
\end{equation*}
where $\GreedyT$ is the finite-horizon greedy policy operator defined in~\eqref{eq:greedy-time-pol-op}. 
At the end, we return $\bspi^K$ and use it as an approximation of $\bspi^*$. 

At each iteration, the state-action value function associated to the current policy can be computed by backward induction. Indeed, for a given policy $\bspi$, $\bfQ^{\bspi}$ satisfies the following Bellman equation, which holds by dynamic programming:
\begin{equation}
    \label{eq:finite-horizon-Bellman}
    \left\{
    \begin{split}
        &\bfQ^{\bspi}_{N_T}(x,a) = 0, \qquad (x,a) \in \states \times \actions,
        \\
        &\bfQ^{\bspi}_n(x,a) = r_n(x,a) +  \gamma \EE\left[\bfQ^{\bspi}_{n+1}(x_{n+1},a_{n+1})  \Big| x_{n+1} \sim p_n(\cdot|x,a), a_{n+1} \sim \pi_{n}(\cdot|x) \right], 
    \\
    &\qquad\qquad (x,a) \in \states \times \actions, n = N_T-1,\dots, 0.
    \end{split}
    \right.
\end{equation}

\subsection{Solving MFGs}
\label{subsec:iterative-solving-mfgs}
As explained at the beginning of this section (see Eq.~\eqref{eq:iterative-methods-scheme}), the main idea underlying the methods we present below is to alternate an update of the population distribution and an update the representative agent's policy.

At iteration $\ell$, we assume we already have a candidate $\mu^\ell$ for the mean field. Inspired by the above methods for standard MDPs, we can distinguish two families of methods for MFGs, depending on whether the policy update consists in computing an optimal policy against $\mu^{\ell}$ or simply improving the current policy. We call these two families of methods respectively \defi{best-response based} and \defi{policy-evaluation based}.

\subsubsection{Methods based on best response computation}
\label{sec:mfg-br-based}
Recall that an MFG equilibrium can be defined as the fixed point of a mapping which is the composition of the best response computation and the mean field computation. Thus, a basic strategy consists in repeatedly applying this mapping. Under suitable conditions, this method converges and the limit is automatically a fixed point.

\paragraph{Stationary MFG. }

In the stationary MFG setting (see Section~\ref{sec:statio-mfg-setting}), we recall that a Nash equilibrium consists of a stationary distribution $\hat\mu\in\Delta_\states$ and a stationary policy $\hat\pi\in\sPol$. The policy $\hat\pi$ is characterized as an optimal policy for a representative player facing the population distribution $\hat\mu$. This problem can be phrased in the framework of MDPs. 

If the stationary mean field is $\mu$, then the MDP that a representative player needs to solve is:
\begin{equation}
\label{eq:statio-MFG-MDP}
    (\states, \actions, p(\cdot,\cdot,\mu), r(\cdot,\cdot,\mu), \gamma),
\end{equation}
where the transition and the reward functions are given by:
$
    p(\cdot,\cdot,\mu): \states \times \actions \to \cP(\states),$  $r(\cdot,\cdot,\mu): \states \times \actions \to \RR.
$
The optimal policy for this MDP is the best response against $\mu$, which is denoted by $\BR_{\mathrm{statio}, \gamma}(\mu)$. It can be obtained for example by applying the policy iteration or the value iteration algorithms as recalled in Section~\ref{sec:iterative-sol-statio-mdp}. Conversely, given a policy $\pi$, the induced mean field is the stationary distribution (assuming it is unique for simplicity) induced by $\pi$ and denoted by $\MF_{\mathrm{statio}}(\pi)$, see Eq.~\eqref{eq:def-Mstatio}.

This is summarized as follows: $\mu^0$ is given, and  for $\ell = 0,\dots,L-1$,
\begin{equation}
\label{eq:mfg-statio-BRbased}
\left\{ 
\begin{split}
        &\pi^{\ell+1} = \BR_{\mathrm{statio}, \gamma} (\mu^{\ell})
        \\
        &\mu^{\ell+1} = \MF_{\mathrm{statio}}(\pi^{\ell+1}). %
\end{split}
\right.
\end{equation}
At the end of these iterations, we use $(\pi^{L}, \mu^{L})$ as a proxy for the MFG equilibrium. Under suitable conditions (such as contraction property of $\MF_{\mathrm{statio}} \circ \BR_{\mathrm{statio}, \gamma}$), it is close to $(\hat\pi,\hat\mu)$ when $L$ is large enough. We come back to the question of convergence in Section~\ref{sec:iterative-convergence-variants} below.

In the above iterative method,  we update the mean field term by using the operator $\MF_{\mathrm{statio}}$, which can be approximated by applying a large number of times the transition matrix defined in~\eqref{eq:evol-mu}. In other words, in practice, $\mu^{\ell+1}$ is often defined by first computing:
\begin{equation*}
        \mu_{n+1} = P_{n,\mu_n,\pi^{\ell+1}}^\top \mu_n, \quad n = 0, \dots, M-1,
\end{equation*}
with $\mu_0$ a given initial distribution. For instance we can take $\mu_0 = \mu^{\ell}$ from the previous iteration. As $M \to +\infty$, we expect $\mu_M \to \MF_{\mathrm{statio}}(\pi^{\ell+1})$, so we use $\mu_M$ as an approximation of $\mu^{\ell+1}$.

In fact, taking $M$ relatively small can have some advantages. In some sense, it amounts to slowing down the updates of the mean-field term. This can bring more stability to the iterative method, particularly when the policy $\pi^{\ell+1}$ is computed approximately (e.g., in a reinforcement learning setup). We will come back to this idea of damping the update of the mean field in Section~\ref{sec:iterative-convergence-variants} below, but let us immediately emphasize that a variant of the above iterative method consists in doing only one application of the transition matrix at each iteration $\ell$. This can be summarized as: $\mu^0$ is given, and for $\ell = 0,\dots,L-1$,
\begin{equation}
\label{eq:algo-update-onetimestep}
\left\{ 
\begin{split}
        &\pi^{\ell+1} = \BR_{\mathrm{statio}, \gamma} (\mu^{\ell})
        \\
        &\mu^{\ell+1} = P_{n,\mu^{\ell},\pi^{\ell+1}}^\top \mu^{\ell}.
\end{split}
\right.
\end{equation}
This method has been used for instance by~\citet{guo2019learningneurips,anahtarci2020qregmfg}. It is also in line with the idea of using a two-timescale approach for mean field Nash equilibria~\citep{subramanian2019reinforcement,mguni2018decentralisedli,angiulifouquelauriere2022unified,pmlr-xie21-learning}. A similar method has been analyzed in~\citep{anahtarci2019learningdiscountedaveragecost,anahtarci2020valueiteration} for ergodic (average reward) MFGs (in the latter work, it is referred to as \defi{value iteration} algorithm for MFGs).

\paragraph{Finite-horizon MFG. } 
In the evolutive MFG setting with a finite horizon $N_T$ (see Section~\ref{sec:evol-mfg-setting}), an equilibrium is a sequence of distributions $\hat\bsmu = (\hat\mu_n)_{n=0,\dots,N_T}$ and a sequence of policies $\hat\bspi = (\hat\pi_n)_{n=0,\dots,N_T}$, indexed by the time steps in the game. Given a sequence of distributions $\hat\bsmu = (\hat\mu_n)_{n=0,\dots,N_T}$, a  representative player needs to solve the following finite-horizon MDP (see Section~\ref{sec:background-finitehorizon-mdp}): 
$$
    (\states, \actions, p_{\bsmu},  r_{\bsmu}, N_T),
$$
where $p_{\bsmu}: \{0,\dots,N_T-1\} \times \states \times \actions \to \cP(\states)$ and $r_{\bsmu}: \{0,\dots,N_T\} \times \states \times \actions \to \RR$ are given by:
$$
    p_{\bsmu}: (n,x,a) \mapsto p_n(\cdot|x,a,\mu_n), \qquad    r_{\bsmu}: (n,x,a) \mapsto r_n(x,a,\mu_n).
$$
The optimal policy for this MDP is the best response against $\bsmu$, which is denoted by $\BR_{\mathrm{evol},m_0,N_T}(\bsmu)$ . It can be obtained as a greedy policy for the optimal value function $\bfQ^{*,\bsmu}$, which can be computed by backward induction as described in Section~\ref{sec:mdp-value-iteration}.  
Alternatively, the optimal policy can be computed by policy iteration as described in Section~\ref{sec:mdp-policy-iteration}. Conversely, given a policy $\bspi=(\pi_n)_{n=0,\dots,N_T}$, the induced mean-field is the sequence of distributions generated by starting from $m_0$ (remember that $m_0$ is fixed in the definition of the MFG, see Section~\ref{sec:evol-mfg-setting}) and using $\pi_n$ at time step $n$, $n=0,\dots,N_T-1$. The resulting mean-field sequence is denoted by $\MF_{\mathrm{evol},m_0,N_T}(\bspi)$, see Eq.~\eqref{eq:defi-MFevolNT}. 

This is summarized below, using the notation introduced in Section~\ref{sec:evol-mfg-setting}: $\bsmu^0$ is given, and for $\ell = 0,\dots,L-1$, 
\begin{equation}
\label{eq:iterative-mfg-evol-policy-eval}
\left\{ 
\begin{split}
        &\bspi^{\ell+1} = \BR_{\mathrm{evol},m_0,N_T} (\bsmu^{\ell})
        \\
        &\bsmu^{\ell+1} = \MF_{\mathrm{evol},m_0,N_T}(\bspi^{\ell+1}). %
\end{split}
\right.
\end{equation}
At the end of these iterations, we use $(\bspi^{L}, \bsmu^{L})$ as a proxy for the MFG equilibrium. Under suitable conditions (such as contraction property of $\MF_{\mathrm{evol},m_0,N_T} \circ \BR_{\mathrm{evol},m_0,N_T}$) on the MFG, this pair is close to $(\hat\bspi,\hat\bsmu)$ when $L$ is large enough.

For the sake of completeness and future reference, we provide here the Bellman equations satisfied by $\bfQ^{*,\bsmu}$ and $\bfQ^{\bspi,\bsmu}$, which can be derived by dynamic programming:
\begin{equation}
\label{eq:mfg-finite-horizon-opt-Bellman}
\left\{
\begin{split}
    &\bfQ^{*,\bsmu}_{N_T}(x,a) = r_{N_T}(x,a,\mu_{N_T})
    \\
    &\bfQ^{*,\bsmu}_n(x,a) = r_n(x,a,\mu_{n}) +  \EE\left[ \max_{a \in \actions} \bfQ^{*,\bsmu}_{n+1}(x_{n+1},a) \Big| x_{n+1} \sim p_n(\cdot|x_{n},a_{n},\mu_{n})\right], 
    \\
    & \qquad\qquad\qquad n = N_T-1,\dots, 0,
\end{split}
\right.
\end{equation}
and
\begin{equation}
    \label{eq:mfg-finite-horizon-Bellman}
    \left\{
    \begin{split}
        &\bfQ^{\bspi,\bsmu}_{N_T}(x,a) = r_{N_T}(x,a,\mu_{N_T})
        \\
        &\bfQ^{\bspi,\bsmu}_n(x,a) = r_n(x,a,\mu_{n}) +   \EE\left[\bfQ^{\bspi,\bsmu}_{n+1}(x_{n+1},a_{n+1})  \Big| x_{n+1} \sim p_n(\cdot|x_{n},a_{n},\mu_{n}), a_{n+1} \sim \pi_{n+1}(\cdot|x_{n+1}) \right], 
        \\
        & \qquad\qquad\qquad n = N_T-1,\dots, 0.
    \end{split}
    \right.
\end{equation}

\citet{perrin2020fictitious,perrin2022generalization} used backward induction  to compute the optimal value function for finite-horizon MFG (embedded in Fictitious Play iterations, see Section~\ref{sec:iterative-convergence-variants}),
which served as a baseline to assess the performance of RL-based methods (see next section). \citet{cui2021approximately} solved finite-horizon MFG using fixed point iterations combined with RL methods and entropy regularization (we come back to this point in Section~\ref{sec:iterative-convergence-variants} below). \cite{mishra2020model} also solved MFGs based on a best-response computation, but by computing a best response backward in time in the spirit of dynamic programming, which requires solving for all possible distributions since the equilibrium mean field sequence is not known a priori. The aforementioned two-timescale approach originally studied in the stationary setting has been extended by~\citet{angiulifouquelauriere2021reinforcementhandbook} to solve finite-horizon MFGs. \citet{muller2022learningpsro} proposed an adaptation to MFGs of the \defi{Policy Space Response Oracles approach}, see~\citep{lanctot2017unified}. This method iteratively computes a best-response to a distribution generated by optimally mixing already-computed best response policies from previous iterations. This method is shown to converge to mean field Nash, correlated and coarse-correlated equilibria in all finite MFGs.

\begin{remark}[Time steps vs iterations]
    As alreay mentioned, several works focusing on the stationary use a single time step in each iteration, hence identifying time steps of the game and iterations of the learning algorithm. See~\eqref{eq:algo-update-onetimestep}. 
    However, in the finite-horizon MFG setting, the index of iterations does not coincide with the index of time in the game. At each iteration $\ell$, the policy and the distributions are updated for all time steps, $n=0,\dots,N_T$. 
\end{remark}

\begin{remark}[Non-uniqueness of best responses]
    Methods based on best response computation (~\eqref{eq:mfg-statio-BRbased} and~\eqref{eq:mfg-finite-horizon-Bellman}) require computing a best response. However, optimal policies are in general non-unique. 
    In such cases, the choice of the optimal policy used in the iteration may be important in order to ensure convergence.  
    This ambiguity can be problematic in practice. See e.g.~\cite[Section 2.3]{lauriere2021numericalams} for a LQ MFG example where fixed point iterations oscillate between two candidates, non of them being a Nash equilibrium. We come back to this aspect in Section~\ref{sec:iterative-convergence-variants} below. Numerical illustrations will be provided in Section~\ref{sec:numerical-experiments}. 
\end{remark}

\subsubsection{Methods based on policy evaluation}

Instead of computing a full-fledged best response for the policy update at each iteration of~\eqref{eq:iterative-methods-scheme}, we can simply do one step of policy improvement. Intuitively, evaluating the current policy should be computationally faster than computing an optimal policy (except when the state space is small or when we have an explicit formula for the optimizer of the value function). To improve the policy, we can first evaluate the current policy given the latest mean field, and then take a greedy policy.

\paragraph{Stationary MFG. } 
In a stationary MFG, we can proceed as follows: we first compute the state-action value function associated to the current policy against the current population distribution (policy evaluation step). We then define the new policy as a greedy policy for the newly computed value function (policy improvement step). Last, we deduce the stationary population distribution induced by this policy (mean field update step). Concretely, the method is: $\mu^0$ and $\pi^0$ are given, and  for $\ell = 0,\dots,L-1$,
\begin{equation}
\label{eq:mfg-statio-policy-eval}
\left\{ 
\begin{split}
        &Q^{\ell+1} = Q^{\pi^\ell, \mu^\ell}
        \\
        &\pi^{\ell+1} \in \Greedy Q^{\ell+1}
        \\
        &\mu^{\ell+1} = \MF_{\mathrm{statio}}(\pi^{\ell+1}). %
\end{split}
\right.
\end{equation}
This method is referred to as the \defi{Policy Iteration} (PI) algorithm for MFGs and was introduced by \cite{cacacecamilligoffi2021pimfg} for continuous time, continuous space MFGs. \cite{camilli2022rates} proved a quadratic rate of convergence under suitable assumptions. It is not to be confused with the method that consists in using standard policy iteration to compute a best response against a given distribution (i.e., replacing $\BR_{\mathrm{statio},\gamma} (\mu^{\ell})$  in~\eqref{eq:mfg-statio-BRbased} by the result of a policy iteration method).

\looseness=-1
In practice, the evaluation step can be done by applying a finite number of times the Bellman operator $B^{\pi^\ell,\mu^\ell}$ as defined in Eq.~\eqref{eq:statio-bellman-op}. Thanks to the contraction property of this operator, we obtain an approximation of $Q^{\pi^\ell, \mu^\ell}$. Furthermore, as discussed above, $\MF_{\mathrm{statio}}(\pi^{\ell+1})$ can be approximated by applying a large but finite number of times the transition matrix.

\paragraph{Finite-horizon MFG. }
In a finite-horizon MFG, the same strategy can be applied, except that we need to take into account the evolutive aspect of the game. Each of the step is done for all the time steps. The method can be summarized as follows:  $\mu^0$ and $\pi^0$ are given, and  for $\ell = 0,\dots,L-1$,
\begin{equation}
\label{eq:finiteHmfg-policy-iteration-method}
\left\{ 
\begin{split}
        &\bfQ^{\ell+1} = \bfQ^{\bspi^\ell, \bsmu^\ell}
        \\
        &\bspi^{\ell+1} \in \GreedyT \bfQ^{\ell+1}
        \\
        &\bsmu^{\ell+1} = \MF_{\mathrm{evol},m_0,N_T}(\bspi^{\ell+1}). 
\end{split}
\right.
\end{equation}
In this setting, $\bfQ^{\bspi^\ell, \bsmu^\ell}$ can be computed by backward induction, thanks to the dynamic programming equation~\eqref{eq:mfg-finite-horizon-Bellman}. Similarly, $\MF_{\mathrm{evol},m_0,N_T}(\bspi^{\ell+1})$ can be computed by following $N_T$ transitions, see~\eqref{eq:evol-mu}. 

\citet{cacacecamilligoffi2021pimfg}, mentioned above, also studied policy iteration in the finite-horizon setting and proved convergence under suitable conditions. Still in the finite-horizon setting, the convergence results were extended to other settings by~\citet{camilli2022rates,lauriere2023policy}. Using a purely greedy policy often leads to instabilities, particularly in the finite state case; see \textit{e.g.}~\citep{cui2021approximately} and the next section for more details. 
For this reason, variants with regularized policies have been introduced, such as the Online Mirror Descent, as we explain below.

\begin{remark}[Non-uniqueness of greedy policies]
    Methods based on policy evaluation (see~\eqref{eq:mfg-statio-policy-eval} and~\eqref{eq:finiteHmfg-policy-iteration-method}) may suffer from the existence of multiple greedy policies, it was the case for best response based methods with respect to optimal policies. We come back to this aspect in Section~\ref{sec:iterative-convergence-variants} below. Numerical illustrations will be provided in Section~\ref{sec:numerical-experiments}. 
\end{remark}

\subsection{Convergence and variants}
\label{sec:iterative-convergence-variants}

\paragraph{Convergence of fixed point iterations. } Intuitively, the scheme described in~\eqref{eq:iterative-methods-scheme} indeed converges towards a fixed point if the mapping $(\mu^{\ell}, \pi^{\ell}) \mapsto (\mu^{\ell+1}, \pi^{\ell+1})$ is a strict contraction on a suitably defined space. In a stationary setting, this property can be ensured by assuming that the reward function and the transition function are smooth enough. Typically, this amounts to assuming that they are Lipschitz continuous with small enough Lipschitz constants. In a finite horizon setting, this condition can sometimes be replaced by an assumption on the smallness of the time horizon. One advantage of having a contraction is that it provides a constructive way to get the equilibrium through Banach-Picard iterations. This technique is commonly used in the literature on MFGs, both to show existence and to derive algorithms. See e.g.~\citet{huangmalhamecaines2006large} in the context of existence and uniqueness of the equilibrium or~\citet{carlini2014fully} in the context of numerical methods. It is in general difficult to formulate sufficient conditions on the model (i.e., the reward and the transition) to ensure the strict contraction property because the mapping involves the policy update step, for which there is in general no explicit formula. In the linear-quadratic case, several sufficient conditions are formulated by~\citet[Proposition~3.1]{hu2021deepfpdiff}. Furthermore, regularizing the policy can help to alleviate some of the conditions ensuring the contraction property, see e.g.~\citet{guo2019learningneurips,anahtarci2020qregmfg}. Using regularization of the policy, \citet{guo2023general} have proved convergence and analyzed the complexity of value-based and policy-based algorithms. 

However, conditions guaranteeing the strict contraction property are generally very restrictive and fails to hold for many games. For example, \citet[Theorem 2]{cui2021approximately} show that non-contractivity is the rule rather than the exception. Without contractivity, Banach-Picard iterations typically lead to oscillations, see e.g.~\citet[Figure 3]{chassagneux2019numerical} in the context of a method based on the probabilistic interpretation of MFGs, or~\citet[Figure 4]{lauriere2021numericalams} in the context of linear-quadratic MFGs. We will also provide numerical illustrations in discrete time and finite state MFGs in Section~\ref{sec:numerical-experiments}. 
To address this issue, several variants of the pure Banach-Picard fixed point iterations have been proposed in the literature, relying on a few key principles.

Before describing these principles, let us mention that besides the aforementioned class of assumptions to ensure contractivity which are somehow \emph{quantitative assumptions} since they boil down to smallness of some coefficients, an alternative class of hypotheses are in some sense \emph{qualitative assumptions} which pertain to the structure of the game. For example, potential structure and MFG satisfying Lasry-Lions monotonicity~\citep{lasrylions-japanMR2295621} can be used to prove convergence of best-response based and policy evaluation based algorithms, see respectively~\citep{cardaliaguethadikhanloo2017learningfictitious,perrin2020fictitious,geist2022concave} and~\citep{hadikhanloo2017learninganonumous,perolat2022scaling}. In particular, the Lasry-Lions monotonicity condition, which basically refers to the fact that players tend to avoid crowded regions,  has been interpreted in terms of exploitability (see Section~\ref{sec:metrics}). These convergence results do not rely on smallness conditions on the coefficients. However, even for MFGs with such nice structure, pure fixed point iterations rarely converge and smoothing the iterations if typically required to ensure convergence. 

\paragraph{Smoothing the mean field updates. } 
First, a simple modification consists in using damping to slow down the updates of the mean field term. Even if the mapping $\mu^{\ell} \to \pi^{\ell} \to \mu^{\ell+1}$ is not contractive, we can hope that the following mapping is contractive, at least for small enough values of $\alpha \in (0,1)$:
\begin{equation}
    \label{eq:damped-fixedpoint-iterations}
    \tilde\mu^{\ell} \to \pi^{\ell} \to \tilde\mu^{\ell+1} := (1-\alpha) \tilde\mu^{\ell} + \alpha \mu^{\ell+1}.
\end{equation}
Here $\mu^{\ell+1}$ is the mean field associated to policy $\pi^{\ell}$ while $\tilde\mu^{\ell}$ is an average over past mean field terms. See e.g.~\citet[Section 2]{lauriere2021numericalams} for an example in which damping with a constant coefficient helps ensuring numerical convergence. We also refer to \citet{tembine2012meansurvey} for more algorithms developed along these lines and presented in the context of static games.

We can also let $\alpha$ change with the iteration index, i.e., take a different $\alpha^{\ell}$ for $\ell=1,2,\dots$. 
One of the most popular versions consist in taking $\alpha^{\ell} = 1/(\ell+1)$ and is called \defi{Fictitious Play}. It was first introduced in two-player games by~\citet{brown1951iterative,robinson1951iterative} and extended to MFG by~\citet{cardaliaguethadikhanloo2017learningfictitious,hadikhanloo2018learningphd,hadikhanloosilva2019finite}. In the context of stationary MFGs for example, \eqref{eq:mfg-statio-BRbased} is replaced by: $\mu^0$ is given, and  for $\ell = 0,\dots,L-1$,
\begin{equation}
\label{eq:mfg-statio-BRbased-fictitiousplay}
\left\{ 
\begin{split}
        &\pi^{\ell+1} = \BR_{\mathrm{statio}, \gamma} (\tilde\mu^{\ell})
        \\
        &\mu^{\ell+1} = \MF_{\mathrm{statio}}(\pi^{\ell+1})
        \\
        &\tilde\mu^{\ell+1} = \frac{\ell}{\ell+1} \tilde\mu^{\ell} + \frac{1}{\ell+1} \mu^{\ell+1}. %
\end{split}
\right.
\end{equation}
Under suitable assumptions, $\tilde\mu^{\ell}$ converges to a stationary MFG equilibrium distribution. It is important to note that in general the last iterate $\pi^{\ell}$ of the policy does \emph{not} generate $\tilde\mu^{\ell}$ and hence does not converge towards an equilibrium policy. If one cares about the equilibrium policy, it is thus required to learn a policy generating $\tilde\mu^{\ell}$. In some cases, convergence of the last iterate towards an equilibrium holds, see e.g.~\citet{cardaliaguethadikhanloo2017learningfictitious}. 

For finite-state MFGs, a rate of convergence of $O(1/t)$ has been obtained by~\citet{perrin2020fictitious} for continuous-time FP under monotonicity condition and by~\citet{geist2022concave,lavigne2023generalized} respectively for discrete-time FP in some potential MFGs. In linear-quadratic MFGs, a rate of convergence has been obtained by~\citet{delarue2024exploration}, who also studied the impact of common noise. Fictitious play has also been used in combination with finite-difference scheme to solve the MFG PDE system in~\citep{inoue2023fictitious}. 

Slowing down the updates of the mean field term is also in line with the idea of using a two-timescale approach for mean field Nash equilibria~\citep{subramanian2019reinforcement,mguni2018decentralisedli,angiulifouquelauriere2022unified,pmlr-xie21-learning}. Here, the distribution and the policy (or the value function) are both updated at every iteration but the distribution is updated at a slower rate than the policy. Intuitively, this implies that the representative agent has enough time to compute an approximate best response before the distribution changes too much. We refer to e.g.~\cite{angiuli2023convergence} for a proof of convergence, under suitable assumptions, using two-timescale and stochastic approximation arguments. 

Whenever Nash equilibria are not unique, one can use a slight alteration of the Fictitious Play algorithm to learn coarse correlated equilibria. \citet{muller2022learningmfce} defined \textbf{Joint Fictitious Play}, in the continuous-time update regime, as:
\begin{equation*}
    \pi^{BR}_t = \argmax_{\pi' \in \Pi} \int \limits_{s=0}^t \langle \mu^{\pi'}, r^{\pi'}(\cdot, \mu^{\pi_{s}}) \rangle ds
    \quad\mbox{and}\quad
    \mu^{\pi_t}(x) = \frac{1}{t} \int \limits_0^t \mu^{\pi^{BR}_s}(x) ds.
\end{equation*}
This algorithm has been proved to converge to a mean field coarse correlated equilibrium. More precisely, the distribution which uniformly samples $\pi^{BR}_t$ with $t \in [0, T]$, and recommends this policy to the whole population, converges towards a mean field coarse correlated equilibrium at a rate of $\mathcal{O}\left ( \frac{1}{T} \right )$.

\looseness=-1
\paragraph{Smoothing the policy updates. } Another way to bring more stability to the iterative method is to regularize the policy update. For instance, the greedy policy operator defined in~\eqref{eq:greedy-statio-pol-op} is very sensitive to perturbations of the state-action value function. Small changes in this value function might lead to significant changes in the induced greedy policy. 
To mitigate this problem, it is common to replace the $\argmax$ by a $\softmax$, meaning that we can define:
\begin{equation}
    \label{eq:softmax-statio-pol-op}
    \pi^{(k+1)}(\cdot|x) = \softmax_\tau Q(x,\cdot),
\end{equation}
where $\tau>0$ is an inverse temperature parameter and $\softmax: \RR^{|\actions|} \to \Delta_{\actions}$ is defined by: for $q = (q_1,\dots,q_{|\actions|})$, 
$$
    \softmax_\tau (q) = \left(\frac{e^{\tau q_i}}{\sum_{j=1}^{|\actions|} e^{\tau q_j}}\right)_{i=1,\dots,|\actions|}.
$$
It transforms a vector of Q-values into a discrete probability distribution on the action space in which the actions with larger value have a higher probability. Using a softmax instead of the argmax generally yields smoother and more stable learning curves, see e.g.~\citet{guo2019learningneurips,anahtarci2020qregmfg}. 

In fact, finite-state finite-action MFGs typically admit only randomized policy equilibria and no pure equilibria. This is also the reason why we generally allow for randomized policies in finite-player games~\citep{nash1950equilibrium,nash1951non}. Hence, iterative methods with pure greedy policies cannot be expected to converge to Nash equilibria in general, and using mixed policies is unavoidable. 

Regularized policies can be obtained e.g. by directly changing the way the policy is obtained from the value function~\citep{guo2019learningneurips,perolat2022scaling} or by adding a penalty in the reward function, which changes the value function and hence the policy, see e.g.~\citet{anahtarci2019fittedqmfg,guo2022entropy,cui2021approximately,firoozi2022exploratory,lauriere2022scalable}. However, it should be noted that regularizing the policies also has drawbacks: if $\pi^{\ell}$ is forced to be smooth, this constraint might prevent the iterative method from converging towards the Nash equilibrium since  $\pi^{\ell}$ can only be smooth version of the equilibrium policy. 

\looseness=-1
One way to circumvent this limitation and to allow the regularized policy to concentrate on optimal actions is to let the underlying Q-function take larger and larger values. This can be achieved by considering a cumulative Q-function, which leads to the \defi{Online Mirror Descent (OMD)} algorithm~\citep{hadikhanloo2017learninganonumous,perolat2022scaling}:
\begin{equation}
\label{eq:mfg-statio-policy-eval-omd}
\left\{ 
\begin{split}
        &Q^{\ell+1} = Q^{\pi^\ell, \mu^\ell}
        \\
        &\tilde Q^{\ell+1} = \tilde Q^{\ell} + \alpha Q^{\ell+1}
        \\
        &\pi^{\ell+1} = \softmax_\tau \tilde Q^{\ell+1}
        \\
        &\mu^{\ell+1} = \MF_{\mathrm{statio}}(\pi^{\ell+1}). %
\end{split}
\right.
\end{equation}
where $\alpha>0$ is a parameter which determines the cumulative factor. 
This algorithm can be viewed as a modification of the policy evaluation method described in~\eqref{eq:mfg-statio-policy-eval} with a cumulative Q-function and a regularized greedy policy.  Instead of the $\softmax$, we can more generally take the gradient of the convex conjugate of a strongly convex regularizer, see~\citet{perolat2022scaling} for more details. 
{In situations where the MFG does not admit a unique Nash equilibrium,  \cite{muller2022learningmfce} showed that recommending an OMD policy at uniformly-sampled times to the whole population yields a mean field coarse-correlated equilibrium as presented in Definition~\ref{def:mfcce}. 

Another way to smooth policy updates is to take the average of the new policy (best response) and the past policies. This method, called {\bf Smoothed Policy Iteration}, has been introduced in \cite{lauriere2023policy,tang2024learning}, which also proved a rate of convergence under suitable assumptions, in continuous time and space MFG models.

\subsection{Solving mean field social optimum}
\label{sec:iterative-methods-mfc}

We recall that the mean field social optimum or MFC problem introduced in Section~\ref{sec:setting-soc-opt} corresponds to the maximization of a social reward. In the evolutive case, it can be reformulated as an MDP by considering the population distribution as the state. Indeed, we can rewrite $J^{\mathrm{social}}_{\mathrm{evol}}$ defined in~\eqref{eq:def-J-social-evol} using the viewpoint of a representative agent, as follows by using the viewpoint of the population distribution:
$$
    J^{\mathrm{social}}_{\mathrm{evol}}(\bspi)
    =  \sum_{n=0}^{N_T-1} \underbrace{\sum_{x \in \states} \sum_{a \in \actions} r_n(x, a, \mu^{m_0,\bspi}_n) \mu^{m_0,\bspi}_n(x) \pi_n(a|x)}_{=:\bar r_n(\mu^{m_0,\bspi}_n, \pi_n)},
$$
subject to the following evolution of the mean field state:
\begin{equation*}
    \left\{
    \begin{split}
        &\mu^{m_0,\bspi}_0 = m_0, 
        \\
        &\mu^{m_0,\bspi}_{n+1}  =  P_{n,\mu^{m_0,\bspi}_{n},\pi_{n}}^\top \mu^{m_0,\bspi}_{n}, \quad n \ge 0.
    \end{split}
    \right.
\end{equation*}
This is an MDP with: state space $\Delta_\states$, 
    action space $\sPol$,
    probability transition function: $\bar p_n(\cdot|\mu, \pi) = (P^{\mu, \pi}_n)^\top \mu$,
    reward function: $ \bar r_n(\mu, \pi) = \sum_{x \in \states} \sum_{a \in \actions} r_n(x, a, \mu) \mu(x) \pi(a|x)$.

We will refer to this MDP as the \defi{mean field MDP} (MFMDP), following the terminology introduced by~\cite{carmona2023model}. An action, taken by the central planner or collectively by the population, is an element of $\overline{\actions} = (\Delta_\actions)^{\states}$. A one-step policy at the level of the population is a function from $\overline{\states}$ to $\Delta_{\overline{\actions}}$. Note that, even if $\states$ and $\actions$ are finite, the state space $\overline{\states}$ and the action space $\overline{\actions}$ of the MFMDP are continuous and hence rigorously defining and analyzing this MDP requires a careful formulation. We refer to the work of e.g.~\cite{gast2012mean,gu2023dynamicmfc,gu2021meanQ,carmona2023model,motte2022mean,bauerle2021meanfieldmdp} for more details on MFMDP.

Let us stress that this MFMDP is not to be confused with the MDP arising in MFGs, which is the MDP for a single representative player when the mean field term is given. In the latter case, the state is simply the agent's state and not the population state.

With this reformulation, the evolutive MFC problem can be analyzed and solved using methods developed from MDP. However, notice that the policies are, in general, functions of both the representative agent's state and the mean field state, which can be a high dimensional vector whose coordinates take continuous values.  
The main challenges thus pertain to the numerical implementation of these methods, since we need to represent efficiently the distribution and the policy. We will come back to this question in Section~\ref{sec:rl-algorithms-mfc}.

\begin{remark}
Note that, in the present model, the evolution of $\bsmu^{m_0,\bspi}$ is in fact completely deterministic once $m_0$ and $\bspi$ are given. Noise affecting the distribution and making its evolution stochastic is referred to as common noise, as already mentioned in Section~\ref{sec:mfg-settings-extensions}. We refer to~\citep{motte2022mean,carmona2023model} for more details on the rigorous treatment of MFMDPs with common noise. Furthermore, motivated by applications in RL, one may want to sample random policies at the population level. In other words, an action in the MFMDP is an element of $\sPol$, so a (mixed) policy is a function $\bar\pi: \Delta_\states \ni \mu \mapsto \bar\pi(\mu) \in \Delta_\sPol$. Sampling from $\bar\pi(\mu)$ amounts to sample an element $\pi$ to be used by the whole population. \citet{carmona2023model} have introduced this type of randomness for both theoretical and numerical purposes. It is referred to this as \defi{common randomness}. 
\end{remark}

\section{Reinforcement learning algorithms}
\label{sec:rl-algorithms}

The iterative methods presented in the previous section are described with ``exact'' updates, in the sense that we assume the model is fully known and that there are no numerical approximations in the computation of the rewards, the transitions or the expectations.

However, in many situations, these methods cannot be implemented as such. A typical scenario is when the model is not completely known from the agent that is trying to learning an optimal behavior. Another instance is when the model is known, but the state space or the action space are too big for us to compute the solution on the whole domain. In such cases, exact dynamic programming cannot be used. Instead \defi{(model-free) reinforcement learning (RL)} methods have been developed. Here we will focus on methods relying on \defi{approximate dynamic programming (ADP)} based on value functions. 
Other model-free methods could be used, such as policy gradient methods but we will mostly omit them in the presentation for the sake of brevity. We will however  mention relevant references in the sequel. 
The question of exploring efficiently the state-action domain plays a crucial role. 

RL ideas have first been developed for finite and small state and action spaces, in which case the algorithms are called  \defi{tabular methods} since the value function can be described by a table (i.e., a matrix). However, many of the recent breakthrough applications of RL have been obtained thanks to a combination of RL methods with neural network approximations and deep learning techniques, which leads to \defi{deep reinforcement learning (DRL) methods}. The flexibility and the generalization capabilities of deep neural networks allow us to efficiently learn solution to highly complex problems. In the context of games, some striking examples that were successfully tackled are ALE (Atari Learning Environment)~\citep{mnih2013playingatari, Bellemare_2013}, Go~\citep{silver2016mastering},  poker~\citep{Brown17Libratus,moravvcik2017deepstack} or Starcraft~\citep{vinyals2019grandmaster}.

In the context of MFGs, we will build on the iterative methods presented in Section~\ref{sec:iterative-methods}. These methods boil down to alternating mean-field updates and policy updates, and the policy updates stem from standard MDP techniques. As a consequence, standard RL techniques can readily be injected at this level to learn policies or value functions.  
We refer to e.g.~\citep{perrin2022scaling} for more background on the motivations of RL and deep RL for MFGs.

In the rest of this section, we will focus on the dynamic settings. RL for static MFGs as introduced in Section~\ref{sec:static-mfne} has thus far been the focus of less attention, with the exception of a few works, such as~\citep{gummadi2013meanbandit,iyer2014meanauction,maghsudi2017distributedbandit,wang2021meanbandit,yardim2023stateless}. However, it is worth noting that some algorithms and results obtained in the dynamic setting can be readily applied to the static setting. Indeed, static MFGs can be recast as dynamic MFGs with a single time step and a single step. This connection has been used for instance by~\cite{muller2022learningpsro,muller2022learningmfce}, which studied several static MFGs with this viewpoint such as mean-field extensions of the celebrated rock-paper-scissors game.

In the rest of this section, starting from exact dynamic programming, we discuss some key ideas underlying ADP and RL methods. We then move on to neural network approximations and DRL. Finally we explain how these ideas can be adapted to the MFG setting.

\subsection{Background on reinforcement learning}
\label{sec:background-RL}

\paragraph{Environment. }
Traditional RL aims at solving a stationary MDP, see Section~\ref{sec:background-statio-mdp}.  In the typical setting, the agent who is trying to find an optimal policy for the MDP interacts with an environment through experiments that can be summarized as follows: 
\begin{enumerate}
    \item The agent observes the current state $x_n$ of the MDP (which is referred to as the state of the environment but could be for instance her own state or the state of the world).
    \item The agent takes an action $a_n$, which is going to influence the state of the MDP through the transition kernel $p$ and produces a reward through the function $r$.
    \item The agent observes the new state $x_{n+1} \sim p(\cdot|x_n,a_n)$ as well as the reward $r(x_n,a_n)$ resulting from her action. 
\end{enumerate}
The agents can repeat such experiments. We provide in Fig.~\ref{fig:rl-env-classical} a schematic representation of this setting. It is often assumed that the agent can reboot the environment from time to time. To avoid remaining stuck in local maxima, it is common to assume that the new state is picked randomly, which is referred to an exploring start. 

We stress that the agent does not observe directly the functions $p$ and $r$ that are used to compute the new state and the reward. The agent only observes the outputs of these functions. In some cases, recovering the functions $p$ and $r$ from such observations is feasible, leading to the concept of \defi{model-based RL}. However, for complex environments (i.e., complex $p$ and $r$), recovering the functions would require such a large number of observations that we generally prefer to directly aim for an optimal policy, which leads to the concept of \defi{model-free RL}. The agent needs to interact multiple times to figure out the most suitable actions for a given state of the world. For more details, we refer the interested reader to e.g.~\citet{Suttonbarto2018,bertsekas2012dynamicbook,szepesvari2010algorithms,meyn2022controlrlbook}.

\begin{figure}
    \centering
    \resizebox{0.5\textwidth}{!}{%
    \begingroup
        \tikzset{every picture/.style={scale=0.3}}%
        \tikzset{every picture/.style={line width=0.75pt}} %

\begin{tikzpicture}[x=0.75pt,y=0.75pt,yscale=-1,xscale=1]

\draw   (220,227.2) .. controls (220,222.67) and (223.67,219) .. (228.2,219) -- (321.8,219) .. controls (326.33,219) and (330,222.67) .. (330,227.2) -- (330,251.8) .. controls (330,256.33) and (326.33,260) .. (321.8,260) -- (228.2,260) .. controls (223.67,260) and (220,256.33) .. (220,251.8) -- cycle ;
\draw    (220,250) -- (122,250) ;
\draw [shift={(120,250)}, rotate = 360] [color={rgb, 255:red, 0; green, 0; blue, 0 }  ][line width=0.75]    (10.93,-3.29) .. controls (6.95,-1.4) and (3.31,-0.3) .. (0,0) .. controls (3.31,0.3) and (6.95,1.4) .. (10.93,3.29)   ;
\draw    (120,250) -- (120,152) ;
\draw [shift={(120,150)}, rotate = 450] [color={rgb, 255:red, 0; green, 0; blue, 0 }  ][line width=0.75]    (10.93,-3.29) .. controls (6.95,-1.4) and (3.31,-0.3) .. (0,0) .. controls (3.31,0.3) and (6.95,1.4) .. (10.93,3.29)   ;
\draw   (235,44) .. controls (235,39.03) and (239.03,35) .. (244,35) -- (301,35) .. controls (305.97,35) and (310,39.03) .. (310,44) -- (310,71) .. controls (310,75.97) and (305.97,80) .. (301,80) -- (244,80) .. controls (239.03,80) and (235,75.97) .. (235,71) -- cycle ;
\draw    (120,50) -- (233,50) ;
\draw [shift={(235,50)}, rotate = 180] [color={rgb, 255:red, 0; green, 0; blue, 0 }  ][line width=0.75]    (10.93,-3.29) .. controls (6.95,-1.4) and (3.31,-0.3) .. (0,0) .. controls (3.31,0.3) and (6.95,1.4) .. (10.93,3.29)   ;
\draw    (220,230) -- (152,230) ;
\draw [shift={(150,230)}, rotate = 360] [color={rgb, 255:red, 0; green, 0; blue, 0 }  ][line width=0.75]    (10.93,-3.29) .. controls (6.95,-1.4) and (3.31,-0.3) .. (0,0) .. controls (3.31,0.3) and (6.95,1.4) .. (10.93,3.29)   ;
\draw    (150,230) -- (150,152) ;
\draw [shift={(150,150)}, rotate = 450] [color={rgb, 255:red, 0; green, 0; blue, 0 }  ][line width=0.75]    (10.93,-3.29) .. controls (6.95,-1.4) and (3.31,-0.3) .. (0,0) .. controls (3.31,0.3) and (6.95,1.4) .. (10.93,3.29)   ;
\draw    (150,70) -- (233,70) ;
\draw [shift={(235,70)}, rotate = 180] [color={rgb, 255:red, 0; green, 0; blue, 0 }  ][line width=0.75]    (10.93,-3.29) .. controls (6.95,-1.4) and (3.31,-0.3) .. (0,0) .. controls (3.31,0.3) and (6.95,1.4) .. (10.93,3.29)   ;
\draw    (430,249.8) -- (430,50) ;
\draw [shift={(430,251.8)}, rotate = 270] [color={rgb, 255:red, 0; green, 0; blue, 0 }  ][line width=0.75]    (10.93,-3.29) .. controls (6.95,-1.4) and (3.31,-0.3) .. (0,0) .. controls (3.31,0.3) and (6.95,1.4) .. (10.93,3.29)   ;
\draw    (310,50) -- (428,50) ;
\draw [shift={(430,50)}, rotate = 180] [color={rgb, 255:red, 0; green, 0; blue, 0 }  ][line width=0.75]    (10.93,-3.29) .. controls (6.95,-1.4) and (3.31,-0.3) .. (0,0) .. controls (3.31,0.3) and (6.95,1.4) .. (10.93,3.29)   ;
\draw    (430,251.8) -- (332,251.8) ;
\draw [shift={(330,251.8)}, rotate = 360] [color={rgb, 255:red, 0; green, 0; blue, 0 }  ][line width=0.75]    (10.93,-3.29) .. controls (6.95,-1.4) and (3.31,-0.3) .. (0,0) .. controls (3.31,0.3) and (6.95,1.4) .. (10.93,3.29)   ;
\draw  [dash pattern={on 4.5pt off 4.5pt}]  (70,150) -- (200,150) ;
\draw    (220,227.2) .. controls (174.46,197.3) and (233.79,164.66) .. (228.38,217.38) ;
\draw [shift={(228.2,219)}, rotate = 277.05] [color={rgb, 255:red, 0; green, 0; blue, 0 }  ][line width=0.75]    (10.93,-3.29) .. controls (6.95,-1.4) and (3.31,-0.3) .. (0,0) .. controls (3.31,0.3) and (6.95,1.4) .. (10.93,3.29)   ;
\draw    (150,150) -- (150,72) ;
\draw [shift={(150,70)}, rotate = 450] [color={rgb, 255:red, 0; green, 0; blue, 0 }  ][line width=0.75]    (10.93,-3.29) .. controls (6.95,-1.4) and (3.31,-0.3) .. (0,0) .. controls (3.31,0.3) and (6.95,1.4) .. (10.93,3.29)   ;
\draw    (120,150) -- (120,52) ;
\draw [shift={(120,50)}, rotate = 450] [color={rgb, 255:red, 0; green, 0; blue, 0 }  ][line width=0.75]    (10.93,-3.29) .. controls (6.95,-1.4) and (3.31,-0.3) .. (0,0) .. controls (3.31,0.3) and (6.95,1.4) .. (10.93,3.29)   ;

\draw (234,233) node [anchor=north west][inner sep=0.75pt]   [align=left] {Environment};
\draw (254,49) node [anchor=north west][inner sep=0.75pt]   [align=left] {Agent};
\draw (75,174) node [anchor=north west][inner sep=0.75pt]   [align=left] {Reward};
\draw (80,193) node [anchor=north west][inner sep=0.75pt]    {$r_{n+1}$};
\draw (154,175) node [anchor=north west][inner sep=0.75pt]   [align=left] {State};
\draw (164,196) node [anchor=north west][inner sep=0.75pt]    {$x_{n+1}$};
\draw (433,139) node [anchor=north west][inner sep=0.75pt]   [align=left] {Action};
\draw (447,160) node [anchor=north west][inner sep=0.75pt]    {$a_{n}$};
\draw (74,84) node [anchor=north west][inner sep=0.75pt]   [align=left] {Reward};
\draw (79,104) node [anchor=north west][inner sep=0.75pt]    {$r_{n}$};
\draw (155,85) node [anchor=north west][inner sep=0.75pt]   [align=left] {State};
\draw (165,106) node [anchor=north west][inner sep=0.75pt]    {$x_{n}$};

\end{tikzpicture}
    \endgroup
    }
    \caption{Reinforcement learning environment: classical single-agent setup. Here, at iteration $n$, the current state of the MDP is $x_n$, the action taken by the agent is $a_n$, the new state is $x_{n+1} \sim p(\cdot|x_n,a_n)$ and the reward is $r_n = r(x_n,a_n)$. The new state $x_{n+1}$ is observed by the agent and is also used for the next step of the environment's evolution. } \label{fig:rl-env-classical}
\end{figure}
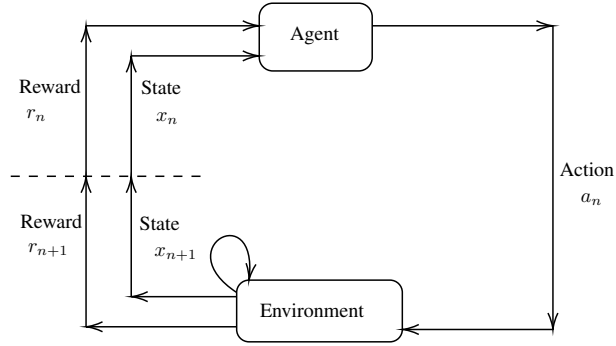

\paragraph{Approximate dynamic programming. } Some of the most popular RL methods are based on approximations of the exact dynamic programming equations satisfied by the value functions. Focusing on a stationary MDP, let us recall that an optimal policy can be computed for instance by value iteration or policy iteration (see Section~\ref{sec:iterative-sol-statio-mdp}), which require computing the state-action value functions $Q^{*}$ or $Q^{\pi}$ respectively. These two functions satisfy fixed-point equations, see~\eqref{eq:mdp-statio-value-iter} and~\eqref{eq:mdp-statio-policy-iter-evalQ}, whose solutions can be approximated by repeatedly applying the corresponding Bellman operators $B^*$ and $B^{\pi}$. This amounts to repeating:
\begin{align*}
    &Q^{\pi}(x,a) \leftarrow r(x,a) + \gamma \mathbb{E}_{x' \sim p(\cdot|x,a), a' \sim \pi(\cdot|x')}[Q(x',a')], \qquad \forall (x,a) \in \states\times\actions
    \\
    &Q^{*}(x,a) \leftarrow r(x,a) + \gamma \mathbb{E}_{x' \sim p(\cdot|x,a)}[ \max_{a'} Q^*(x',a')], \qquad \forall (x,a) \in \states\times\actions.
\end{align*}
The arrow is used to denote that the value of $Q^{\pi}(x,a)$ is replaced by the value in the right hand side. As written above, the updates are \defi{synchronous} in the sense that all state-action pairs are updated simultaneously. Furthermore, the update is based on an expectation, which is in general impossible to compute perfectly.

In the context of RL, we assume that the agent does not know $r$ or $p$, so she cannot perform the above updates. However, these updates can be performed approximately provided we assume that the agent can query the environment and ask, for any pair $(x,a)$, the value of $r(x,a)$ and a sample $x' \sim p(\cdot|x,a)$ (picked independently at each realization). Then to update $Q^{\pi}(x,a)$ and $Q^*(x,a)$, the agent can query multiple times the pair $(x,a)$ and replace the expectations by empirical averages:
\begin{align*}
    &\tilde Q^{\pi}(x,a) \leftarrow r(x,a) + \gamma \frac{1}{I} \sum_{i=1}^I \tilde Q(x^i,a^i), \qquad x^i \sim p(\cdot|x,a), a^i \sim \pi(\cdot|x^i), i=1,\dots,I, \quad \forall (x,a) \in \states\times\actions
    \\
    &\tilde Q^{*}(x,a) \leftarrow r(x,a) + \gamma \frac{1}{I} \sum_{i=1}^I  \max_{a'} \tilde Q^*(x^i,a'), \qquad x^i \sim p(\cdot|x,a), i=1,\dots,I, \quad \forall (x,a) \in \states\times\actions,
\end{align*}
where the Monte Carlo samples $x^i$ and $a^i$ are independent. However, it is generally too computationally expensive to update every pair $(x,a)$ using a batch of $I$ samples. Furthermore, in many scenarios the agent does not have the freedom to query any state $x$. Instead, she is usually bound to observe the state of the environment, which is updated iteration after iteration in a sequential manner by following the dynamics of the state. She can influence the evolution of the state, but she cannot pick any new state that she wants at every iteration. In such scenarios, the agent can only perform updates by using the available information at each iteration. 

To be specific, let us assume that the agent has a policy $\tilde\pi$ that she uses to generate a trajectory by interacting with the environment: $x_0 \sim m_0$ and for $n \ge 0$, $a_n \sim \tilde\pi(\cdot|x_n), \quad x_{n+1} \sim p(\cdot|x_n, a_n)$. 
The fixed-point equation satisfied by $Q^*$ says that $\tilde Q^*(x,a)$ is well estimated if:
\begin{equation}
\label{eq:mdp-optQ-equality-n}
    \tilde Q^{*}(x,a) = r(x,a) + \gamma  \EE_{x' \sim p(\cdot | x, a), a' \sim \pi(\cdot|x')}\left[\max_{a'} \tilde Q^*(x',a')\right].
\end{equation}
So it is natural to use the discrepancy between the right hand side and the left hand side to improve the estimate $\tilde Q^*$ of $Q^*$. Since we are bound to follow the trajectory, we cannot get the expectation over $x'$ and, instead, we perform sampled-based updates using one sample at each step and a learning $\alpha>0$:
\begin{align*}
    &\tilde Q^{*}(x_{n},a_{n}) \leftarrow \tilde Q^{*}(x_{n},a_{n}) + \alpha \left[ r(x_{n},a_{n}) + \gamma  \max_{a'} \tilde Q^*(x_{n+1},a') - \tilde Q^{*}(x_{n},a_{n})\right].
\end{align*}
This leads to the celebrated \defi{Q-learning} algorithm introduced by~\cite{watkins1989learningphd} and whose convergence under suitable conditions has been proved by~\citet{watkins1992q}. Estimating correctly the whole function $Q^*$ can be ensured if every pair $(x,a)$ is visited infinitely often, which can be guaranteed under some assumptions on the dynamics of the state and by taking $\tilde\pi$ for instance as an $\epsilon$-greedy policy (according to which in every state, every action has some probability to be selected). To estimate $\tilde Q^{\pi}$, a similar strategy can be used.

\paragraph{Deep reinforcement learning. } 
When state and action spaces are finite, a state-action value function is simply a matrix, which can be stored in memory and processed easily when the spaces are small enough. Thanks to this, we can update the value function point by point (one point being a state-action pair in the case of Q-functions). However, when the spaces are very large or even continuous, it becomes impossible to evaluate precisely every pair $(x,a)$. Furthermore, it is also impossible to visit all pairs during training, implying that the question of \defi{generalization} (i.e., performance on unvisited pairs) cannot be avoided. Motivated by both efficiency and generalization capabilities, we can approximate the state-action value functions by parameterized non-linear functions such as neural networks. For example, let us approximate $Q^{*}$ by a neural network $Q_\theta$ with a given architecture and parameters $\theta$. Going back to~\eqref{eq:mdp-optQ-equality-n}, we note that now, not only we do not know the expected value, but also we cannot update the function only at a specific pair without changing its value at other pairs. Instead, we use the discrepancy between the left hand side and an estimation of the right hand side to define a loss function that can be used to train the neural network $Q_\theta$. Since this neural network appears in both sides of the equation, to make the learning process more stable, we introduce an auxiliary neural network $Q_{\theta_{\mathrm{target}}}$ called \defi{target network} and we use it to replace $Q_\theta$ in the right hand side. The parameters $\theta_{\mathrm{target}}$ are fixed when we update $\theta$, and are updated from time to time but less frequently than $\theta$. To be specific, we define the loss:
\begin{align}
    L(\theta; \theta_{\mathrm{target}})
    = \EE_{x,a}\left[ \left| Q_{\theta}(x,a) - r(x, a) - \EE_{x' \sim p(\cdot |x,a)}\left[\max_{a'} Q_{\theta_{\mathrm{target}}}(x',a') \right]\right|^2 \right],
    \label{eq:critic_loss}
\end{align}
where the expectation is over state-action pairs. We do not specify here the distribution of this pair, but in practice it typically comes from played trajectories stored in a replay buffer.
In practice, this loss is minimized using stochastic gradient descent on mini-batches sampled from this replay buffer. This leads to the \defi{DQN algorithm} introduced by~\cite{mnih2013playingatari}.

The above approach uses the fact that we can easily compute the maximal value of $Q_{\theta_{\mathrm{target}}}(x',\cdot)$ over the action space. This is typically possible only if the action space is finite and not too large. Otherwise, we can use another neural network -- called \defi{policy network} -- to encode a policy and train this neural network so that it approximates an optimal policy, using a so-called policy loss. Then, in loss \eqref{eq:critic_loss}, the term $\max_{a'} Q_{\theta_{\mathrm{target}}}(x',a')$ is replaced by the expectation of the target $Q$-value according to the learnt policy. The resulting agent is called an \defi{actor-critic} (the actor is the policy, the critic is the state-action value function). Since our goal is not to present an exhaustive list of methods, we simply mention below three popular approaches, to give an idea of the variety of algorithms:
\begin{itemize}
    \item If we consider only deterministic policies, we can replace the policy by a parameterized function $\varphi_{\omega}: \states \to \actions $ with parameters $\omega$. In this case, the corresponding policy loss optimises for
    $$
        \max_{\omega}\EE_{x'}[Q_{\theta_{\mathrm{target}}}(x',\varphi_{\omega}(x'))].
    $$
    Its gradient can be obtained using the chain rule.
    This leads to an algorithm that is reminiscent of the the \defi{Deep Deterministic Policy Gradient (DDPG)} of~\cite{lillicrap2016continuous}. 
    \item Another approach is to look for a general stochastic policy $\pi$, in which case we have the interpretation:
    $$
        \EE_{a' \sim \pi(\cdot|x')}[Q_{\theta_{\mathrm{target}}}(x',a')] = \int_{ \actions} Q_{\theta_{\mathrm{target}}}(x',a') \pi(a'|x') da'.
    $$
    Taking the gradient and using the log trick yields the state-wise gradient:
    $$
        \EE_{a' \sim \pi(\cdot|x')}\left[ Q_{\theta_{\mathrm{target}}}(x',a') \nabla \log \pi(a'|x') \right].
    $$
    \looseness=-1
    The resulting algorithm is reminiscent of \defi{REINFORCE} \citep{williams1992simple}.
    The related empirical gradient requires sampling from the learnt policy and is usually of high variance. An alternative approach consists in using the reparameterization trick.
     For example, we can restrict our attention to Gaussian policies
    $$
        \pi_{\omega}(\cdot|x) = \Phi_{ \cN(m_{\omega}(x),  \sigma_{\omega}(x) I)}(\cdot),
    $$
    where $\Phi_{ \cN(m_{\omega}(x),  \sigma_{\omega}(x) I)}$ denotes the density function of the normal distribution $\cN(m_{\omega}(x),  \sigma_{\omega}(x) I)$, with $m_\omega$ and $\sigma_\omega$ being two parameterized functions with parameters $\omega$. Here $I$ denotes the identity matrix on $\actions$. This leads to the following approximation:
    $$
        \EE_{a' \sim \pi(\cdot|x')}[Q_{\theta_{\mathrm{target}}}(x',a')]
        \approx 
        \EE_{a' \sim \pi_\omega(\cdot|x')}[Q_{\theta_{\mathrm{target}}}(x',a')]
        = \EE_{\epsilon \sim \cN(0,I)} [Q_{\theta_{\mathrm{target}}}(x',m_\omega(x) + \sigma_\omega(x) \epsilon)].
    $$
    Here again, we can replace the expectation by an average over a finite number of realizations of $\epsilon$. In this way, we obtain an algorithm which is similar to \defi{TD3} \citep{fujimoto2018addressing}.
\end{itemize}

Since the goal of this section is simply to describe some of the key ideas behind DRL, we do not go further into a detailed presentation of the variety of existing methods. We refer the interested reader to e.g.~\citet{arulkumaran2017deeprlsurvey,franccois2018introduction}.

\subsection{Reinforcement learning for MFGs}

We now turn our attention to (model-free) RL methods for MFGs. The iterative methods presented in Section~\ref{sec:iterative-methods} allow us to solve an MFG by iteratively updating the population distribution and the policy, and the policy update can be done using standard MDP techniques. As a consequence, RL techniques to solve MDPs can be directly adapted to solve MFGs in a model-free fashion.

\subsubsection{Environments with mean field interactions} 

To study RL methods for MFGs, the first question is the definition of the environment, which will generate  individual agent's rewards and transitions. In MFGs, the transitions and the rewards depend on the population distribution, which should thus be part of environment. Since we are going to focus on how a representative agent learns an equilibrium policy, we also include the state of this representative agent in the state of the environment. 

The next question is: What is the information available to the agent who is learning? In other words, we should decide what  the output of one query to the environment is. Remember that for a given population distribution (or sequence of distributions in the evolutive setting), the agent tries to solve an MDP parameterized by this distribution but the policy is not a function of the population distribution  (see e.g. Section~\ref{sec:background-RL} in the stationary MFG case). From this point of view, to learn an optimal policy, she does not need to observe the rest of the population: it is sufficient to observe the result of the reward function and some samples of transitions. 
 
\begin{remark}
    The fact that the representative agent does not need to observe the rest of the population in order to learn an optimal policy is specific to the mean-field setting with an infinite number of players (and without common noise, see Section~\ref{sec:mfg-settings-extensions}). In a finite-player game, the equilibrium policy of each player generally depends on the configuration of the rest of the population, even when the interactions are symmetric or when a mean-field approximation is used, see e.g.~\citet{yang2018mean,yang2020overviewmarl,zhangyangbasar2021multiagentrloverview,cui2022survey} in the context of MARL. This is because in a mean-field setting, the law of large numbers allows to get rid of the randomness of the evolution of the crowd, provided that the crowd is always starting from the same initial distribution and all the players are anonymous, identical and use the same policy (which is exactly the mean-field setting). With a finite number of players, even if the players are always starting from the same set of states, having a stochastic environment (\textit{e.g.} a stochastic transition kernel) compels the agents to keep track of the current states of other agents. Thus, focusing on population-independent policies (which are sometimes referred to as decentralized policies) is one of the main advantage of the MFG approach compared with a finite-player game framework. We stress that this is possible because the agents always start from the same initial distribution. Relaxing this constraint requires population-dependent policies \citep{perrin2022generalization,wu2024population}.
\end{remark}

We summarize the environment in Fig.~\ref{fig:mfg-env}, which is very similar to the classical RL setup described in Fig.~\ref{fig:rl-env-classical} except that the distribution is involved in the environment. 

\begin{remark}[On the implementation of the environment]
In some cases, the environment is truly based on the mean field state corresponding to the regime with an infinite number of agents. This can be the case for example when the state space and the action space are finite and small, and the evolution of the distribution or the stationary distribution can be computed exactly using the transition matrix. This approach is implemented in \texttt{OpenSpiel}, for a wide class of MFGs; see Section~\ref{sec:numerical-experiments} and Appendix~\ref{app:intro-mfg-openspiel} of the long version~\citep{lauriere2022learning}. However, in general, the environment relies on some approximate version of the population distribution (e.g., using an empirical distribution with a finite number of agents, or using some function approximations). Compared with the ideal environment with the true mean-field distribution, this adds an extra layer of approximation which can be neglected if one is purely interested in the performance of RL algorithms. We come back to this point in Section~\ref{sec:rl-algorithms-mfc} below, in the context of MFC.
\end{remark}

\begin{figure}
    \centering
    \resizebox{0.5\textwidth}{!}{%
    \begingroup
        \tikzset{every picture/.style={scale=0.3}}%
        \tikzset{every picture/.style={line width=0.75pt}} %

\begin{tikzpicture}[x=0.75pt,y=0.75pt,yscale=-1,xscale=1]

\draw   (220,227.2) .. controls (220,222.67) and (223.67,219) .. (228.2,219) -- (321.8,219) .. controls (326.33,219) and (330,222.67) .. (330,227.2) -- (330,251.8) .. controls (330,256.33) and (326.33,260) .. (321.8,260) -- (228.2,260) .. controls (223.67,260) and (220,256.33) .. (220,251.8) -- cycle ;
\draw    (220,250) -- (122,250) ;
\draw [shift={(120,250)}, rotate = 360] [color={rgb, 255:red, 0; green, 0; blue, 0 }  ][line width=0.75]    (10.93,-3.29) .. controls (6.95,-1.4) and (3.31,-0.3) .. (0,0) .. controls (3.31,0.3) and (6.95,1.4) .. (10.93,3.29)   ;
\draw    (120,250) -- (120,152) ;
\draw [shift={(120,150)}, rotate = 450] [color={rgb, 255:red, 0; green, 0; blue, 0 }  ][line width=0.75]    (10.93,-3.29) .. controls (6.95,-1.4) and (3.31,-0.3) .. (0,0) .. controls (3.31,0.3) and (6.95,1.4) .. (10.93,3.29)   ;
\draw   (235,44) .. controls (235,39.03) and (239.03,35) .. (244,35) -- (301,35) .. controls (305.97,35) and (310,39.03) .. (310,44) -- (310,71) .. controls (310,75.97) and (305.97,80) .. (301,80) -- (244,80) .. controls (239.03,80) and (235,75.97) .. (235,71) -- cycle ;
\draw    (120,50) -- (233,50) ;
\draw [shift={(235,50)}, rotate = 180] [color={rgb, 255:red, 0; green, 0; blue, 0 }  ][line width=0.75]    (10.93,-3.29) .. controls (6.95,-1.4) and (3.31,-0.3) .. (0,0) .. controls (3.31,0.3) and (6.95,1.4) .. (10.93,3.29)   ;
\draw    (220,230) -- (152,230) ;
\draw [shift={(150,230)}, rotate = 360] [color={rgb, 255:red, 0; green, 0; blue, 0 }  ][line width=0.75]    (10.93,-3.29) .. controls (6.95,-1.4) and (3.31,-0.3) .. (0,0) .. controls (3.31,0.3) and (6.95,1.4) .. (10.93,3.29)   ;
\draw    (150,230) -- (150,152) ;
\draw [shift={(150,150)}, rotate = 450] [color={rgb, 255:red, 0; green, 0; blue, 0 }  ][line width=0.75]    (10.93,-3.29) .. controls (6.95,-1.4) and (3.31,-0.3) .. (0,0) .. controls (3.31,0.3) and (6.95,1.4) .. (10.93,3.29)   ;
\draw    (150,70) -- (233,70) ;
\draw [shift={(235,70)}, rotate = 180] [color={rgb, 255:red, 0; green, 0; blue, 0 }  ][line width=0.75]    (10.93,-3.29) .. controls (6.95,-1.4) and (3.31,-0.3) .. (0,0) .. controls (3.31,0.3) and (6.95,1.4) .. (10.93,3.29)   ;
\draw    (430,249.8) -- (430,50) ;
\draw [shift={(430,251.8)}, rotate = 270] [color={rgb, 255:red, 0; green, 0; blue, 0 }  ][line width=0.75]    (10.93,-3.29) .. controls (6.95,-1.4) and (3.31,-0.3) .. (0,0) .. controls (3.31,0.3) and (6.95,1.4) .. (10.93,3.29)   ;
\draw    (310,50) -- (428,50) ;
\draw [shift={(430,50)}, rotate = 180] [color={rgb, 255:red, 0; green, 0; blue, 0 }  ][line width=0.75]    (10.93,-3.29) .. controls (6.95,-1.4) and (3.31,-0.3) .. (0,0) .. controls (3.31,0.3) and (6.95,1.4) .. (10.93,3.29)   ;
\draw    (430,251.8) -- (332,251.8) ;
\draw [shift={(330,251.8)}, rotate = 360] [color={rgb, 255:red, 0; green, 0; blue, 0 }  ][line width=0.75]    (10.93,-3.29) .. controls (6.95,-1.4) and (3.31,-0.3) .. (0,0) .. controls (3.31,0.3) and (6.95,1.4) .. (10.93,3.29)   ;
\draw  [dash pattern={on 4.5pt off 4.5pt}]  (70,150) -- (200,150) ;
\draw    (220,227.2) .. controls (174.46,197.3) and (233.79,164.66) .. (228.38,217.38) ;
\draw [shift={(228.2,219)}, rotate = 277.05] [color={rgb, 255:red, 0; green, 0; blue, 0 }  ][line width=0.75]    (10.93,-3.29) .. controls (6.95,-1.4) and (3.31,-0.3) .. (0,0) .. controls (3.31,0.3) and (6.95,1.4) .. (10.93,3.29)   ;
\draw    (150,150) -- (150,72) ;
\draw [shift={(150,70)}, rotate = 450] [color={rgb, 255:red, 0; green, 0; blue, 0 }  ][line width=0.75]    (10.93,-3.29) .. controls (6.95,-1.4) and (3.31,-0.3) .. (0,0) .. controls (3.31,0.3) and (6.95,1.4) .. (10.93,3.29)   ;
\draw    (120,150) -- (120,52) ;
\draw [shift={(120,50)}, rotate = 450] [color={rgb, 255:red, 0; green, 0; blue, 0 }  ][line width=0.75]    (10.93,-3.29) .. controls (6.95,-1.4) and (3.31,-0.3) .. (0,0) .. controls (3.31,0.3) and (6.95,1.4) .. (10.93,3.29)   ;
\draw [color={rgb, 255:red, 208; green, 2; blue, 27 }  ,draw opacity=1 ]   (330,225) .. controls (370.59,189.36) and (308.27,172.34) .. (319.63,218.58) ;
\draw [shift={(320,220)}, rotate = 254.85] [color={rgb, 255:red, 208; green, 2; blue, 27 }  ,draw opacity=1 ][line width=0.75]    (10.93,-3.29) .. controls (6.95,-1.4) and (3.31,-0.3) .. (0,0) .. controls (3.31,0.3) and (6.95,1.4) .. (10.93,3.29)   ;

\draw (234,233) node [anchor=north west][inner sep=0.75pt]   [align=left] {Environment};
\draw (254,49) node [anchor=north west][inner sep=0.75pt]   [align=left] {Agent};
\draw (75,174) node [anchor=north west][inner sep=0.75pt]   [align=left] {Reward};
\draw (80,193) node [anchor=north west][inner sep=0.75pt]    {$r_{n+1}$};
\draw (154,175) node [anchor=north west][inner sep=0.75pt]   [align=left] {State};
\draw (164,196) node [anchor=north west][inner sep=0.75pt]    {$x_{n+1}$};
\draw (310,147) node [anchor=north west][inner sep=0.75pt]  [color={rgb, 255:red, 208; green, 2; blue, 27 }  ,opacity=1 ] [align=left] {Distribution};
\draw (339,167) node [anchor=north west][inner sep=0.75pt]  [color={rgb, 255:red, 208; green, 2; blue, 27 }  ,opacity=1 ]  {$\mu _{n}$};
\draw (433,139) node [anchor=north west][inner sep=0.75pt]   [align=left] {Action};
\draw (447,160) node [anchor=north west][inner sep=0.75pt]    {$a_{n}$};
\draw (74,84) node [anchor=north west][inner sep=0.75pt]   [align=left] {Reward};
\draw (79,104) node [anchor=north west][inner sep=0.75pt]    {$r_{n}$};
\draw (155,85) node [anchor=north west][inner sep=0.75pt]   [align=left] {State};
\draw (165,106) node [anchor=north west][inner sep=0.75pt]    {$x_{n}$};

\end{tikzpicture}
    \endgroup
    }
\caption{Environment for MFGs: Here,  $x_n$ is the representative agent's state, $\mu_n$ is the population distribution, $a_n$ is the action taken by the agent. The new state is $x_{n+1} \sim p(\cdot|x_n,a_n,\mu_n)$ and the reward is $r_n = r(x_n,a_n,\mu_n)$. The new state $x_{n+1}$ is observed by the agent and is also used for the next step of the environment's evolution along with $\mu_n$. }
\label{fig:mfg-env}
\end{figure}
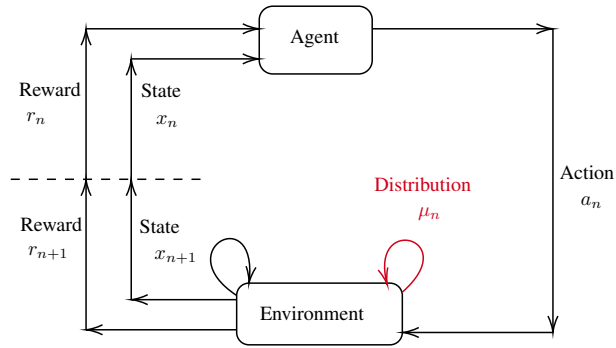

\subsubsection{Reinforcement learning for MFGs}

We focus on two settings: stationary and finite horizon. The ideas developed in these cases can be adapted to tackle static and infinite horizon MFGs.

\paragraph{Stationary MFG setting. } In a nutshell, in the stationary MFG setting, when using one of the iterative methods presented in Section~\ref{sec:iterative-methods}, at each iteration the representative agent faces a stationary MDP parameterized with a fixed distribution $\mu$. We can thus use off-the-shelf RL methods.

To be more specific, we assume that a representative agent is encoded by a stationary policy $\pi \in \sPol$, either explicitly or implicitly (through a $Q$-function) and can interact with the environment in the following way: at each step, the agent observes its current state $x$, chooses action $a \sim \pi(\cdot|x)$, and the environment returns a realization of $x' \sim p(\cdot|x,a,\mu)$ and $r(x,a,\mu)$. Note that the agent does not need to observe directly the mean field sequence $\mu$, which is stored in the environment and simply enters as a parameter of the transition and reward functions $p$ and $r$. In this stationary setting, in Fig.~\ref{fig:mfg-env}, $\mu_n$ is constant equal to $\mu$ for all $n$. Based on such samples, the representative agent can implement any of the RL methods (e.g. the ones discussed in Section~\ref{sec:background-RL}) for standard MDPs. 

Notice that, at each new step of the iterative method described in~\eqref{eq:iterative-methods-scheme}, after the mean field update the environment needs to be updated with the new population distribution. That is to say, when the mean field state is $\mu^{\ell}$, the agent uses the environment of Fig.~\ref{fig:mfg-env} with $\mu_n = \mu^{\ell}$ for every $n$ to learn a best-response or evaluate a policy. Here $n$ is the index of the RL method iteration. Then, the new mean field $\mu^{\ell+1}$ is computed. For the next iteration, the MDP is updated so the agent interacts with the environment of Fig.~\ref{fig:mfg-env} but now with $\mu_n = \mu^{\ell}$ for every $n$.

For stationary MFG equilibria, \cite{guo2019learningneurips} introduced a best-response based iterative method with tabular Q-learning to compute the best response at each iteration. Moreover, they proved convergence using bounds on classical Q-learning, combined with a strict contraction argument.
\citet{guo2023general} generalized this idea  using a policy gradient approach in lieu of Q-learning.
A similar algorithm but combined with fitted Q-learning instead of tabular Q-learning was analyzed and proved to converge by~\citet{anahtarci2019fittedqmfg}. Furthermore, \citet{anahtarci2020qregmfg,anahtarci2021learningfittedaverage} showed that  some of the conditions to obtain convergence in the tabular Q-learning case can be relaxed if the MDP is regularized.

\citet{subramanian2019reinforcement,angiulifouquelauriere2022unified} used a two-timescale approach combined with model-free RL to compute stationary equilibria. The convergence has been proved under suitable conditions on the underlying ODEs by using stochastic approximation techniques~\citep{borkar2009stochasticapproxbook}.

In the $\gamma$-discounted setting, \cite{elie2020convergencemodelfreemfg} analyzed the propagation of error in Fictitious Play (i.e., how errors made in the computation of the best response propagate through the algorithm) and implemented this scheme with an actor-critic DRL method (namely, DDPG~\citep{lillicrap2016continuous}) to compute the best response. Fictitious play and DRL combined with neural network approximation of the population distribution allowed \citet{perrin2021mfgflockrl}  to solve a flocking model with continuous and high-dimensional space.  Still in the $\gamma$-discounted setting, \citet{perrin2020fictitious} provided a convergence rate for continuous-time Fictitious Play under monotonicity assumption, while \citet{geist2022concave} established a rate of convergence for discrete-time Fictitious Play in MFGs with a potential structure. \cite{mao2022mean} applied a natural actor-critic method with neural networks to a cloud resource management problem. \cite{angiuli2023deepcontinuousspaces} proposed an actor-critic deep RL method to solve MFGs and MFC problems in continuous spaces, building upon a multi-timescale approach introduced in~\cite{angiulifouquelauriere2022unified}. \cite{algumaei2023regularization} proposed  PPO-based algorithm and applied it to solve grid-world examples.

\paragraph{\bf Finite horizon MFG setting. } To learn finite horizon MFG solutions in a model-free way,  we assume that a representative agent is encoded by a time-dependent policy $\bspi = (\pi_n)_{n=0,\dots,N_T-1}$ and can interact with the environment to realize episodes. Each episode is done in the following way: the environment picks $x_0 \sim m_0$ and reveals it to the agent; then for $n=0,\dots,N_T$, the agent observes $x_n$, chooses action $a_n \sim \pi_n(\cdot|x_n)$, and the environment returns a realization of $x_{n+1} \sim p_n(\cdot|x_n,a_n,\mu_n)$ as well as the value of $r_n(x_n,a_n,\mu_n)$. Note that the agent does not need to observe directly the mean field sequence $(\mu_n)_{n=0,\dots,N_T}$, which simply enters as a parameter of the transition and reward functions $p_n$ and $r_n$. 

Based on such episodes, the agent can for example estimate a policy $\bspi$ by approximately computing the state-action value function $\bfQ^{\bspi,\bsmu}$, or compute a best response by first approximating the optimal value function $\bfQ^{*,\mu}$. The value functions can be estimated by backward induction as in \eqref{eq:mfg-finite-horizon-Bellman} and \eqref{eq:mfg-finite-horizon-opt-Bellman}, replacing the expectation by empirical averages over Monte Carlo samples. 

\citet{perrin2020fictitious} solved finite-horizon MFG by a Fictitious Play method in which the best responses are computing using tabular Q-learning. \cite{mishra2020model} proposed a combination of RL and backward induction to solve finite-horizon MFGs by approximating the policy starting from the terminal time. \cite{cui2021approximately} applied best-response based and policy-evaluation based methods combined with DRL techniques and studied numerically the impact of entropy regularization on the convergence of these methods. Although DRL methods offer many promises in terms of scalability, it is in general hard to average or sum non-linear function approximators such as neural networks.  \cite{lauriere2022scalable} proposed best-response based and policy-evaluation based methods (namely Fictitious Play and OMD) with DRL techniques in such ways that average or sum of neural networks can be approximated efficiently. This leads to scalable model-free methods for finite-horizon MFGs. Extensions of these methods to master (i.e., population-dependent) policies have been studied in~\cite{wu2024population}.

\subsubsection{Some remarks about the distribution}

\paragraph{Observing the mean field. }
In the above presentation, we assume that the agent does not observe the distribution, or at least does not exploit this information to learn the equilibrium policy. Although this is the most common approach in the RL and MFGs literature, the question of learning \defi{population-dependent policies} arises quite naturally since one could expect that agents learn how to react to the current distribution they observe. This is usual in MARL, see e.g.~\citet{yang2018mean} who consider Q-functions  depending on the actions of all the other players. In MFGs, we can expect that, by learning a population-dependent policy, the agent will be able to \emph{generalize}, i.e., to behave (approximately) optimally even for population configurations that have not been encountered during training. 

The concept of a value function depending on the population distribution is connected to the so-called \defi{Master equation} in MFGs. Introduced by~\cite{lionsCDF} in continuous MFGs (continuous time, continuous state and action spaces), this partial differential equation (PDE) corresponds to the limit of systems of Hamilton-Jacobi-Bellman PDEs characterizing Nash equilibria in symmetric $N$-player games. We refer the interested reader to e.g.~\citet{BENSOUSSAN20151441,cardaliaguet2019master} for more details on this topic.

With this approach, value functions and policies take as input a distribution, which is a high-dimensional object. As a consequence, they are much more challenging to approximate than population-independent policies. 
\cite{perrin2022generalization}, introduced the concept of \defi{master policies}, which are population-dependent policies allowing to recover an equilibrium policy for any observed population distribution. They proposed to approximately compute master policies by a combination of Fictitious play, DRL, and a suitable randomization of the initial distribution. \cite{wu2024population} extended the approach to non-stationary master policies, and proposed an adaptation of the Munchausen OMD algorithm introduced in~\citep{lauriere2022scalable} to compute policies taking the distribution as an input.

\paragraph{Distribution estimation. }
When the state space is finite but very large, storing the population distribution in a tabular way for every state and computing the evolution of this distribution in an exact way is prohibitive in terms of memory and computational time. In continuous spaces, representing and updating the distribution is even more challenging, even if it is just for the purpose of implementing the RL environment and not to use it as an input to the policies. In this case, one needs to rely on approximations. As already mentioned above, a possible method consists in using an empirical distribution, whose evolution can be implemented by Monte Carlo samples of an interacting agent system. This amounts to using a finite population of agents to simulate the environment. For example, in linear-quadratic MFGs the interactions are only through the mean, which can be estimated even using a single agent, see~\citep{angiulifouquelauriere2022unified,angiuli2022reinforcementmixedmfcg} in the stationary setting and~\citep{angiulifouquelauriere2021reinforcementhandbook,uz2020reinforcementnonstatio,miehling2022reinforcementmultipop,zaman2023oracle} in the finite-horizon setting.  However, it should be noted that even if a finite number of agents is used in the environment, this approach does not directly reduce the problem to a MARL problem because the goal is still to learn the equilibrium policy for the MFG instead of the finite-agent equilibrium policy.

Another approach consists in representing efficiently the distribution using function approximation. This raises the questions of the choice of parameterization and of the training method for the parameters. This approach can be implemented in a model-free way using Monte Carlo samples, which is particularly suitable for spaces that are too large to be explored in an exhaustive fashion. For example, \citet{perrin2021mfgflockrl} used a kind of deep neural networks, namely normalizing flows \citep{rezende2015variational,papamakarios2021normalizing}, to represent the distribution of agents in a flocking model.

\subsection{Reinforcement learning for Mean Field Control and MFMDP}
\label{sec:rl-algorithms-mfc}

As discussed in Section~\ref{sec:iterative-methods-mfc}, the problem of maximizing the social reward can be interpreted in the MDP framework through a mean-field MDP in which the state incorporates the whole mean field state. Adapting in a straightforward way the RL framework represented in Fig.~\ref{fig:rl-env-classical}, we can consider the environment described in Fig.~\ref{fig:mfc-env} where the state is $\mu \in \overline{\states} = \Delta_\states$ instead of $x$, and the reward and the transition are given respectively by $\bar r$ and $\bar p$. An action, taken by the central planner or collectively by the population, is an element of $\overline{\actions} = (\Delta_\actions)^{\states}$. 

The problem can be interpreted as a situation in which all the agents in the population cooperate to learn a socially optimal behavior. Alternatively, we can adopt the point of view of a central planner trying to find an optimal policy that leads to a social optimum if it is followed by all the agents. In both cases, we assume here that the agent who is learning observes the whole population distribution (which is sometimes referred to as the centralized setting). The value function and the policy can thus depend on the state of the mean field, which is consistent with the dynamic programming equations presented in Section~\ref{sec:iterative-methods-mfc}. 

\begin{figure}
    \centering
    \resizebox{0.5\textwidth}{!}{%
    \begingroup
        \tikzset{every picture/.style={scale=0.3}}%
        \tikzset{every picture/.style={line width=0.75pt}} %

\begin{tikzpicture}[x=0.75pt,y=0.75pt,yscale=-1,xscale=1]

\draw   (220,227.2) .. controls (220,222.67) and (223.67,219) .. (228.2,219) -- (321.8,219) .. controls (326.33,219) and (330,222.67) .. (330,227.2) -- (330,251.8) .. controls (330,256.33) and (326.33,260) .. (321.8,260) -- (228.2,260) .. controls (223.67,260) and (220,256.33) .. (220,251.8) -- cycle ;
\draw    (220,250) -- (122,250) ;
\draw [shift={(120,250)}, rotate = 360] [color={rgb, 255:red, 0; green, 0; blue, 0 }  ][line width=0.75]    (10.93,-3.29) .. controls (6.95,-1.4) and (3.31,-0.3) .. (0,0) .. controls (3.31,0.3) and (6.95,1.4) .. (10.93,3.29)   ;
\draw    (120,250) -- (120,152) ;
\draw [shift={(120,150)}, rotate = 450] [color={rgb, 255:red, 0; green, 0; blue, 0 }  ][line width=0.75]    (10.93,-3.29) .. controls (6.95,-1.4) and (3.31,-0.3) .. (0,0) .. controls (3.31,0.3) and (6.95,1.4) .. (10.93,3.29)   ;
\draw   (225,44) .. controls (225,39.03) and (229.03,35) .. (234,35) -- (311,35) .. controls (315.97,35) and (320,39.03) .. (320,44) -- (320,71) .. controls (320,75.97) and (315.97,80) .. (311,80) -- (234,80) .. controls (229.03,80) and (225,75.97) .. (225,71) -- cycle ;
\draw    (120,50) -- (223,50) ;
\draw [shift={(225,50)}, rotate = 180] [color={rgb, 255:red, 0; green, 0; blue, 0 }  ][line width=0.75]    (10.93,-3.29) .. controls (6.95,-1.4) and (3.31,-0.3) .. (0,0) .. controls (3.31,0.3) and (6.95,1.4) .. (10.93,3.29)   ;
\draw    (220,230) -- (152,230) ;
\draw [shift={(150,230)}, rotate = 360] [color={rgb, 255:red, 0; green, 0; blue, 0 }  ][line width=0.75]    (10.93,-3.29) .. controls (6.95,-1.4) and (3.31,-0.3) .. (0,0) .. controls (3.31,0.3) and (6.95,1.4) .. (10.93,3.29)   ;
\draw    (150,230) -- (150,152) ;
\draw [shift={(150,150)}, rotate = 450] [color={rgb, 255:red, 0; green, 0; blue, 0 }  ][line width=0.75]    (10.93,-3.29) .. controls (6.95,-1.4) and (3.31,-0.3) .. (0,0) .. controls (3.31,0.3) and (6.95,1.4) .. (10.93,3.29)   ;
\draw    (150,70) -- (223,70) ;
\draw [shift={(225,70)}, rotate = 180] [color={rgb, 255:red, 0; green, 0; blue, 0 }  ][line width=0.75]    (10.93,-3.29) .. controls (6.95,-1.4) and (3.31,-0.3) .. (0,0) .. controls (3.31,0.3) and (6.95,1.4) .. (10.93,3.29)   ;
\draw    (430,249.8) -- (430,50) ;
\draw [shift={(430,251.8)}, rotate = 270] [color={rgb, 255:red, 0; green, 0; blue, 0 }  ][line width=0.75]    (10.93,-3.29) .. controls (6.95,-1.4) and (3.31,-0.3) .. (0,0) .. controls (3.31,0.3) and (6.95,1.4) .. (10.93,3.29)   ;
\draw    (320,50) -- (428,50) ;
\draw [shift={(430,50)}, rotate = 180] [color={rgb, 255:red, 0; green, 0; blue, 0 }  ][line width=0.75]    (10.93,-3.29) .. controls (6.95,-1.4) and (3.31,-0.3) .. (0,0) .. controls (3.31,0.3) and (6.95,1.4) .. (10.93,3.29)   ;
\draw    (430,251.8) -- (332,251.8) ;
\draw [shift={(330,251.8)}, rotate = 360] [color={rgb, 255:red, 0; green, 0; blue, 0 }  ][line width=0.75]    (10.93,-3.29) .. controls (6.95,-1.4) and (3.31,-0.3) .. (0,0) .. controls (3.31,0.3) and (6.95,1.4) .. (10.93,3.29)   ;
\draw  [dash pattern={on 4.5pt off 4.5pt}]  (70,150) -- (200,150) ;
\draw [color={rgb, 255:red, 0; green, 0; blue, 255 }  ,draw opacity=1 ]   (220,227.2) .. controls (174.46,197.3) and (233.79,164.66) .. (228.38,217.38) ;
\draw [shift={(228.2,219)}, rotate = 277.05] [color={rgb, 255:red, 0; green, 0; blue, 255 }  ,draw opacity=1 ][line width=0.75]    (10.93,-3.29) .. controls (6.95,-1.4) and (3.31,-0.3) .. (0,0) .. controls (3.31,0.3) and (6.95,1.4) .. (10.93,3.29)   ;
\draw    (150,150) -- (150,72) ;
\draw [shift={(150,70)}, rotate = 450] [color={rgb, 255:red, 0; green, 0; blue, 0 }  ][line width=0.75]    (10.93,-3.29) .. controls (6.95,-1.4) and (3.31,-0.3) .. (0,0) .. controls (3.31,0.3) and (6.95,1.4) .. (10.93,3.29)   ;
\draw    (120,150) -- (120,52) ;
\draw [shift={(120,50)}, rotate = 450] [color={rgb, 255:red, 0; green, 0; blue, 0 }  ][line width=0.75]    (10.93,-3.29) .. controls (6.95,-1.4) and (3.31,-0.3) .. (0,0) .. controls (3.31,0.3) and (6.95,1.4) .. (10.93,3.29)   ;

\draw (234,233) node [anchor=north west][inner sep=0.75pt]   [align=left] {Environment};
\draw (239,49) node [anchor=north west][inner sep=0.75pt]   [align=left] {Population};
\draw (74,168) node [anchor=north west][inner sep=0.75pt]   [align=left] {Reward};
\draw (79,191) node [anchor=north west][inner sep=0.75pt]    {$\overline{r}_{n+1}$};
\draw (154,169) node [anchor=north west][inner sep=0.75pt]  [color={rgb, 255:red, 0; green, 0; blue, 255 }  ,opacity=1 ] [align=left] {MF State};
\draw (164,192) node [anchor=north west][inner sep=0.75pt]  [color={rgb, 255:red, 0; green, 0; blue, 255 }  ,opacity=1 ]  {$\mu _{n+1}$};
\draw (433,139) node [anchor=north west][inner sep=0.75pt]   [align=left] {Action};
\draw (447,160) node [anchor=north west][inner sep=0.75pt]    {$\overline{a}_{n}$};
\draw (74,84) node [anchor=north west][inner sep=0.75pt]   [align=left] {Reward};
\draw (79,104) node [anchor=north west][inner sep=0.75pt]    {$\overline{r}_{n}$};
\draw (155,85) node [anchor=north west][inner sep=0.75pt]  [color={rgb, 255:red, 0; green, 0; blue, 255 }  ,opacity=1 ] [align=left] {MF State};
\draw (165,106) node [anchor=north west][inner sep=0.75pt]  [color={rgb, 255:red, 0; green, 0; blue, 255 }  ,opacity=1 ]  {$\mu _{n}$};

\end{tikzpicture}
    \endgroup
    }
\caption{Environment for MFC and MFMDP.}
\label{fig:mfc-env}
\end{figure}
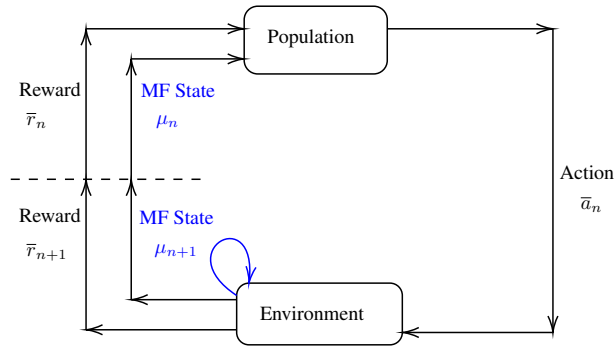

From here, standard RL techniques can be adapted to solve an MFMDP. In their implementation, the main challenge is the representation of the population distribution. A few noticeable cases are the following:
\begin{itemize}
    \item In continuous space linear-quadratic case, the interaction is only through the mean so we do not need to give the full distribution as an input to the policy but only its first moment. In this case, policy gradient for the parameters of a suitable representation of the policy can be implemented and shown to converge, \citep{carmonalaurieretan2019linearpg,wang2021globallqmfcg,gu2020qlearningmfc,gu2024meanmarldecentralized}. This approach can also be extended to more complex settings such as mean-field type games~\citep{carmona2020policyzsmftg,carmonahamidouchelaurieretan2021linearjdg}.
    \item When the state space $\states$ is finite, the mean field state can be represented as an element of the simplex, identified as a subset of $\RR^{|\states|}$:  $\{\mu \in [0,1]^{|\states|}\,:\, \sum_{i=1}^{|\states|} \mu_i = 1\}$. We can then use two different approaches. 
    \begin{itemize}
        \item First, this simplex can be discretized and replaced by a finite set $\tilde{\states} \subset \overline{\states}$. We can then approximate the MFMDP by an MDP with this finite state space. The action space is in principle $\Delta_{\tilde{\states}}$, which is continuous, but if we are also willing to discretize this space, then we obtain a finite state space, finite action space MDP for which tabular RL methods can be used. For example tabular Q-learning can be shown to converge under suitable conditions~\citep{carmona2023model,gu2020qlearningmfc}.
        \item 
        \looseness=-1
        Alternatively, the original MFMDP can be tackled without space discretization by using RL techniques for continuous space MDPs. For example, ~\citet{carmona2023model} used deep RL  to learn optimal policies as functions of the population distribution viewed as an element of $\{\mu \in [0,1]^{|\states|}\,:\, \sum_{i=1}^{|\states|} \mu_i = 1\}$. 
    \end{itemize}
\end{itemize}

It can be argued that the environment described in Fig.~\ref{fig:mfc-env} is not very realistic because in general, we cannot assume that an agent observes the mean field distribution. Indeed, this distribution corresponds to the regime with an infinite number of agents while in practice, the number of agents is always finite. One can thus replace the ``ideal'' \defi{McKean-Vlasov environment} by a more realistic \defi{finite-population environment}. The former can be viewed as an approximation of the latter. The quality of the approximation gets better as the number of agents $N$ in the environment increases. These two types of environments are discussed e.g. by~\citet{carmonalaurieretan2019linearpg}, in which the finite-population environment analysis benefits from the analysis of the McKean-Vlasov environment. The connection between RL for MFC and finite-agent problems has also been analyzed by~\citet{wang2020breaking,chen2021pessimism,li2021permutationmarl}. Furthermore, RL for continuous-time MFC has been investigated in~\cite{frikha2023actor}. Extensions to non-homogeneous populations have been considered in~\cite{mondal2022approximationheterogeneous,cui2023learningdecpo,cui2023multi,hu2023graphon}.

Lastly, as in the MFG setting, for some MFC problems, it has been shown that observing the state of a single agent is sufficient to approximate the mean field distribution and learn the optimal behavior, see~\citet{angiulifouquelauriere2022unified,angiulifouquelauriere2021reinforcementhandbook,angiuli2023convergence}.

\section{Numerical experiments}\label{sec:numerical-experiments}

We now present numerical experiments to illustrate some of the techniques introduced in the previous sections. We first discuss metrics that can be used to assess convergence to Nash equilibrium, and several algorithms based on the material presented in the previous sections. We then present an MFG model and numerical results. The experiments are implemented using the open-source library \texttt{OpenSpiel}~\citep{lanctot2019openspiel}.\footnote{\url{https://github.com/deepmind/open_spiel}}

\subsection{Metrics}
\label{sec:metrics}
Here we discuss ways to measure convergence of the iterative methods discussed in Section~\ref{sec:iterative-methods}. First, since many methods are based on fixed point iterations, we can measure the distance between successive mean field terms. Second, we can also measure convergence in terms of the policy's exploitability.

\paragraph{Wasserstein distance. }
Let us recall that the iterative methods described previously are based on the scheme summarized in~\eqref{eq:iterative-methods-scheme}. The pair $(\mu^\ell, \pi^\ell)$ computed at iteration $\ell$ is expected to converge to a Nash equilibrium. If the Nash equilibrium $\hat{\mu}$ is known (which is the case in some examples of the literature which are used to benchmark new methods), we can measure the distance between the mean field $\mu^{\ell}$ at the current iteration and the Nash equilibrium. However, in general $\hat\mu$ is not known. But if the iterations converge, then the distance between $\mu^{\ell}$ and $\mu^{\ell+1}$ should decay to $0$.\footnote{Depending on the way the policy is updated, the distance between $\pi^{\ell}$ and $\pi^{\ell+1}$ may also decay to $0$ but it is not always the case. For instance in best-response based iterations, there could be multiple optimal policies, and successive iterations may compute different best responses, even after the mean field term has converged.} 
Since the mean field is a probability distribution, we can use for instance the Wasserstein distance, which has the advantage to be well-defined even when the underlying space is continuous. 

Let us focus on the mean field and look at the macroscopic behavior, at the scale of the whole population (one can proceed similarly with the policy). For simplicity, let us assume the state space $\states$ is a finite set endowed with a distance denoted by $d$. The \defi{Wasserstein distance} $\cW$ (or earth mover's distance) measures the minimum cost of turning one distribution into another and is defined as follows: for $\smf,\smf' \in \Delta_\states$, 
$
    \cW(\smf,\smf')
    = \inf_{\nu \in \Gamma(\smf,\smf')} \sum_{(x,x') \in \states \times \states} d(x, x') \nu(x,x'),
$
where $\Gamma(\smf,\smf')$ is the set of probability distributions on $\states\times\states$ with marginals $\smf$ and $\smf'$.

In a finite-horizon setting with horizon $N_T$, the mean field is a sequence of distributions. We can average the distances over the $N_T +1$ time steps to get the following distance between mean field sequences: for $\mf,\mf' \in (\Delta_\states)^{N_T+1}$, 
$$
	\cW_T(\mf,\mf')
	= \frac{1}{N_T+1}\sum_{n=0}^{N_T} \cW(\mf_n, \mf'_n).
$$

This distance can be used measure the distance between two successive iterates of the scheme~\eqref{eq:iterative-methods-scheme}, namely, $\cW(\mu^{\ell}, \mu^{\ell+1})$. Although there is in general no guarantee that this distance should decrease monotonically, it goes to zero if the method converges.

\paragraph{Exploitability. }
Although the idea of computing the distance between successive iterates (of mean fields, policies, or value functions) is very natural, there is a major drawback: even if the distance goes to $0$, it does not prove that the algorithm has converged to a Nash equilibrium. In fact, in some methods such as Fictitious Play, the updates are becoming smaller and smaller as iterations go by, so the method discussed in the previous subsection does not provide sufficient information to assess the convergence to a Nash equilibrium of the MFG. 

Instead, one can check whether the policy $\pi^{\ell}$ at the current iteration is approximately a Nash equilibrium by using the reward function.
Indeed, the definition of an approximate Nash equilibrium can be formalized by  measuring to what extent a representative player can improve their reward by deviating from the policy used by the rest of the population.  This leads to the notion of exploitability. The definition in MFG is inspired by analogous concepts introduced in the context of computational game theory~\citep{zinkevich2007regret,lanctot2009monte}.

To fix the ideas, let us consider the finite horizon evolutive setting for instance. The exploitability of a policy $\pi$ is defined as follows:
\begin{definition}
The \defi{exploitability} of policy $\bspi$ is defined as:
$
	\mathcal{E}(\bspi) = \sup_{\bspi'} J_{\mathrm{evol}}(\bspi'; \bsmu^{\bspi}) -  J_{\mathrm{evol}}(\bspi; \bsmu^{\bspi}),
$
where $\bsmu^{\bspi} = \MF_{\mathrm{evol},m_0,N_T}(\bspi)$ is the mean field distribution sequence induced by $\bspi$ as defined in~\eqref{eq:evol-mu}, and $J_{\mathrm{evol}}$ is defined in~\eqref{eq:evol-J}. 
\end{definition}
Notice that, in contrast with the definition of the total reward $J_{\mathrm{evol}}(\bspi;\bsmu)$, which assumes a given mean field $\bsmu$, the exploitability $\mathcal{E}(\bspi)$ depends only on the policy under consideration. Furthermore, computing the exploitability does not require finding the Nash equilibrium, but only computing one optimal policy against the mean field induced by the given policy. 

Exploitability can be defined in a similar way for other settings besides the evolutive one. Furthermore, besides its use to assess numerical convergence, exploitability has also been used to prove the convergence of several algorithms such as fictitious play or online mirror descent. For more details, we refer for instance to~\citep{perrin2020fictitious,geist2022concave} for the $\gamma$-discounted setting, \citep{perrin2020fictitious,cui2021approximately,cheng2023meanregret,ramponi2024imitation} for the finite horizon setting, and \citep{perrin2022generalization} for the stationary setting.

Using this notion, we can rephrase the definition of mean field Nash equilibrium (see Definition~\ref{def:statioMFNE}) as: $\hat\pi$ is a stationary MFNE policy if and only if:
$
	\mathcal{E}(\hat\bspi) = 0.
$
In fact, $\mathcal{E}$ is always non-negative, and for $\epsilon \ge 0$,
$
	\mathcal{E}(\bspi) \le \epsilon
$
corresponds to saying that the policy $\bspi$ is an \defi{$\epsilon$-Nash equilibrium}, meaning that a representative player can improve her reward by at most $\epsilon$ by unilaterally deviating from the policy $\bspi$ used by the rest of the population. As such, the exploitability offers a different perspective than the Wasserstein distance discussed above to assess the convergence of learning methods. 
In the context of iterations described by~\eqref{eq:iterative-methods-scheme}, the quantity $\mathcal{E}(\bspi^\ell)$ measures convergence from the point of view of the potential reward improvement that a player could get by deviating from $\bspi^\ell$ while the rest of the population would use $\bspi^\ell$. 

In practice, computing the exploitability of a policy $\bspi$ necessitates computing  $\sup_{\bspi'} J_{\mathrm{evol}}(\bspi'; \bsmu^\bspi)$. This requires solving the MDP for a representative player facing the mean field $\bsmu^\bspi$. For many problems, there is no explicit formula for this value and if the environment is complex, exact methods cannot be used. However an approximation can be computed by learning an approximate best response to $\bsmu^\bspi$, for example using RL. If this step is still too computationally expensive, then we can replace the  supremum over {\it all} policies by a maximum over a {\it finite set}, say $\widetilde\Pi$. For example, for this set, we can use the set of policies computed in previous iterations. This amounts to check how much a player could be better off by using one of the past policies instead of the current one. This leads to the notion of \defi{approximate exploitability} used in~\citep{perrin2021mfgflockrl,perrin2022generalization} and defined as:
$$
	\widetilde{\mathcal{E}}(\bspi) = \max_{\pi' \in \widetilde\Pi} J_{\mathrm{evol}}(\bspi'; \bsmu^\bspi) -  J_{\mathrm{evol}}(\bspi; \bsmu^\bspi).
$$

We conclude this presentation of metrics by mentioning that in the case of MFC, the convergence can be assessed through the social cost, see Section~\ref{sec:setting-soc-opt}. For numerical illustrations involving deep RL and using the social cost (or reward) as a measure of success, we refer to e.g.~\citep{carmona2023model}.

\subsection{Algorithms} 
\label{sec:num-algo}

For the sake of illustration, we consider below the following iterative methods. 
\begin{itemize}
    \item \textbf{Fixed point: } $\bspi^{\ell+1}$ is a best response against $\bsmu^{\ell}$ and $\bsmu^{\ell+1}$ is the mean field induced by $\bspi^{\ell+1}$.
    \item \textbf{Fictitious play: } As described in~\eqref{eq:mfg-statio-BRbased-fictitiousplay}, we update the mean field by averaging over past iterations. 
    \item \textbf{Online Mirror Descent: } As described in~\eqref{eq:mfg-statio-policy-eval-omd}, the policy is updated by first computing a cumulative $Q$-function and then taking a softmax. 
    \item \textbf{Damped Fixed Point: } As described in~\eqref{eq:damped-fixedpoint-iterations}, the mean field term is updated by taking the average of the previous mean field and the mean field induced by the most recent policy, with constant weights for the average. 
    \item \textbf{Softmax fixed point: } This method is like the fixed point iterations except that the policy is a softmax of the $Q$-function instead of being an argmax as in the pure best response case.  
    \item \textbf{Softmax Fictitious Play: } This method is like the Fictitious Play iterations except that the policy is a softmax of the $Q$-function instead of being an argmax as in the pure Fictitious Play case.
    \item \textbf{Boltzmann policy iteration: }  This method corresponds to the policy iteration method described in~\eqref{eq:finiteHmfg-policy-iteration-method} except that the policy is a softmax of the $Q$-function instead of being an argmax as in the pure Fictitious Play case. We stick to this terminology, which has been used in the literature, although this method could also be called \textbf{softmax policy iteration} to be consistent with the terminology used for the other methods. 
\end{itemize}
In the numerical results below, the computations are done without RL but these methods can be combined with RL, see Appendix~\ref{sec:expe-RL} of the long version~\citep{lauriere2022learning} for some examples.

\subsection{MFG Model} 
For the sake of illustration, we focus on a model that is in the finite-horizon evolutive setting, in which the state space represents a grid world and the actions correspond to movements. The mean field through which the interactions occur is the density of the population over the grid world. The reward function is typically composed of three additive terms:
$$
    r(x,a,\mu) = r_{\mathrm{pos}}(x) + r_{\mathrm{move}}(a,\mu(x)) + r_{\mathrm{pop}}(\mu(x)),
$$
where $\mu(x)$ is the value of the distribution at state $x$, hence the dependence on the distribution is local. The first term will encode spatial preferences, the second term will represent a cost of moving (which may depend on the distribution), and the third term will capture mean field interactions, independently of any movement.  The model we present here corresponds to a crowd evolving in a maze. More examples are presented in Appendix~\ref{app:more-examples-num} of the long version~\citep{lauriere2022learning}.

\paragraph{A Model for crowd motion with congestion in maze. }

We consider the following model, which is a finite horizon problem (see Section~\ref{sec:evol-mfg-setting}) and has been introduced in~\citep{lauriere2022scalable}:
\begin{itemize}
    \item {\bf State space: } $\states = \{0,\dots,N_x^1\} \times \{0,\dots,N_x^2\}$, which represents a 2D grid world.
    \item {\bf Action space: } $\actions = \{(-1,0), (1,0), (0,0), (0,1), (0,-1)\}$, which represents movement in the 4 directions and the absence of movement. They correspond respectively to: left, right, stay, up, down. 
    \item {\bf Transitions: } At time $n$, the agent chooses to stay at the current position or to move to one of the neighboring positions. There are walls which form a maze, see~Figure~\ref{fig:expe-maze-distributions} where the walls are represented in grey color. Furthermore, a random disturbance potentially affects the dynamics. The next state is computed according to the dynamics:
    $$
        x_{n+1} = 
        \begin{cases}
            x_{n} + a_{n} + \epsilon_{n+1}, \quad &\hbox{ if } x_{n} + a_{n} + \epsilon_{n+1} \hbox{ is not in a forbidden state,} 
            \\
            x_{n}, \quad &\hbox{otherwise,} 
        \end{cases}
    $$
    where $(\epsilon_{n})_n$ is a sequence of i.i.d. random variables taking values in $\actions$. In the experiment, we take $\epsilon_{n} = (0,0)$ with probability $1-p$ where $p \in [0,1]$ and otherwise $\epsilon_{n}$ is one of the other four actions, each having probability $p/4$. We refer to $p$ as the noise intensity.
    \item {\bf Time horizon: } There is no discount and time goes until a terminal time horizon $N_T$.
    \item {\bf Rewards: } The players pay a cost (negative reward) to move, and this cost increases when the density of agents increases. There is also an incentive to reach the target position. The reward function is given by:
    $
        r(x,a,\mu) = c_{\mathrm{pos}} r_{\mathrm{pos}}(x) + r_{\mathrm{move}}(a,\mu(x)), 
    $
    where 
    \begin{itemize}
        \item[$\bullet$] $r_{\mathrm{pos}}(x) = - c_{\mathrm{pos}} \mathrm{dist}(x, x_{\mathrm{ref}})$ is the distance to the target position $x_{\mathrm{ref}}$; $c_{\mathrm{pos}}$ is a positive coefficient.
        \item[$\bullet$] $r_{\mathrm{move}}(a,\mu(x)) = -\mu(x)\|a\|$ is a penalty for moving ($\|a\|=1$) which increases with the density $\mu(x)$ at $x$; this is called congestion effect in the literature.
    \end{itemize}  
    The target position is in the bottom right corner of the maze.
    \item {\bf Initial distribution: }  The initial distribution is concentrated in the top left corner.
\end{itemize}

\paragraph{Implementation in OpenSpiel. } The games and the algorithms are implemented in \texttt{OpenSpiel}~\citep{lanctot2019openspiel} and are publicly available, along with more examples and algorithms. \texttt{OpenSpiel} is a framework for learning in games, including with reinforcement learning. It encompasses a wide variety of games besides MFGs; see \url{https://github.com/deepmind/open_spiel/}. More details on MFGs in \texttt{OpenSpiel} are provided in Appendix~\ref{app:intro-mfg-openspiel} of the long version~\citep{lauriere2022learning}.

\paragraph{Numerical setup. }
In the experiments below, we use the following parameters:
    We use $N_x^1 = N_x^2 = 22$, with walls on the sides, and walls inside the domain to form a maze (see Figure~\ref{fig:expe-maze-distributions} below).
    The time horizon is $N_T=70$.
    In the rewards, we use $c_{\mathrm{pos}} = 5$.
    The learning rate is $0.05$ for Online Mirror Descent, $0.01$ for Damped Fixed Point, and the temperature is $0.1$ for softmax in Softmax Fixed Point, Softmax Fictitious Play and Boltzmann Policy Iteration.

\paragraph{Numerical results. }

Figure~\ref{fig:expe-maze-distributions} displays the distribution evolution generated by using the policy learnt by each algorithm. Rows correspond to different algorithms while columns correspond to different time steps. Figure~\ref{fig:expe-maze-exploitabilities} shows, for each algorithm, the exploitability as a function of the iteration number.

We see that Online Mirror Descent has the fastest convergence. Fictitious Play and Softmax Fictitious Play also perform well, although we can expect the latter to saturate at some point due to the limited approximation capabilities of softmax policies. The exploitability of Damped Fixed Point is decreasing but rather slowly. It is possible that the convergence would be faster with a smaller damping coefficient (i.e., larger learning rate). However, we see that pure fixed point iterations (corresponding to Damped Fixed Point algorithm with a damping of $0$) does not converge. Boltzmann Policy Iteration also fails to converge in this example. See Appendix~\ref{app:more-examples-num} of the long version~\citep{lauriere2022learning} for other examples.

\begin{figure}[tbh!]
    \centering
    \begin{minipage}{.7\linewidth}
    \includegraphics[width=\linewidth]{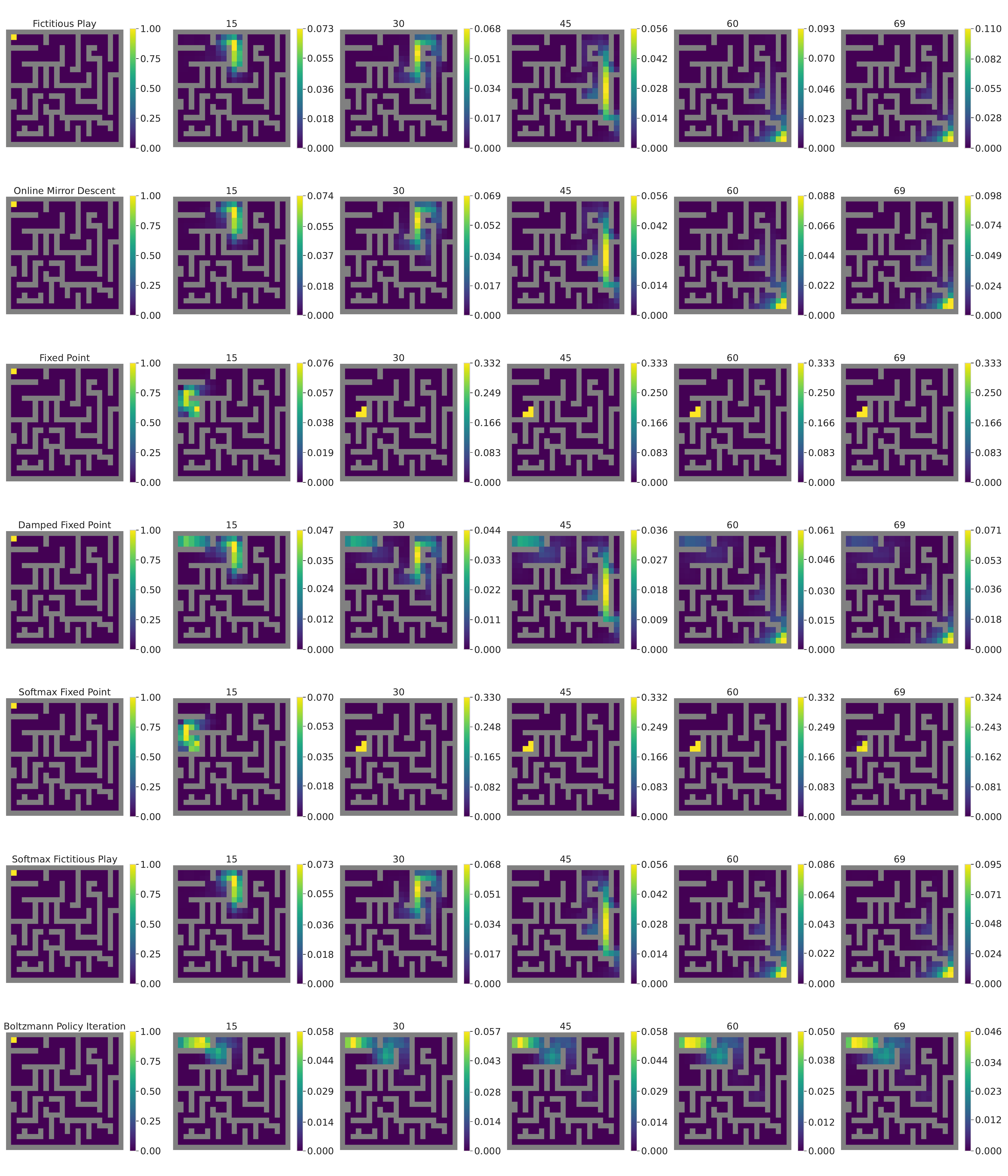}
    \end{minipage}
    
    \caption{Crowd in a maze: Evolution of the distributions for the methods discussed in Section~\ref{sec:num-algo}. Each row corresponds to one method, each column corresponds to one time step in $\{0, 15, 30, 45, 60, 69\}$.}
    \label{fig:expe-maze-distributions}
\end{figure}

\begin{figure}[tbh!]
    \centering
    \begin{minipage}{.7\linewidth}
    \includegraphics[width=\linewidth]{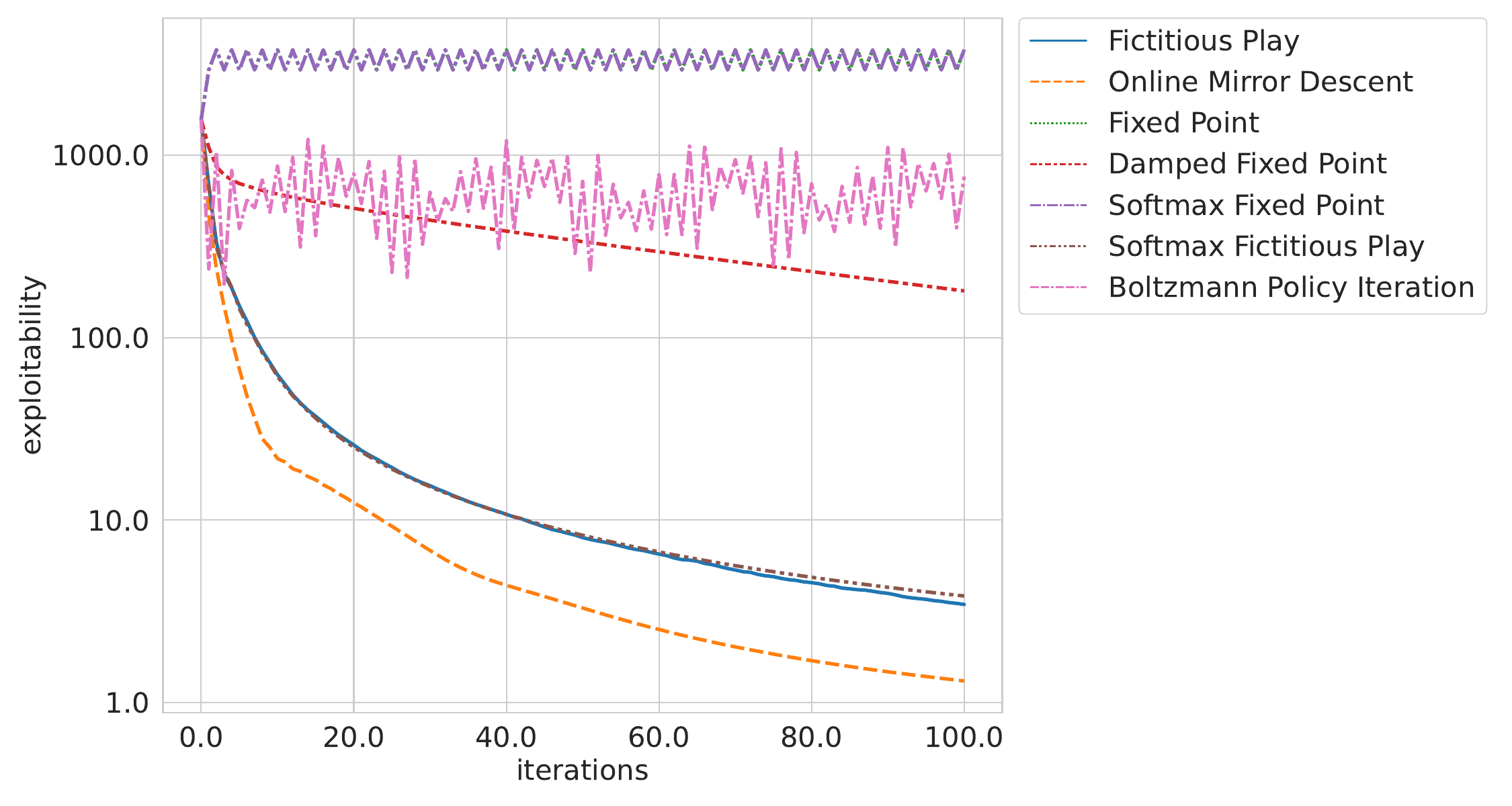}
    \end{minipage}
    
    \caption{Crowd in a maze: Exploitability curves for the methods discussed in Section~\ref{sec:num-algo}.}
    \label{fig:expe-maze-exploitabilities}
\end{figure}

\section{Conclusion and perspectives} 
\label{sec:conclusion}

In this work, we have surveyed the main recent developments related to the question of learning MFGs and MFC solutions. We first clarified the definitions of several classical settings that have appeared in the literature. As far as we know, it is the first time that these settings are summarized and discussed in comparison with each other. Second, we proposed an overview of iterative methods to learn MFG and MFC solutions by updating the mean field and the policy. Starting from simple fixed-point iterations, we explained how these procedures can be enhanced by incorporating various smoothing methods. Along the way, we clarified the link with the framework of MDPs. Third, building on this connection with MDPs, we presented RL and deep RL methods for MFGs and MFC. Finally, we provided some numerical results on a simple benchmark problem. 

Several aspects were not discussed in detail here. Among others, we can cite:
\begin{itemize}
    \item games with multiple groups or sub-populations~\citep{subramanian2018reinforcementmfteams,subramanian2020multi,perolat2022scaling,yang2020bayesianmultitype,uz2023reinforcementlqmulti}, graphon-type interactions~\citep{cui2021learninggraphonmfg,fabian2022mean,cui2022hypergraphon,fabian2023learning,hu2023graphon,zhang2024learning}, mean field control games and mixed equilibria~\citep{angiuli2022reinforcementbank,angiuli2022reinforcementmixedmfcg,angiuli2023convergence,carmona2023nash}, mean field type games~\citep{carmona2020policyzsmftg,carmonahamidouchelaurieretan2021linearjdg,guan2023zero,zaman2024independent}, major-minor MFGs~\citep{cui2023learningdecpo,cui2023learning},
    \item static models and bandit problems~\citep{gummadi2013meanbandit,iyer2014meanauction,maghsudi2017distributedbandit,wang2021meanbandit},
    \item games with strategic complementarities~\citep{adlakha2013meancomplementarities,lee2020learningtremblinghand,lee2021reinforcementcomplementarities},
    \item problems with more complex structures such as partially observable problems~\citep{ganapathi2021partially,cui2023learningdecpo,yang2023partially}, %
    \item other types of equilibria, such as correlated equilibria~\citep{muller2022learningpsro,muller2022learningmfce,wang2022unifieddeepequilibrium},
    \item other forms of RL techniques such as model-based RL~\citep{pasztor2023efficientmodelbased,huang2023statistical,huang2024model} or inverse reinforcement learning~\citep{yang2018learningdeepmfg,chen2021agentinverseRL,chen2022individualinverseRL,ramponi2024imitation,anahtarci2024maximum},
    \item decentralized and independent learners~\cite{yongacoglu2022independent,subramanian2022decentralized,yardim2023policy,benjamin2023networked,yardim2024mean}.
\end{itemize}

Apart from the points covered in this survey, many directions remain to be investigated. For example, the question of approximation of the distribution has received relatively less interest than the question of approximating the policy and the value function. Efficiently representing and learning the distribution is important for mean field problems, particularly for large and complex environments for which exact tabular representations are not suitable. 

Furthermore, the literature is quickly expanding in various directions and the numerical results are not always easy to compare. We hope that this survey will contribute to harmonizing the research on this topic although many aspects remain to be unified. Along these lines, having common benchmark problems and a common framework seem important. Recently, MFGs have been incorporated to the \texttt{OpenSpiel} library~\citep{lanctot2019openspiel}.\footnote{\url{https://github.com/deepmind/open_spiel}} 

Last but not least, one of the main motivations to use RL methods for MFGs is to be able to compute Nash equilibria at a large scale. We thus hope that the methods presented here and their extensions will find concrete applications in the near future.

\section*{Acknowledgements}
Mathieu Lauriere is affiliated with the Shanghai Frontiers Science Center of Artificial Intelligence and Deep Learning, and with the NYU-ECNU Institute of Mathematical Sciences, NYU Shanghai, 567 West Yangsi Road, Shanghai, 200126, People’s Republic of China.
The authors would like to thank their collaborators on the topic of this survey for fruitful discussions and collaborations, but also for contributions developing and maintaining the \texttt{OpenSpiel} code used in this survey, and in particular: %
Andrea Angiuli, Olivier Bachem, Tamer Basar, Theophile Cabannes, Ren\'e Carmona, Kai Cui, G\"ok{\c c}e Dayanikli, Jean-Pierre Fouque, Maximilien Germain, Xin Guo, Niao He, Kenza Hamidouche, Ruimeng Hu, Ayush Jain, Pavel Kolev, Alec Koppel, Raphael Marinier, Rémi Munos, Huy\^en Pham, Georgios Piliouras, Giorgia Ramponi, Mark Rowland, Kai Shao, Zongjun Tan, Karl Tuyls, Muhammad Aneeq uz Zaman, Mengrui Zhang.

\bibliographystyle{apalike}
\bibliography{bib}

\newpage

\appendix

\section{More numerical examples}
\label{app:more-examples-num}

In this section we study numerically several other examples of MFGs. They show that performance of each algorithm varies significantly depending on the problem.

\subsection{Exploration in four rooms}
\label{sec:experiments-four-rooms-exploration}

We consider a game in which there are four connected rooms and the agents want to avoid crowded regions, which leads to exploration of the spatial domain. This model has been introduced in~\citep{geist2022concave}.

\paragraph{Model. }
We consider the following model:
\begin{itemize}
    \item {\bf State space: } $\states = \{0,\dots,N_x^1\} \times \{0,\dots,N_x^2\}$, which represents a 2D grid world.
    \item {\bf Action space: } $\actions = \{(-1,0), (1,0), (0,0), (0,1), (0,-1)\}$, which represents movement in the 4 directions and the absence of movement. They correspond respectively to: left, right, stay, up, down. 
    \item {\bf Transitions: } At time $n$, the agent chooses to stay at the current position or to move to one of the neighboring positions. There are walls which form four rooms connected by doors, see~Figure~\ref{fig:exploration-no-rl-settings}. Furthermore, a random disturbance potentially affects the dynamics. The next state is computed according to the dynamics:
    $$
        x_{n+1} = 
        \begin{cases}
            x_{n} + a_{n} + \epsilon_{n+1}, \quad &\hbox{ if } x_{n} + a_{n} + \epsilon_{n+1} \hbox{ is not in a forbidden state,} 
            \\
            x_{n}, \quad &\hbox{otherwise,} 
        \end{cases}
    $$
    where $(\epsilon_{n})_n$ is a sequence of i.i.d. random variables taking values in $\actions$. In the experiment, we take $\epsilon_{n} = (0,0)$ with probability $1-p$ where $p \in [0,1]$ and otherwise $\epsilon_{n}$ is one of the other four actions, each having probability $p/4$. We refer to $p$ as the noise intensity.
    \item {\bf Time horizon: } The rewards are accumulated without discount and only until a terminal time horizon $N_T$.
    \item {\bf Rewards: } $r(x,a,\mu) = -\log(\mu(x))$, which increases when the distribution at $x$ decreases. This reward gives an incentive to the agents to avoid being in a crowded region. It is sometimes referred to crowd aversion in the literature. Overall, we expect the population to spread as much as possible. In fact, at the macroscopic level of the population, the average one-step reward if the population distribution is $\mu$ is:
    $$
        \EE_{x \sim \mu}[-\log(\mu(x)] = -\sum_{x} \mu(x) \log(\mu(x)),
    $$
    which is the entropy of $\mu$. So maximizing the average reward amounts to maximizing the entropy. This model has been introduced and studied by~\citet{geist2022concave}, in which it is shown that the Nash equilibrium can in fact be interpreted as the social optimum of an MFC problem. 
    \item {\bf Initial distribution: }  The initial distribution is concentrated in the top left corner of the top left room (see Figure~\ref{fig:exploration-no-rl-settings}~{(a)}.
\end{itemize}

\paragraph{Numerical setup. }
In the experiments below, we use the following parameters:
\begin{itemize}
    \item {\bf State space: } We use $N_x^1 = N_x^2 = 13$, with walls on the sides, and walls inside the domain to split it into 4 connected rooms. 
    \item {\bf Time horizon: } $N_T=40$.
    \item {\bf Noise intensity: } $p = 1.0$ (large noise), $p = 0.5$ (medium), $p = 0.1$ (small noise). %
    \item {\bf Hyperparameters: } Unless otherwise specified, the learning rate is $0.05$ for Online Mirror Descent, $0.01$ for Damped Fixed Point, and the temperature is $0.1$ for softmax in Softmax Fixed Point, Softmax Fictitious Play and Boltzmann Policy Iteration. We present below results for various hyperparameters, such as learning rate and temperature of softmax policies. 
\end{itemize}

\paragraph{Results with a uniform policy. }
Since the goal is to spread throughout the domain, we first check the performance of the uniform policy, i.e., the policy which gives the uniform distribution over the set of possible actions. This policy generates the distribution shown in Figure~\ref{fig:exploration-no-rl-settings}~{(b)}; the plot represents the log-density. %

\begin{figure}[tbh]
    \centering
    \begin{minipage}{.3\linewidth}
    \includegraphics[width=.85\linewidth]{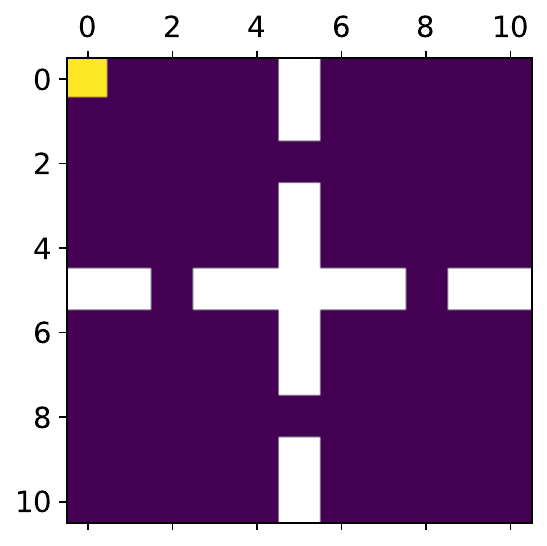}
    \end{minipage}
    \begin{minipage}{.3\linewidth}
    \includegraphics[width=\linewidth]{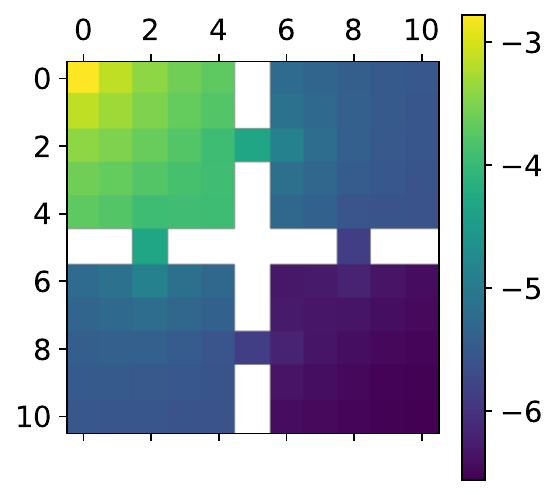}
    \end{minipage}
    \caption{%
    Reading order: \textbf{(a)} Initial distribution concentrated in the yellow state (yellow $ = 1$); the walls are in white; \textbf{(b)} The log-density of the distribution generated by the uniform policy.
    }
    \label{fig:exploration-no-rl-settings}
\end{figure}

\paragraph{Numerical results with fixed hyperparameters. }

We first consider the model with large noise intensity, i.e., $p=1.0$. Figure~\ref{fig:expe-four-room-explo-noise1.0-distributions} displays the distribution evolution generated by using the policy learnt by each algorithm. Rows correspond to different algorithms while columns correspond to different time steps. Figure~\ref{fig:expe-four-room-explo-noise1.0-exploitabilities} shows, for each algorithm, the exploitability as a function of the iteration number.

From these two figures, we can see that Fictitious Play and Online Mirror Descent seem to converge best. On the other hand, (pure) Fixed Point and Softmax Fixed Point methods fail to converge. Between these two extremes, Damped Fixed Point (with a suitable level of damping -- see below for the sweep) seems to be converging. Softmax Fictitious Play's exploitability decays faster than (normal) Fictitious Play's at the beginning, but after a while it stagnates, probably due to the fact that the best response at each iteration is of softmax type, with a fixed temperature, which prevents it from approximating well a true best response.

Figures~\ref{fig:expe-four-room-explo-noise0.5-distributions} and~\ref{fig:expe-four-room-explo-noise0.5-exploitabilities} display analogous results for the model with medium noise intensity, $p=0.5$. 

Finally, Figures~\ref{fig:expe-four-room-explo-noise0.1-distributions} and~\ref{fig:expe-four-room-explo-noise0.1-exploitabilities} display analogous results for the model with small noise intensity, $p=0.1$. 

We see that decreasing the noise intensity makes the problem more difficult, at least for some of the methods. For instance Fictitious Play now converges more slowly than Online Mirror Descent.

\begin{figure}[tbh!]
    \centering
    \begin{minipage}{.7\linewidth}
    \includegraphics[width=\linewidth]{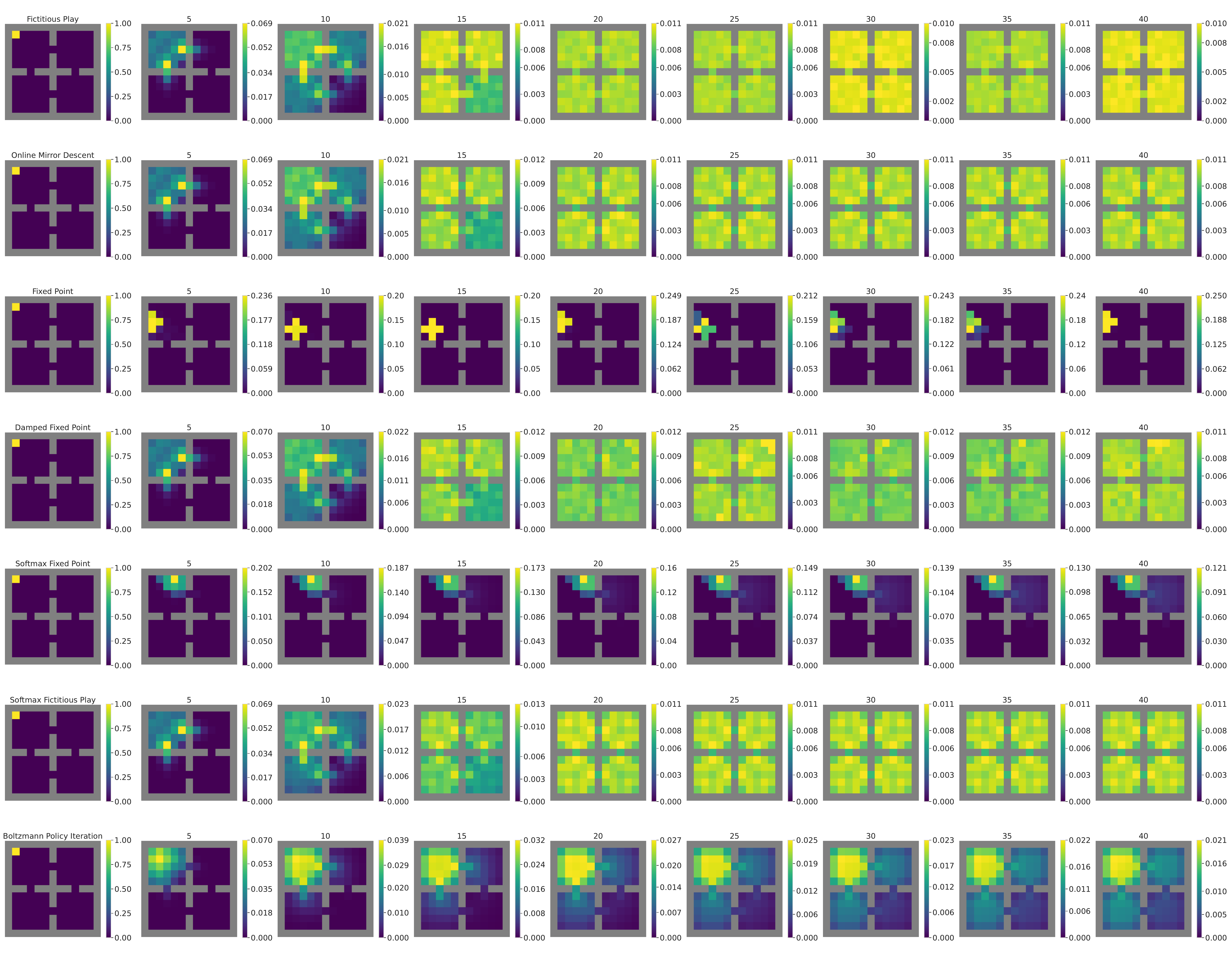}
    \end{minipage}
    
    \caption{Four room exploration with noise intensity $p=1.0$. Distribution evolution.}
    \label{fig:expe-four-room-explo-noise1.0-distributions}
\end{figure}

\begin{figure}[tbh!]
    \centering
    \begin{minipage}{.7\linewidth}
    \includegraphics[width=\linewidth]{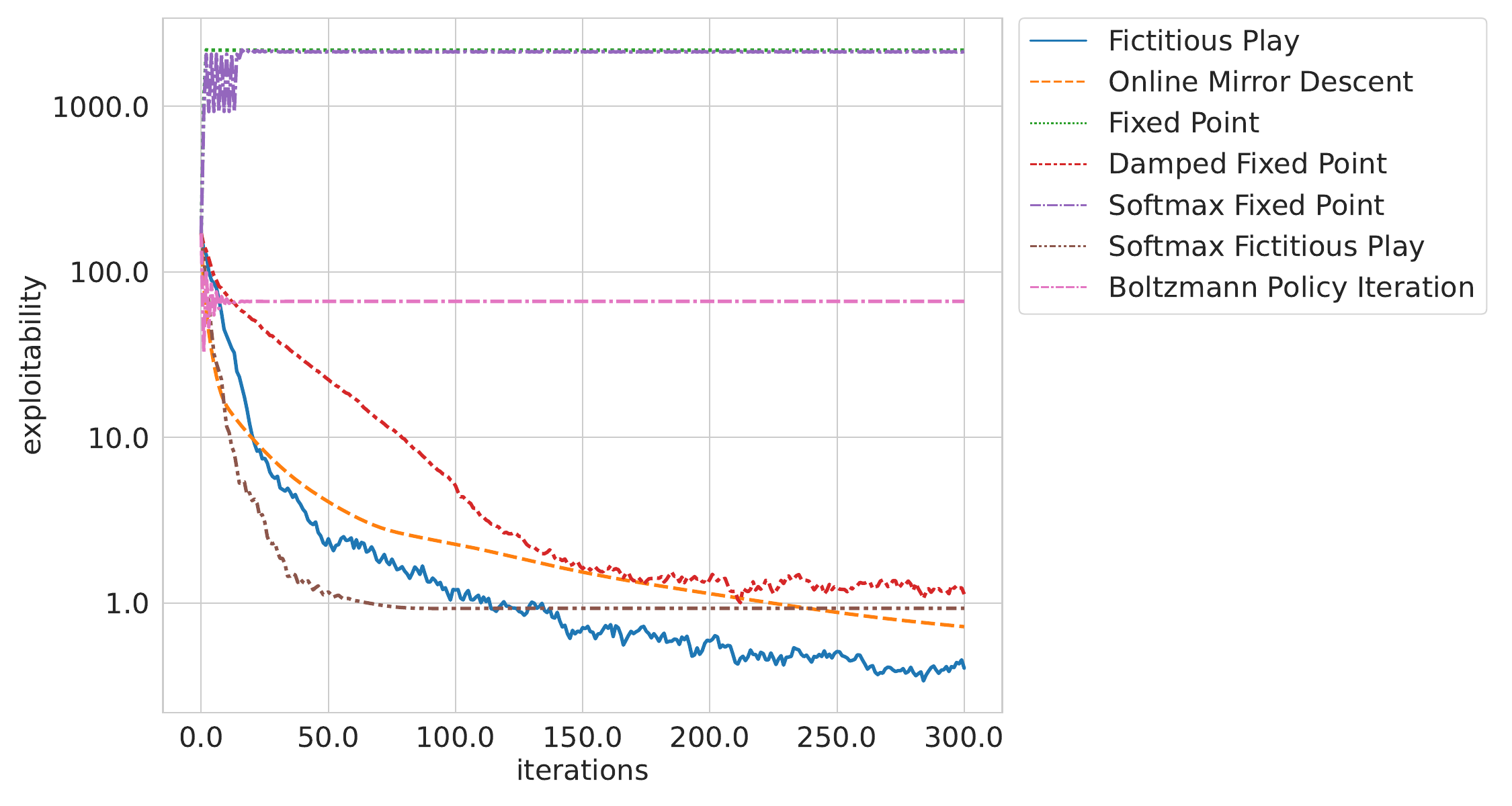}
    \end{minipage}
    
    \caption{Four room exploration with noise intensity $p=1.0$. Exploitability.}
    \label{fig:expe-four-room-explo-noise1.0-exploitabilities}
\end{figure}

\begin{figure}[tbh!]
    \centering
    \begin{minipage}{.7\linewidth}
    \includegraphics[width=\linewidth]{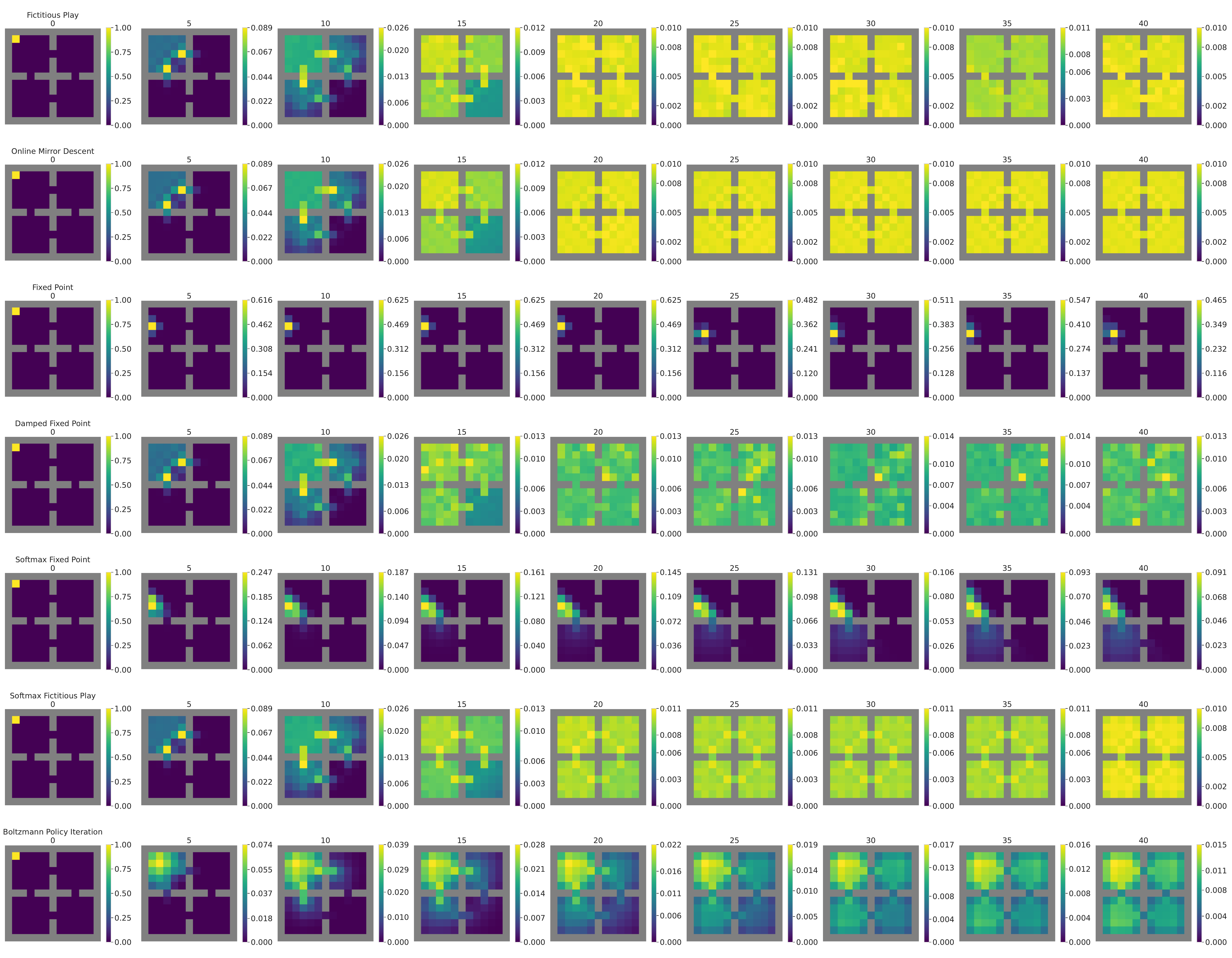}
    \end{minipage}
    
    \caption{Four room exploration with noise intensity $p=0.5$. Distribution evolution.}
    \label{fig:expe-four-room-explo-noise0.5-distributions}
\end{figure}

\begin{figure}[tbh!]
    \centering
    \begin{minipage}{.7\linewidth}
    \includegraphics[width=\linewidth]{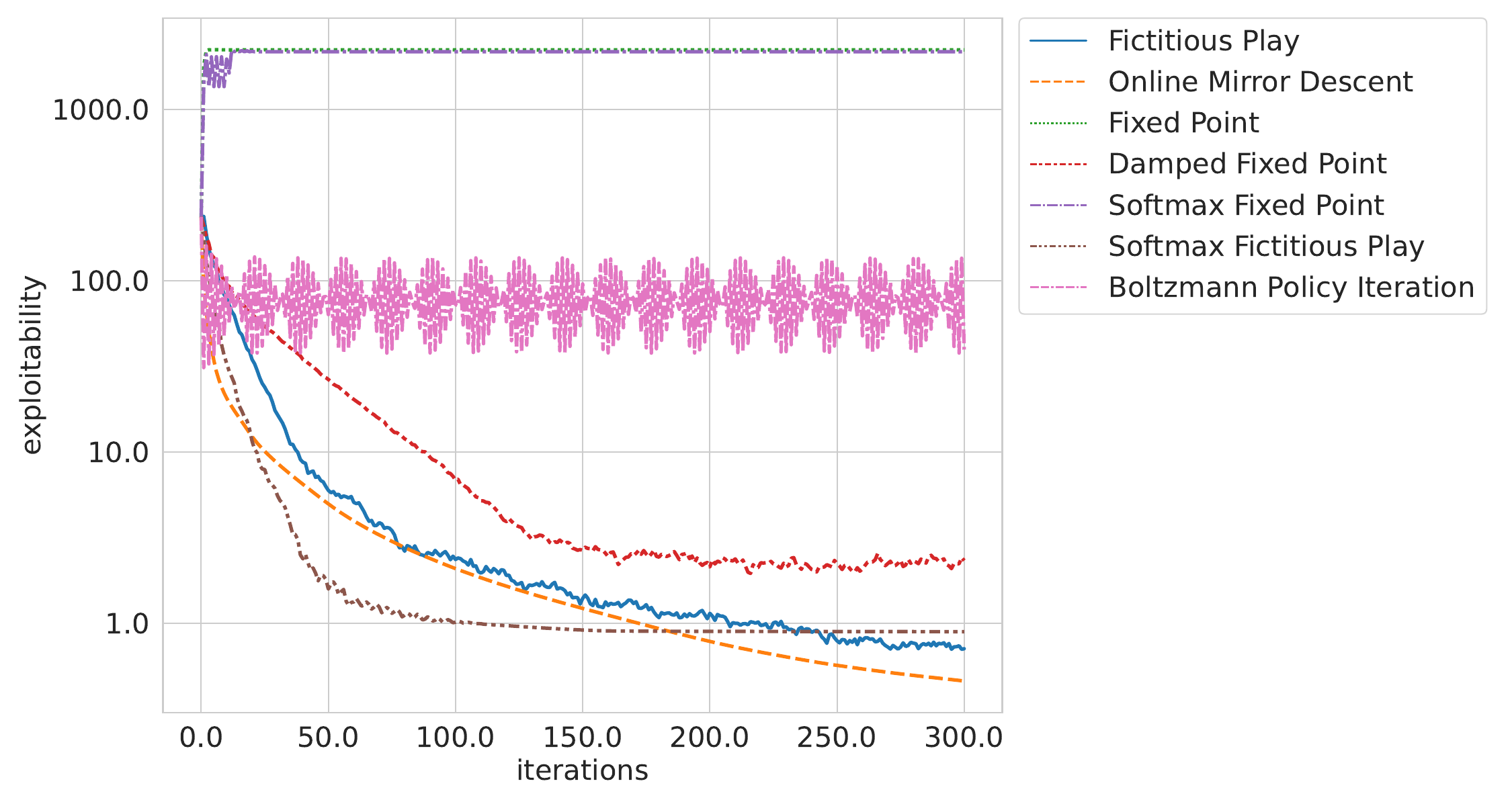}
    \end{minipage}
    
    \caption{Four room exploration with noise intensity $p=0.5$. Exploitability.}
    \label{fig:expe-four-room-explo-noise0.5-exploitabilities}
\end{figure}

\begin{figure}[tbh!]
    \centering
    \begin{minipage}{.7\linewidth}
    \includegraphics[width=\linewidth]{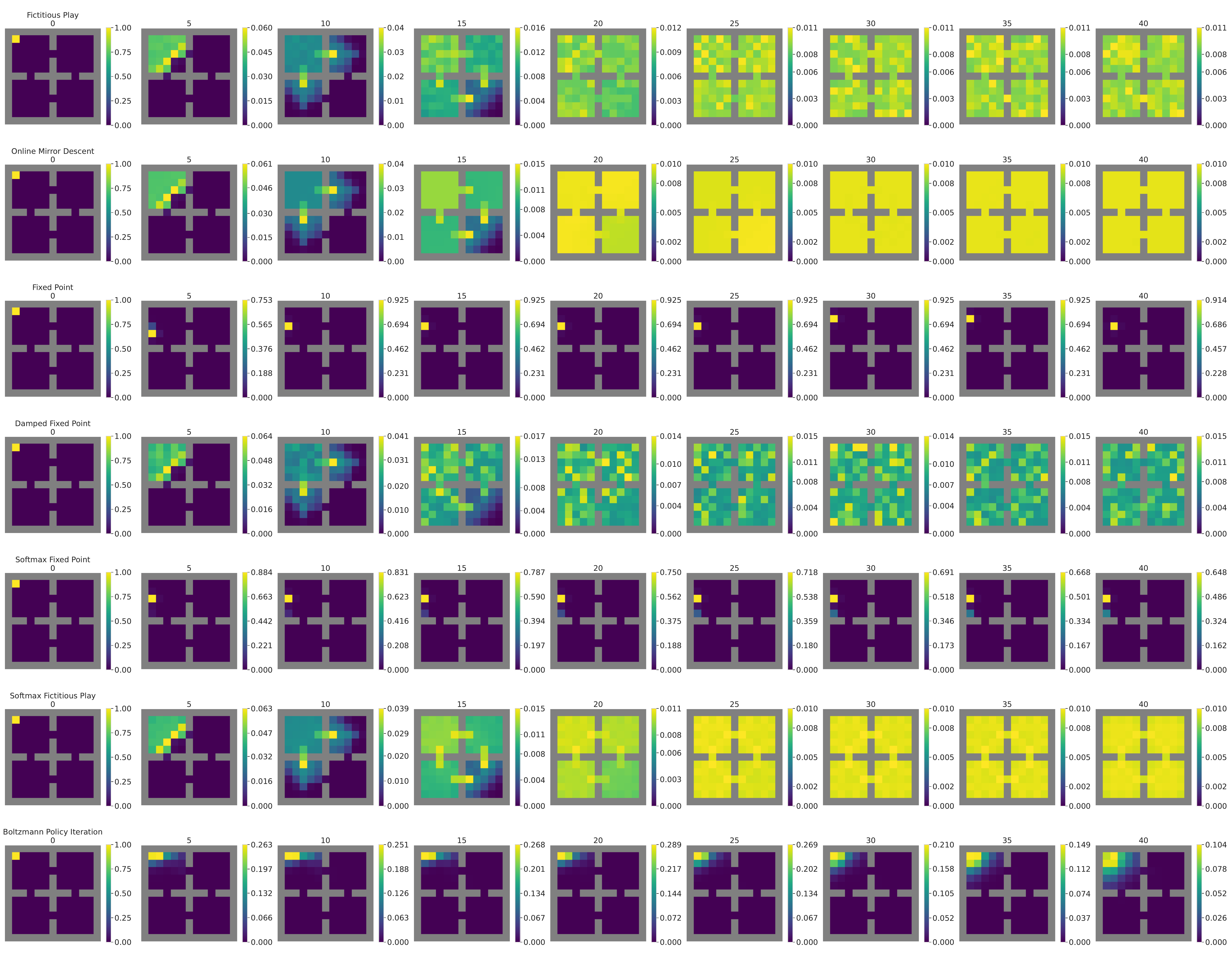}
    \end{minipage}
    
    \caption{Four room exploration with noise intensity $p=0.1$. Distribution evolution.}
    \label{fig:expe-four-room-explo-noise0.1-distributions}
\end{figure}

\begin{figure}[tbh!]
    \centering
    \begin{minipage}{.7\linewidth}
    \includegraphics[width=\linewidth]{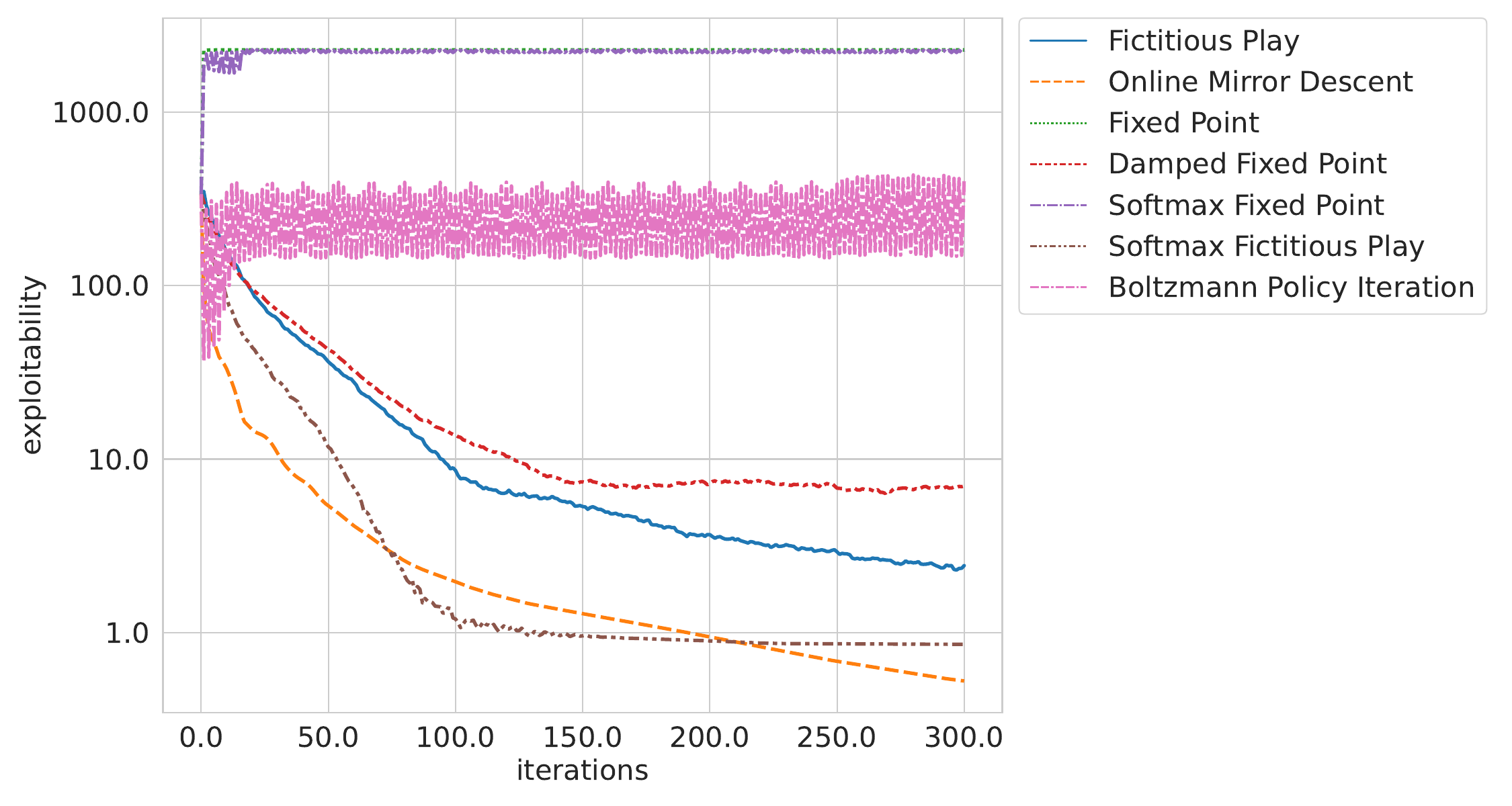}
    \end{minipage}
    
    \caption{Four room exploration with noise intensity $p=0.1$. Exploitability.}
    \label{fig:expe-four-room-explo-noise0.1-exploitabilities}
\end{figure}

\paragraph{Numerical results for sweep over hyperparameters. }
To better understand the impact of the learning rates and softmax temperatures, we provide below the results for sweeps over these hyperparameters. %

Figure~\ref{fig:expe-four-room-explo-omd-sweep-exploitabilities} shows the results with Online Mirror Descent for various learning rate values.
Figure~\ref{fig:expe-four-room-explo-dampedfp-sweep-exploitabilities} shows the results with Damped Fixed Point for various damping values.
Figure~\ref{fig:expe-four-room-explo-softmaxfp-sweep-exploitabilities} shows the results with Softmax Fixed Point for various softmax temperature values. 
Figure~\ref{fig:expe-four-room-explo-bpi-sweep-exploitabilities} shows the results with Boltzmann Policy Iteration for various learning rates.

We see that when the learning rates or damping values are too small or too large, the algorithm works less well or even does not converge at all. Softmax Fixed Point seems to always fail on this example. Botlzmann temperature works relatively well for moderately large values of the learning rate, but never manages to converge because the softmax prevents it from learning the true Nash equilibrium policy. 

\begin{figure}[tbh!]
    \centering
    \begin{minipage}{.7\linewidth}
    \includegraphics[width=\linewidth]{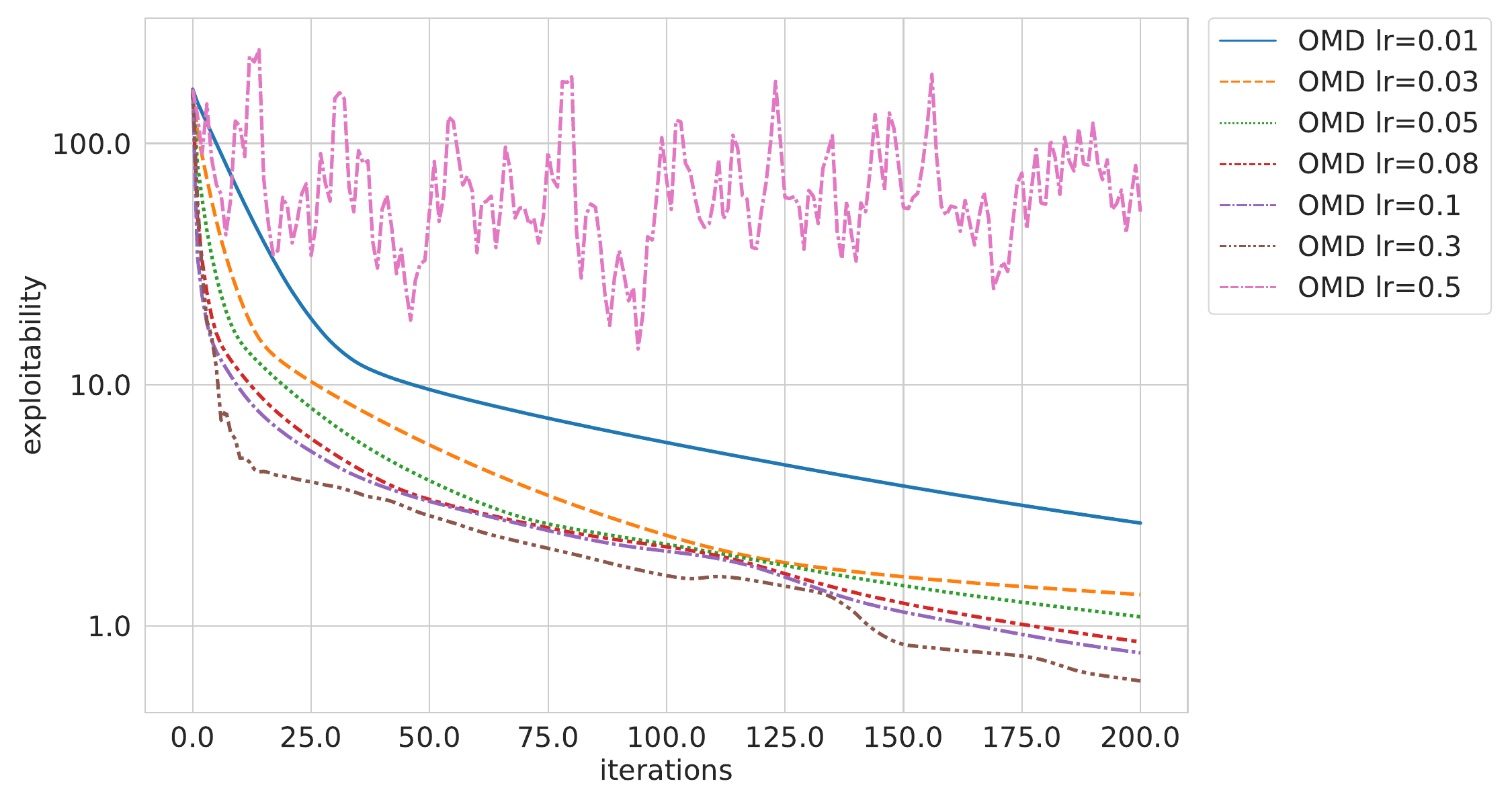}
    \end{minipage}
    
    \caption{Four room exploration with Online Mirror Descent for various learning rates. Exploitability.}
    \label{fig:expe-four-room-explo-omd-sweep-exploitabilities}
\end{figure}

\begin{figure}[tbh!]
    \centering
    \begin{minipage}{.7\linewidth}
    \includegraphics[width=\linewidth]{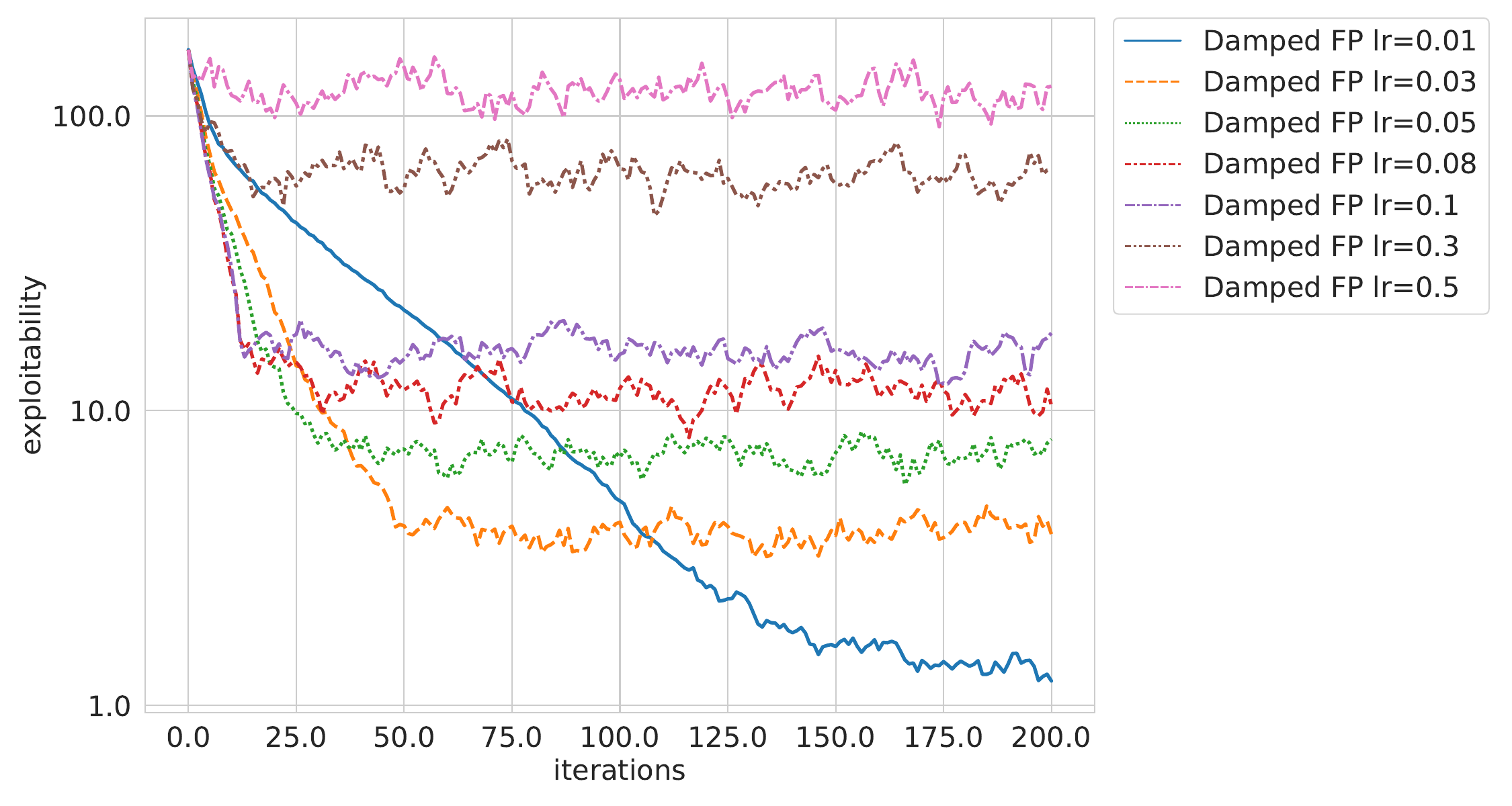}
    \end{minipage}
    
    \caption{Four room exploration with Damped Fixed Point for various damping coefficients. Exploitability.}
    \label{fig:expe-four-room-explo-dampedfp-sweep-exploitabilities}
\end{figure}

\begin{figure}[tbh!]
    \centering
    \begin{minipage}{.7\linewidth}
    \includegraphics[width=\linewidth]{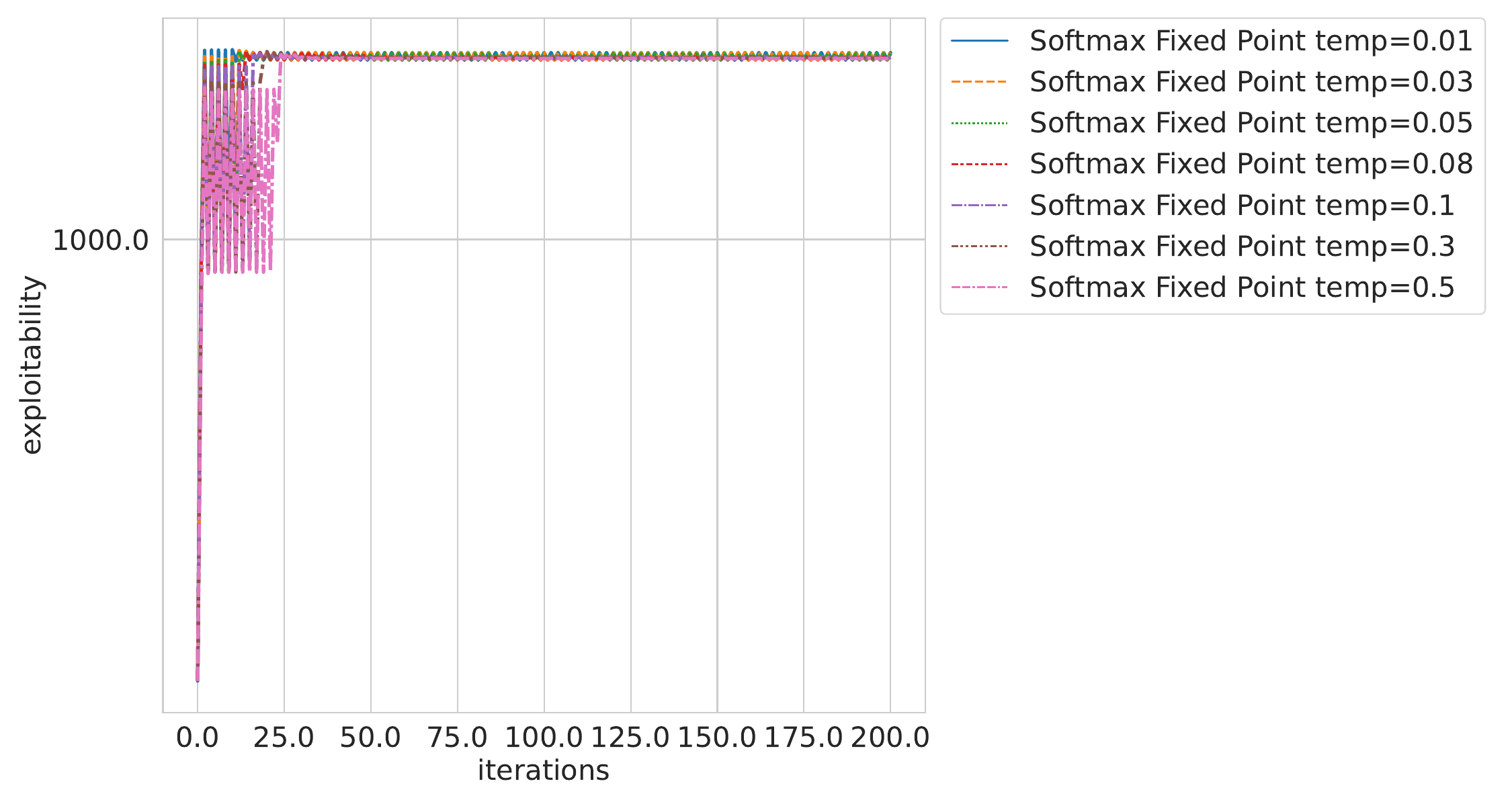}
    \end{minipage}
    
    \caption{Four room exploration with Softmax Fixed Point for various softmax temperatures. Exploitability.}
    \label{fig:expe-four-room-explo-softmaxfp-sweep-exploitabilities}
\end{figure}

\begin{figure}[tbh!]
    \centering
    \begin{minipage}{.7\linewidth}
    \includegraphics[width=\linewidth]{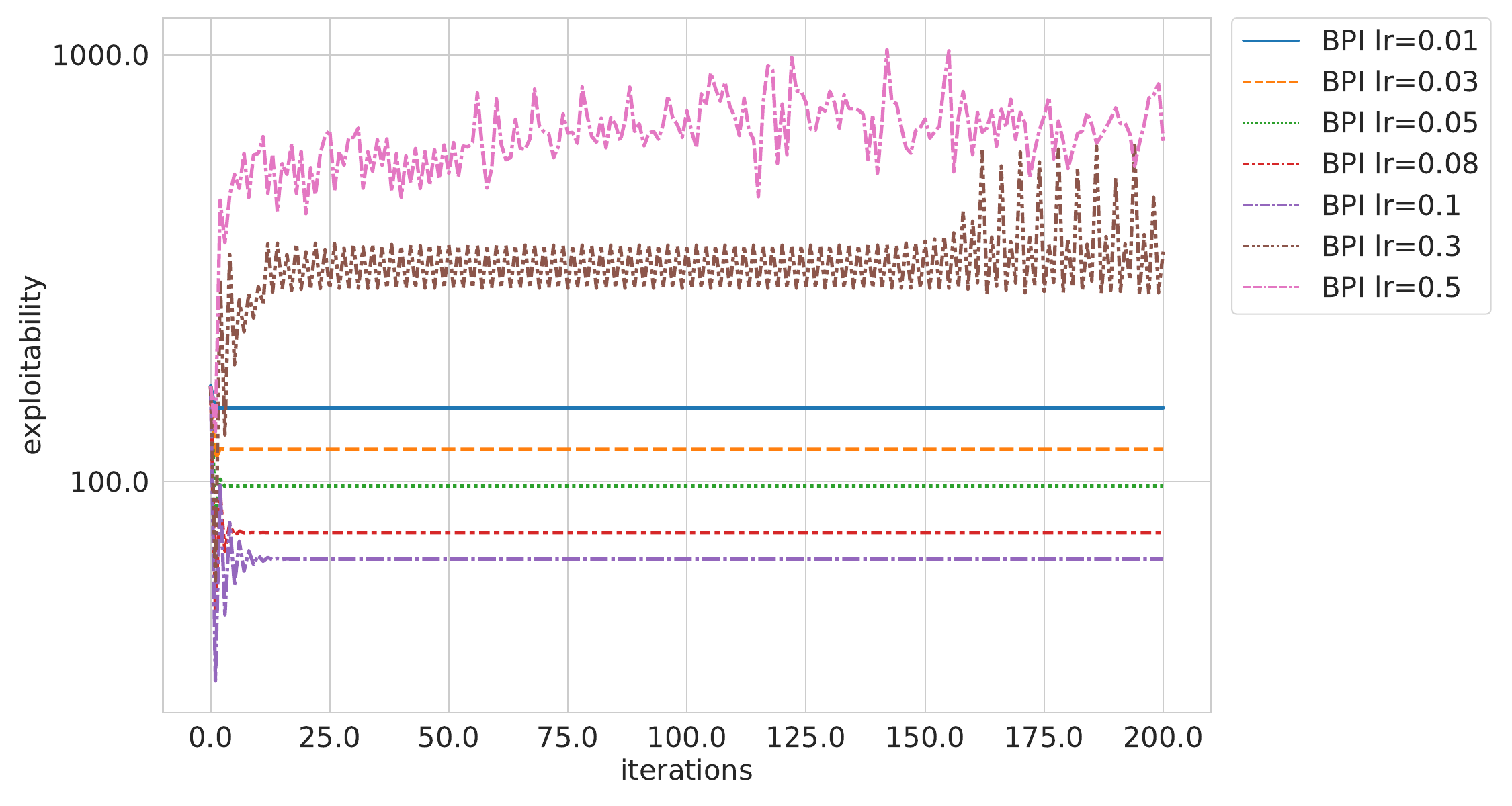}
    \end{minipage}
    
    \caption{Four room exploration with Boltzmann Policy Iteration for various learning rates. Exploitability.}
    \label{fig:expe-four-room-explo-bpi-sweep-exploitabilities}
\end{figure}

\subsection{Avoidance game}
We then consider a game with two populations of players. At time $n$, the description of the mean field is a pair of probability distributions, $\mu^1_n \in \Delta_\states$ and $\mu^2_n \in \Delta_\states$, which represent the distribution of states of each population. 
There are walls forming two rooms connected by two doors. Each population starts in a room (the initial distribution is concentrated in that room) and wants to go to the other one (there is a positional reward in that room). The populations start in different rooms. The reward functions for population $i \in \{1,2\}$ are given by:
\begin{itemize}
    \item $r_{\mathrm{pos}}^i(x) = - \mathrm{dist}(x, x_{\mathrm{ref}}^i)$ is the distance to the target position $x_{\mathrm{ref}}^i$ for population $i$;
    \item $r_{\mathrm{move}}(a,\mu^1(x),\mu^2(x)) = 0$;
    \item $r_{\mathrm{pop}}^i(\mu^1(x),\mu^2(x)) = -c_{\mathrm{cong}}\mu^{-i}(x)$, where $\mu^{-i}(x)$ is the density of players of the other population at the location $x$.
\end{itemize}  
Here, $c_{\mathrm{cong}}$ is a non-negative constant. When it is non-zero, the players are incentivized to avoid crossing the path of the other population. This pushes the two populations to use different doors.

 We then consider a game with two populations of players evolving in two rooms connected by two doors. At time $n$, the description of the mean field is a pair of probability distributions, $\mu^1_n \in \Delta_\states$ and $\mu^2_n \in \Delta_\states$, which represent the distribution of states of each population. Each population starts in one room and wants to go to the other room. If there is a congestion cost, they want to avoid each other, so they tend to use different doors.  

\paragraph{Model. }
We consider the following model:
\begin{itemize}
    \item {\bf State space} Similar to the four room exploration task (see Section~\ref{sec:experiments-four-rooms-exploration}).
    \item {\bf Action space: } Similar to the four room exploration task (see Section~\ref{sec:experiments-four-rooms-exploration}).
    \item {\bf Transitions: } Similar to the four room exploration task (see Section~\ref{sec:experiments-four-rooms-exploration}), except that the walls (forbidden states) are located to form a maze.
    \item {\bf Time horizon: } Similar to the four room exploration task (see Section~\ref{sec:experiments-four-rooms-exploration}).
    \item {\bf Rewards: } The players pay a cost (negative reward) to move, and this cost increases when the density of agents increases. There is also an incentive to reach the target position. The reward function is given by:
    $$
        r(x,a,\mu^1,\mu^2) =  r_{\mathrm{pos}}(x) +r_{\mathrm{pop}}^i(\mu^1(x),\mu^2(x)), 
    $$
    where 
    \begin{itemize}
        \item[$\bullet$] $r_{\mathrm{pos}}^i(x) = - \mathrm{dist}(x, x_{\mathrm{ref}}^i)$ is the distance to the target position $x_{\mathrm{ref}}^i$ for population $i$;
        \item[$\bullet$] $r_{\mathrm{pop}}^i(\mu^1(x),\mu^2(x)) = -c_{\mathrm{cong}}\mu^{-i}(x)$, where $\mu^{-i}(x)$ is the density of players of the other population at the location $x$.
    \end{itemize}  
    For each population, the target position is in the opposite room.
    \item {\bf Initial distribution: } Each population start in a different room. The first population starts int he left room, while the second population starts in the right room. Each initial distribution is split into two parts, one in from of each door. But the first population has more mass in front of the top door, while the second population has more mass in front of the bottom door.
\end{itemize}

\paragraph{Numerical setup. }
In the experiments below, we use the following parameters:
\begin{itemize}
    \item {\bf State space: } We use $N_x^1 = N_x^2 = 7$, with walls on the sides, and walls inside the domain to form a maze.
    \item {\bf Time horizon: } $N_T=11$.
    \item {\bf Noise intensity: } $p = 0$ (no noise).
    \item {\bf Rewards: } We use $c_{\mathrm{cong}} = 7.0$ (high level of congestion), and $c_{\mathrm{cong}} = 0.0$ (no congestion).
    \item {\bf Hyperparameters: } The learning rate is $0.05$ for Online Mirror Descent, $0.01$ for Damped Fixed Point, and the temperature is $0.1$ for softmax in Softmax Fixed Point, Softmax Fictitious Play and Boltzmann Policy Iteration. %
\end{itemize}

\paragraph{Numerical results. }

We start with the case without congestion, $c_{\mathrm{cong}}=0$. Figure~\ref{fig:expe-avoidance-congboth-distributions} (left plot) displays the distribution evolution generated by using the policy learnt by each algorithm. Pairs of rows correspond to different algorithms (first row for first population, and second row for second population) while columns correspond to different time steps. Figure~\ref{fig:expe-avoidance-congboth-exploitabilities} (bottom plot) shows, for each algorithm, the exploitability as a function of the iteration number. We see that, in this example, (pure) Fixed Point converges extremely quickly. We see that the first population moves from the left room to the right room as expected, and each part of the mass uses the door in front of it. (which is the shortest path).  
Softmax Fixed Point does very well but the exploitability is bounded away from $0$ due to the fact that it uses softmax policies. 

Figures~\ref{fig:expe-avoidance-congboth-distributions} (right plot) and~\ref{fig:expe-avoidance-congboth-exploitabilities} (bottom plot) display similar results but for the case with large congestion coefficient, $c_{\mathrm{cong}}=7$. Fixed Point still converges very well. But now, we see that the smaller part of the mass does not use the shortest path. For example, for population $1$, the mass in front of the bottom door in fact uses the top door. This can be explained by the congestion cost: if that part of the mass was using the bottom door, it would cross the second population and pay a large cost. Furthermore, in this setting, it seems that Online Mirror Descent is not converging (or only extremely slowly). In the distribution plot, we see that the distributions generated by the result of the OMD algorithm are simply diffusing in each room instead of moving to the other room.

\begin{figure}[tbh!]
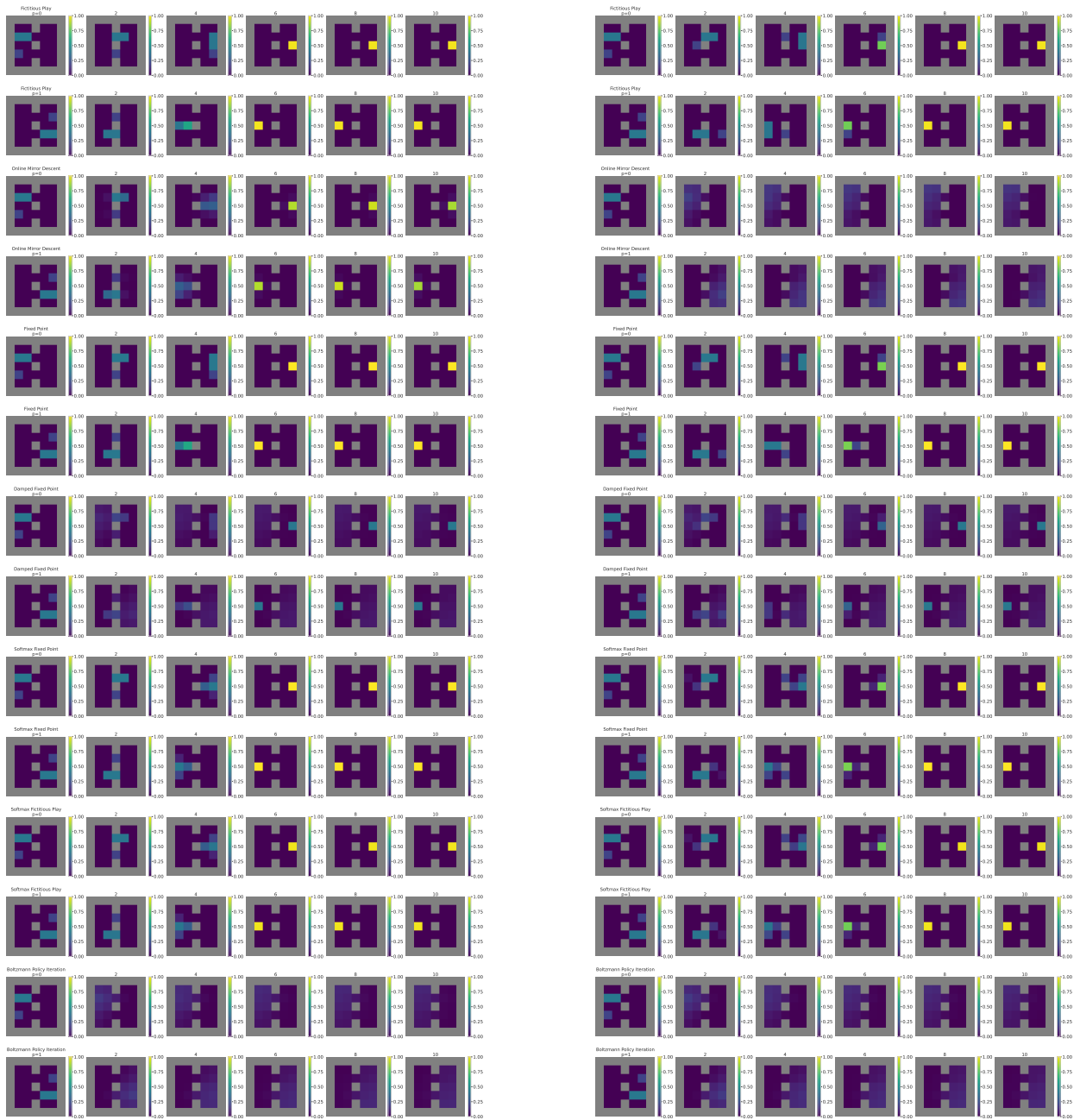

    \centering
    \begin{minipage}{.45\linewidth}
    \includegraphics[width=\linewidth]{FIGURES/EXPERIMENTS-202307/3_MFG-Avoidance-2rooms2pop/avoidance-cong0.0-target1.0-initpolneutral_distributions}
    \end{minipage}
    \hfill
    \begin{minipage}{.45\linewidth}
    \includegraphics[width=\linewidth]{FIGURES/EXPERIMENTS-202307/3_MFG-Avoidance-2rooms2pop/avoidance-cong7.0-target1.0-initpolneutral_distributions}
    \end{minipage}
    
    \caption{Avoidance Game with no congestion $c_{\mathrm{cong}}=0$  (left) and with congestion $c_{\mathrm{cong}}=7$ (right). Distribution.}
    \label{fig:expe-avoidance-congboth-distributions}
\end{figure}

\begin{figure}[tbh!]
    \centering
    \begin{minipage}{.8\linewidth}
    \includegraphics[width=\linewidth]{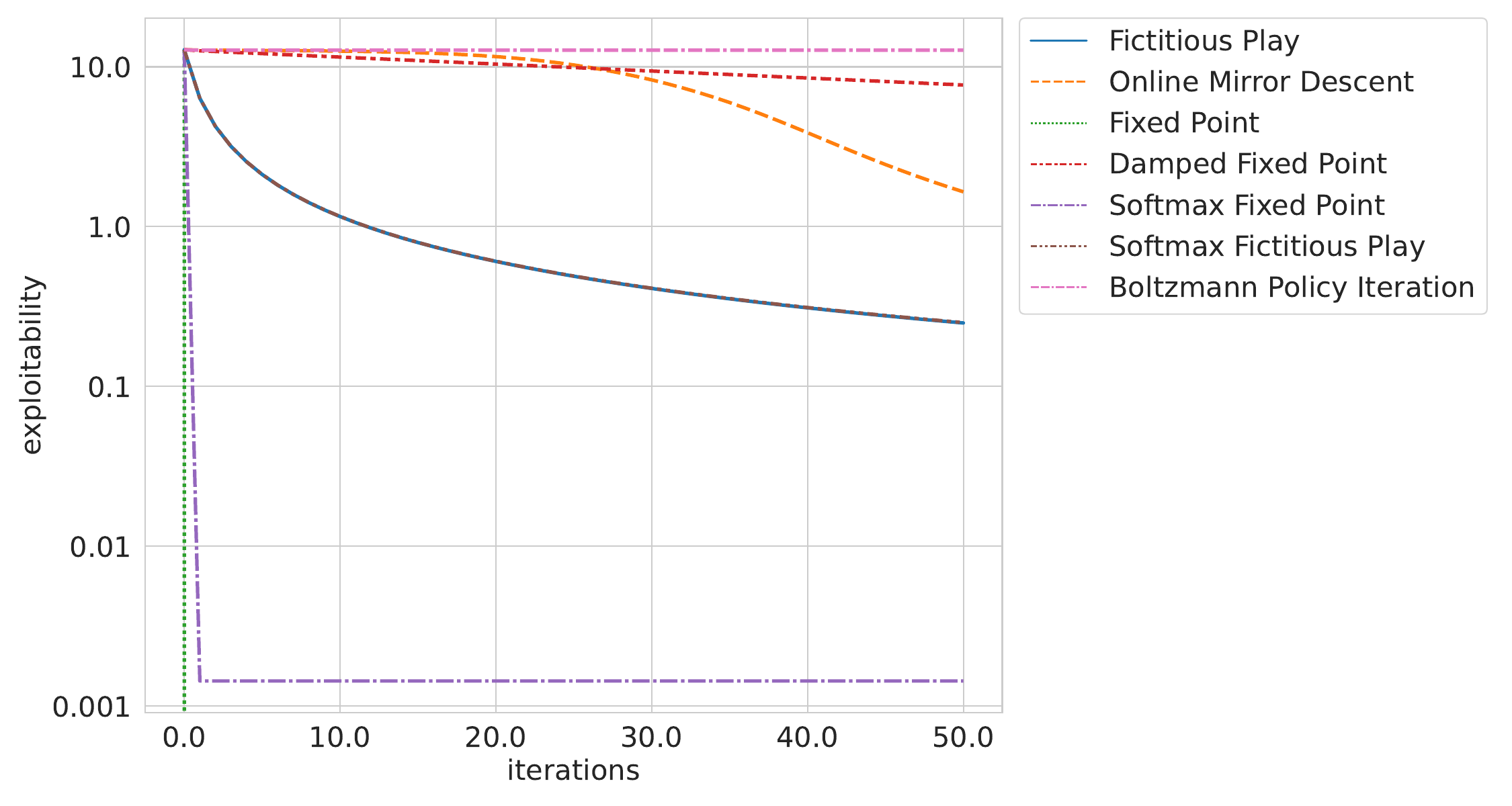}
    \end{minipage}
    
    \begin{minipage}{.8\linewidth}
    \includegraphics[width=\linewidth]{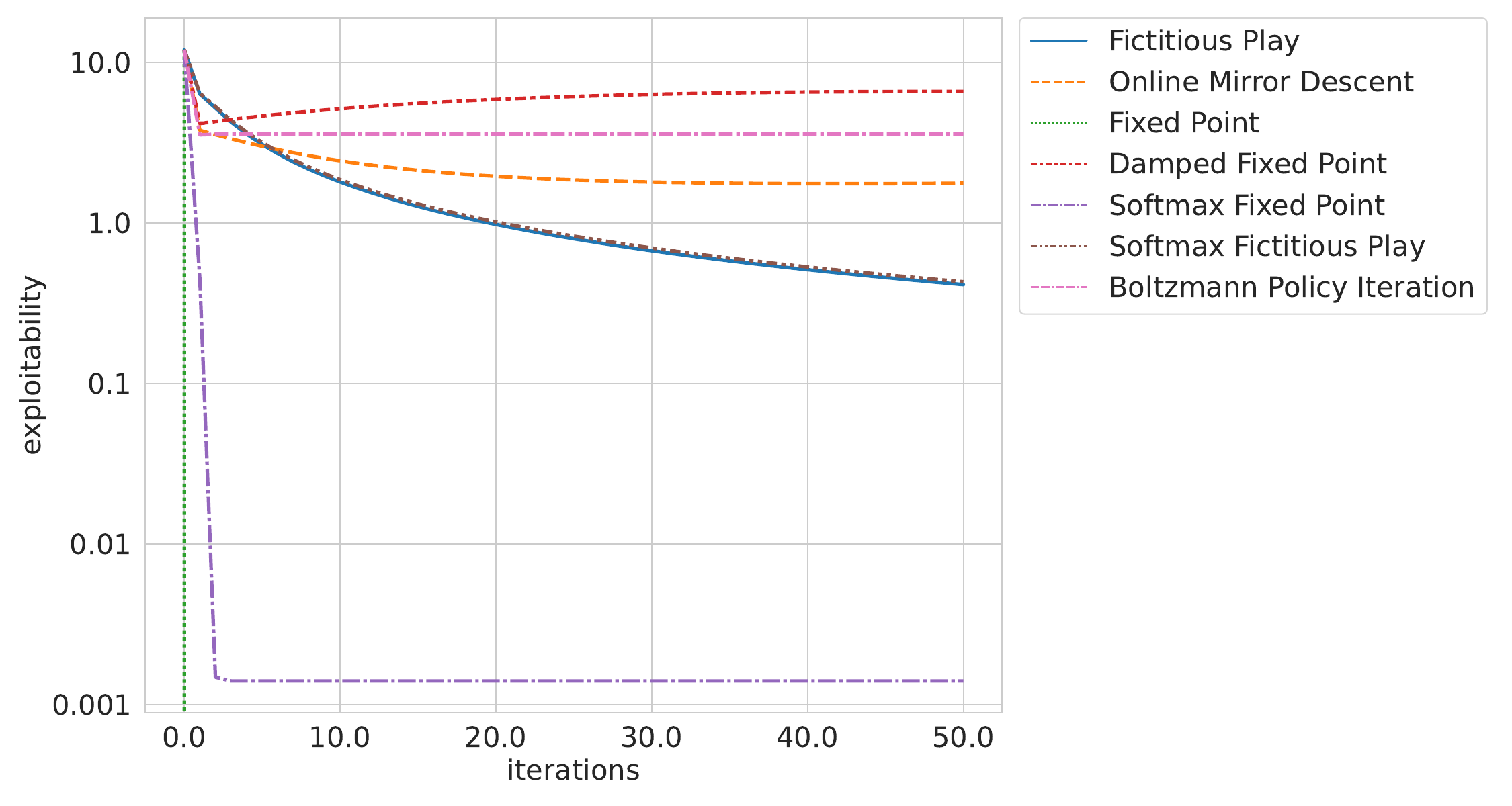}
    \end{minipage}
    
    \caption{Avoidance Game with no congestion $c_{\mathrm{cong}}=0$  (top) and with congestion $c_{\mathrm{cong}}=7$ (bottom). Exploitability.}
    \label{fig:expe-avoidance-congboth-exploitabilities}
\end{figure}

\clearpage

\subsection{Predator-prey game}

We then consider a game with $N_p = 4$ populations of players, in a 2D grid world. Each population chases the next one (their prey), and tries to avoid the previous one (their predator). So the players in population $p$ are attracted to the distribution of population $p+1$ and want to avoid the distribution of population $p-1$ (where $p+1$ and $p-1$ are understood modulo $N_p$). 
If there are $N_p$ populations, at time $n$, the description of the mean field is a tuple of probability distributions, $\mu^p_n \in \Delta_\states$, $p=0,\dots,N_p-1$. This game is analogous to the predator-prey example in~\citep{perolat2022scaling}.

\paragraph{Model. }
We consider the following model:
\begin{itemize}
    \item {\bf State space} Similar to the four room exploration task (see Section~\ref{sec:experiments-four-rooms-exploration}).
    \item {\bf Action space: } Similar to the four room exploration task (see Section~\ref{sec:experiments-four-rooms-exploration}).
    \item {\bf Transitions: } Similar to the four room exploration task (see Section~\ref{sec:experiments-four-rooms-exploration}) but there are no forbidden states inside the domain (only walls around it, to prevent them from exiting).
    \item {\bf Time horizon: } Similar to the four room exploration task (see Section~\ref{sec:experiments-four-rooms-exploration}).
    \item {\bf Rewards: } Each agent gets rewarded for being at the same location as their prey and gets penalized for being at the same location as their predator. There is also a congestion cost within each population. The reward function for population $p \in \{1,\dots,N_p\}$ is given by:
    $$
        r(x,a,\mu^1,\dots,\mu^{N_p-1}) = r_{\mathrm{pop}}^p((\mu^p(x))_{p=0\dots,N_p-1}) = -c_{\mathrm{cong}}\mu^{p}(x) + \mu^{p+1}(x) - \mu^{p-1}(x), 
    $$
    where $c_{\mathrm{cong}}$ is a non-negative constant. 
    For each population, the target position is in the opposite room.
    \item {\bf Initial distribution: } Each population start in a different corner.
\end{itemize}

\paragraph{Numerical setup. }
In the experiments below, we use the following parameters:
\begin{itemize}
    \item {\bf State space: } We use $N_x^1 = N_x^2 = 5$, with walls on the sides, and walls inside the domain to form a maze.
    \item {\bf Time horizon: } $N_T=11$.
    \item {\bf Noise intensity: } $p = 0.1$. 
    \item {\bf Rewards: } We use $c_{\mathrm{cong}} = 0.1$. 
    \item {\bf Hyperparameters: } The learning rate is $0.05$ for Online Mirror Descent, $0.01$ for Damped Fixed Point, and the temperature is $0.1$ for softmax in Softmax Fixed Point, Softmax Fictitious Play and Boltzmann Policy Iteration. %
\end{itemize}

Figure~\ref{fig:expe-predatorprey-distributions} displays the distribution evolution generated by using the policy learnt by each algorithm. Pairs of rows correspond to different algorithms (first row for first population, and second row for second population) while columns correspond to different time steps. We see that the populations chase each other while spreading throughout the domain.  Figure~\ref{fig:expe-predatorprey-exploitabilities} shows, for each algorithm, the exploitability as a function of the iteration number. We see that, in this example, (pure) Fixed Point converges extremely quickly. We see that the first population moves from the left room to the right room as expected, and each part of the mass uses the door in front of it. (which is the shortest path).  
Softmax Fixed Point does very well but the exploitability is bounded away from $0$ due to the fact that it uses softmax policies.

\begin{figure}[tbh!]
    \centering
    \begin{minipage}{.4\linewidth}
    \includegraphics[width=\linewidth]{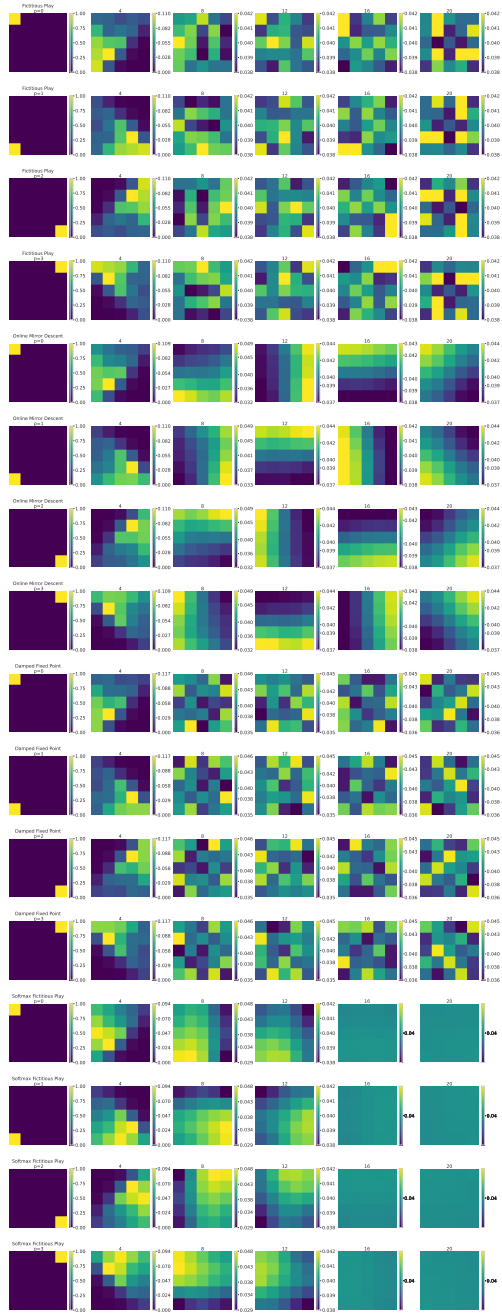}
    \end{minipage}
    
    \caption{Predator-Prey with 4 populations. Distribution for the best performing algorithms only.}
    \label{fig:expe-predatorprey-distributions}
\end{figure}

\begin{figure}[tbh!]
    \centering
    \begin{minipage}{.48\linewidth}
    \includegraphics[width=\linewidth]{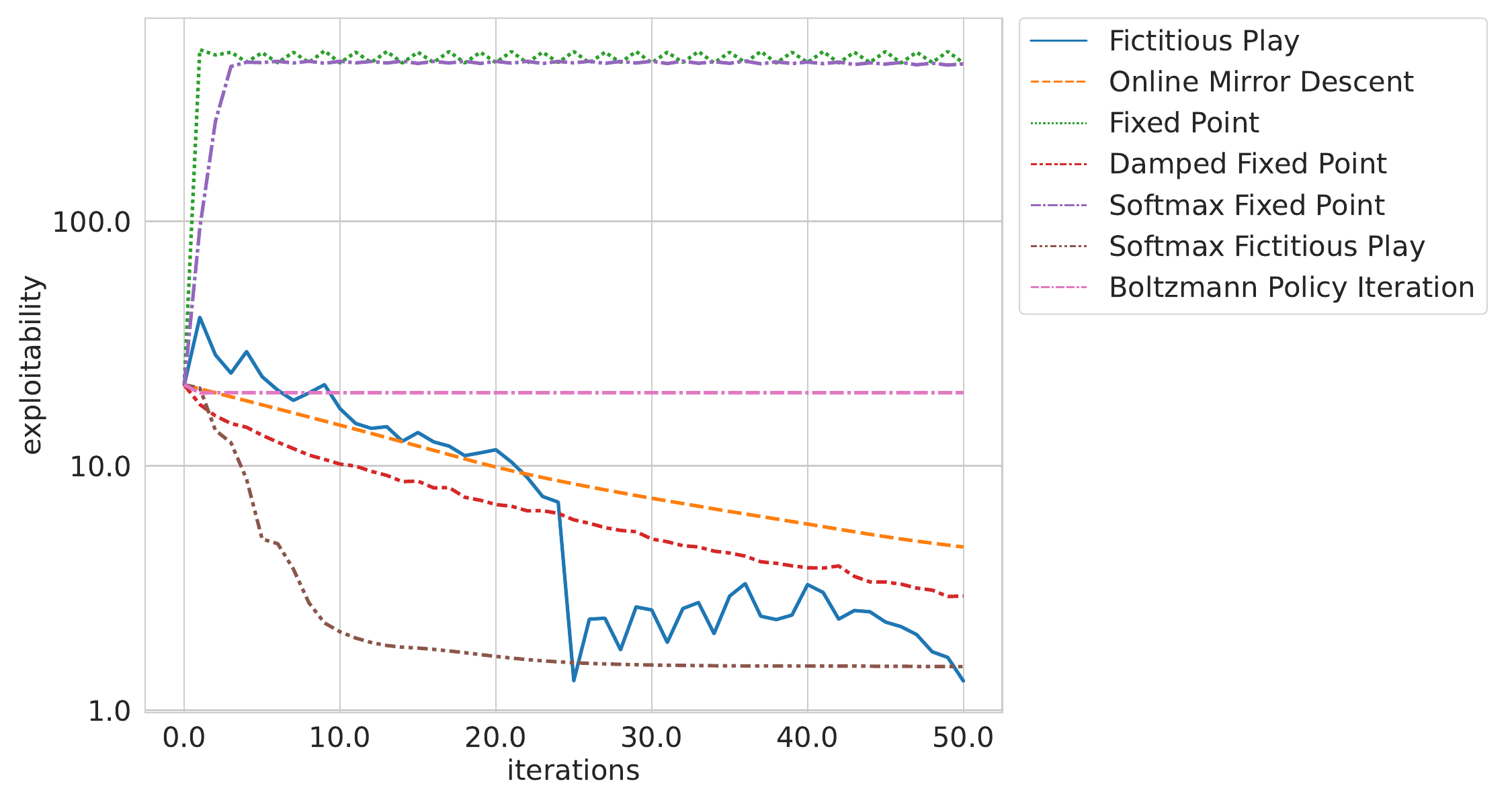}
    \end{minipage}
    \begin{minipage}{.48\linewidth}
    \includegraphics[width=\linewidth]{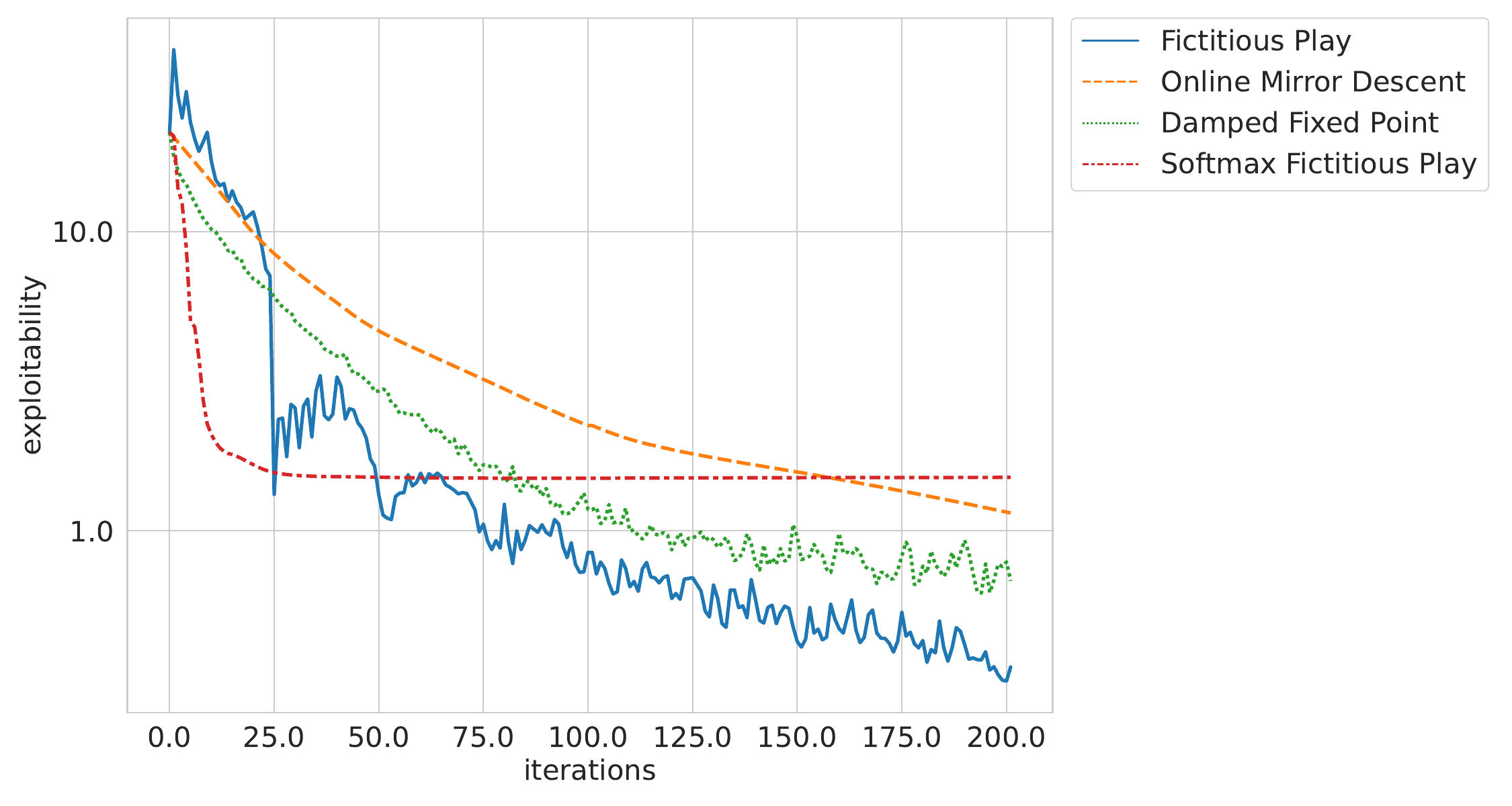}
    \end{minipage}
    
    \caption{Predator-Prey with 4 populations. Exploitability. Left: Same algorithms as in previous tests, over a small number of iterations. Right: Only the best performing algorithms, over a larger number of iterations.}
    \label{fig:expe-predatorprey-exploitabilities}
\end{figure}

\clearpage

\subsection{Experiments with RL}
\label{sec:expe-RL}

As illustrated by the above experiments, algorithms with ``exact'' updates may have different behaviors depending on the model at hand and further investigations in this direction will be required to gain a finer understanding of when each method performs best. Things get even more complex when these methods are used in combination with model-free RL. For the sake of illustration, we provide here one example, implemented avain in \texttt{OpenSpiel}, which contains various tabular and deep RL methods. Although (deep) RL methods are useful mostly for large-scale examples, here we will consider a small-scale example only for the sake of illustration. 

\paragraph{Model. }
We consider the following model, which is inspired by the model studied in~\citep{elie2020convergencemodelfreemfg}:
\begin{itemize}
    \item {\bf State space: } $\states = \{0,\dots,N_x-1\}$, which represents a 1D grid world.
    \item {\bf Action space: } $\actions = \{-1,0,-1\}$, which represents movement in the 2 directions and the absence of movement. They correspond respectively to: left, stay, right.
    \item {\bf Transitions: } At time $n$, the agent chooses to stay at the current position or to move to one of the neighboring positions. Furthermore, a random disturbance potentially affects the dynamics. The next state is computed according to the dynamics:
    $$
        x_{n+1} = 
        \begin{cases}
            x_{n} + a_{n} + \epsilon_{n+1}, \quad &\hbox{ if } x_{n} + a_{n} + \epsilon_{n+1} \hbox{modulo $N_x$} 
            \\
            x_{n}, \quad &\hbox{otherwise,} 
        \end{cases}
    $$
    where $(\epsilon_{n})_n$ is a sequence of i.i.d. random variables taking values in $\actions$. In the experiments below, we take a uniform distribution of noise. Notice that the ``modulo $N_x$'' induces periodic boundary conditions. 
    \item {\bf Time horizon: } The rewards are accumulated without discount and only until a terminal time horizon $N_T$.
    \item {\bf Rewards: } $r(x,a,\mu) = (1 - \frac{1}{N_x/2}|x - \frac{N_x}{2}|) -\frac{|a|}{N_x} -\log(\mu(x))$, which has the following effects: it encourages the agent to move towards the center of the domain, it discourages the agent from moving when it is not beneficial, and it encourages the agent to avoid crowded areas. The constants can be adjusted to strengthen some of these three effects.
    \item {\bf Initial distribution: }  The initial distribution is uniform over the state space.
\end{itemize}

\paragraph{Numerical setup. }
In the experiments below, we use the following parameters:
\begin{itemize}
    \item {\bf State space: } We use $N_x=10$, with periodic boundary conditions. 
    \item {\bf Time horizon: } $N_T=10$.
    \item {\bf Noise: } uniform distribution over the action space.
\end{itemize}

\paragraph{Numerical results. }Figure~\ref{fig:expe-1dgrid-comp-ex-rl} shows numerical results obtained on this example using three different variants of fictitious play, which differ by how the BR is computed: 
\begin{itemize}
    \item Exact Fictitious Play: the BR is computed using backward induction, without RL at all;
    \item Tabular Q-learning Fictitious Play: the BR is computed using tabular Q learning;
    \item Deep Fictitious Play: here we use the average network deep fictitious play introduced in~\cite{lauriere2022scalable}; an ``average'' neural network for the Q-function corresponding to the average policy is trained over the fictitious play iterations; during each iteration of fictitious play, DQN is used to learn a best response, which is in turn used to improve the average neural network. 
\end{itemize}
In each case, the mean field induced by the average policy is computed exactly.

The top plot displays the distribution at several time steps using the policy learnt by fictitious play with exact updates. We can see that the population, starting from a uniform distribution moves towards a distribution which is concentrated around the center of the spatial domain (but not completely, due to crowd aversion effects), and then takes a different shape close to the terminal horizon because there is not much incentive to be close to the center of the domain when there is no time left. This behavior can be viewed as an instance of the turnpike phenomenon; see~\cite{BricenoAriasetalCEMRACS2017,porretta2018turnpike,trusov2020numerical,cirant2021long}. 

The bottom plot displays the exploitability curves, as a function of the Fictitious Play iterations. For fictitious play with tabular Q-learning updates, we use 50 episodes of Q-learning to compute the BR during each iteration fictitious play. For the Deep Fictitious Play, we compare two versions: one which uses only 50 episodes of training for the BR and the average network during each iteration of fictitious play, and another one which uses 200 episodes. For each of the three algorithms involving RL, we ran the algorithm $10$ times and computed the average exploitability; the shaded region shows the mean $\pm$ standard deviation over these $10$ runs. 
We see that the version with exact updates and the tabular RL version have similar performance, which is not surprising on a small-scale example such as the one considered here. As for the deep RL method, the performance is not as good, which might be due to the fact that we have not tried to find the best hyperparameters. However, we notice that increasing the number of training steps inside each fictitious play iteration seems to improve the exploitability and decreases the variance. 

\begin{figure}[tbh!]
    \centering
    \begin{minipage}{.9\linewidth}
    \includegraphics[width=\linewidth]{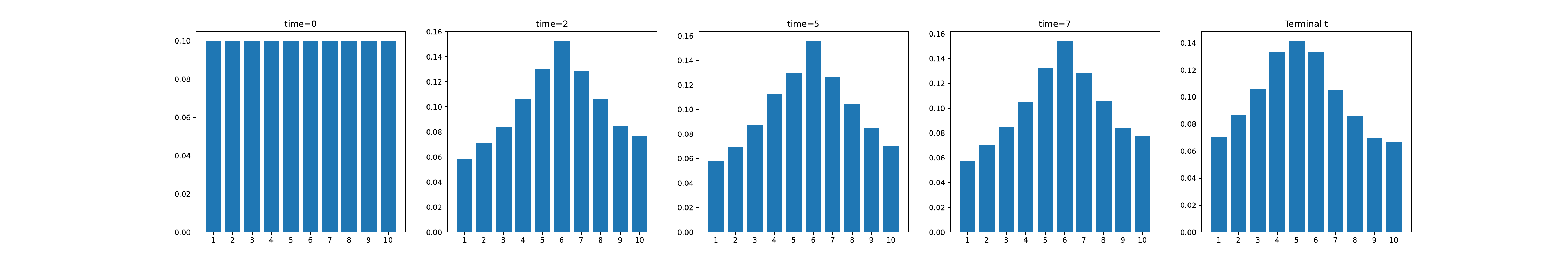}
    \end{minipage}
    
    \begin{minipage}{.45\linewidth}
    \includegraphics[width=\linewidth]{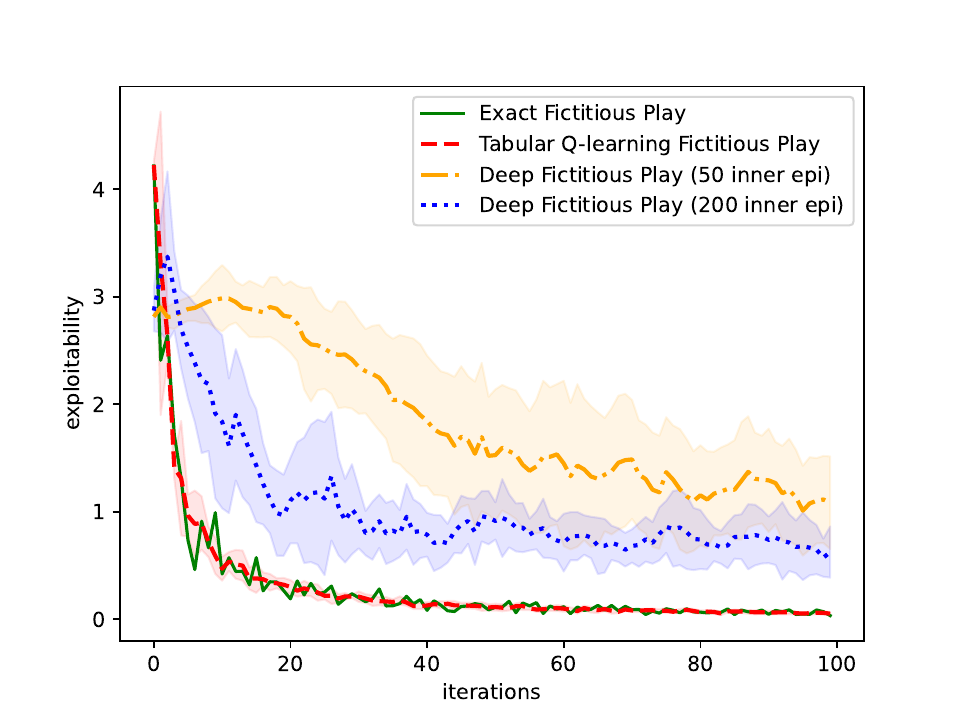}
    \end{minipage}
    
    \caption{1D grid world model. 
    Top: distribution at several time steps using the policy learnt by fictitious play with ``exact'' updates. Bottom: exploitability of several algorithms, as described in the text.}
    \label{fig:expe-1dgrid-comp-ex-rl}
\end{figure}

\section{An introduction to MFGs in \texttt{OpenSpiel}}
\label{app:intro-mfg-openspiel}

\texttt{OpenSpiel} covers a variety of implemented environments for MFGs as well as state of the art learning algorithms. It provides a relevant playground to develop additional MFG environments, as well as test and benchmark new algorithms. This section provides a brief introduction to this open source library, focusing on MFGs.

We distinguish \defi{games} (or environments in the RL terminology) and \defi{algorithms}. Intuitively, a game contains everything that is needed to define the model, whereas an algorithm is dedicated to computing a solution to the problem. 

For now, \texttt{OpenSpiel} focuses primarily on the evolutive setting described in Section \ref{sec:evol-mfg-setting}. The stationary setting is indirectly supported through transforming stationary games into evolutive ones, and other settings could also be implemented. In the sequel, we restrict our attention to evolutive MFGs.

\subsection{Games structure} 

Games can be implemented in C++ or in python. We discuss here the implementation in Python, but implementation in C++ follows the same lines. The games in Python are located in open\_spiel/tree/master/open\_spiel/python/mfg/games. 

\paragraph{Players. } 
\texttt{OpenSpiel} has been originally designed to support game structures with a finite number of players. For the sake of consistency with other games implemented in \texttt{OpenSpiel}, the MFG evolution is represented using the notion of \texttt{OpenSpiel} players. More specifically, single population MFGs are implemented within a framework of two \texttt{OpenSpiel} players: the representative player called \texttt{DEFAULT\_PLAYER\_ID} is one player and the population called \texttt{MEAN\_FIELD} is the other one. When randomness occur in the game dynamics, it is encoded as an additional third player called \texttt{CHANCE}.
To compute the evolution of one player's state, the dynamics can be viewed as sequence of nodes. One round consists in updating both the representative player state and the mean field (i.e. population) one. The idea is that three players: the representative  \texttt{DEFAULT\_PLAYER\_ID} one, the population \texttt{MEAN\_FIELD} one and the \texttt{CHANCE} one, influence one transition between two nodes, in turn. %
This structure is the main difference i comparison to the way MFGs are presented in this survey (and more generally in the literature). 

\paragraph{Initial and terminal states. } Randomness at initial time is also encoded as an action of the chance player. However, this randomness is based on a distribution over the state space and not over the action space. 
The end of the game (e.g., the finite horizon) is encoded through the notion of \defi{terminal state}. The game starts from an initial state and runs until a terminal state is reached. This can be used to represent finite-horizon MFGs by considering that the state of the game is not only the state of the representative player but also the time index. In this way, we can define the terminal states of the game as all the states for which the time index is equal to the time horizon. 

\paragraph{Implementation details. }
From an implementation viewpoint, we stress the following points. We can use the \texttt{crowd\_modeling}\footnote{\url{https://github.com/deepmind/open_spiel/blob/master/open_spiel/python/mfg/games/crowd_modelling.py}} game as typical example.
\begin{itemize}
    \item The two main building blocks are:
    \begin{itemize}
        \item One class for the game, which inherits from \texttt{pyspiel.Game}. 
        \item One class for the state, which inherits from \texttt{pyspiel.State}
    \end{itemize}
    \item The state's evolution is implemented in the state's class.
    {\footnotesize
    $$
        \texttt{CHANCE} \xrightarrow[]{\texttt{\_apply\_action}}
        \texttt{MEAN\_FIELD} \xrightarrow[]{\texttt{update\_distribution}}
        \texttt{DEFAULT\_PLAYER\_ID} \xrightarrow[]{\texttt{\_apply\_action}}
        \texttt{CHANCE}  \dots
    $$
    }
    where we write over the arrow the method that is used for the update. These two methods are:
    \begin{itemize}
        \item \texttt{\_apply\_action}: takes an action as an input and uses it to update the state; notice that the representative player is not the only one taking actions as the chance player can similarly do so: the actions of the chance player represent the randomness from the environment that impacts the evolution of the representative player's state.  
        \item \texttt{update\_distribution}: updates the distribution using the one that is passed as an input (and which needs to be computed externally)
    \end{itemize}
    \item Other functions of the state class include:
    \begin{itemize}
        \item \texttt{chance\_outcomes}: returns the probabilities of all actions when at a chance node; this can be viewed as the (fixed) policy of the chance player; 
        \item \texttt{\_legal\_actions}: returns the actions that are admissible in the current state; this can be used in practice to forbid actions; some movements in a grid world can for example be forbidden due to the presence of obstacles; 
        \item \texttt{\_rewards}: computes the reward of the representative player in the current state.
    \end{itemize}
    
\end{itemize}

\subsection{Algorithms}

The algorithms for MFGs are located in open\_spiel/tree/master/open\_spiel/python/mfg/algorithms. As of February 2024, existing algorithms are as follows. First, there are algorithms with ``exact'' updates, i.e., without RL:
\begin{itemize}
    \item {\bf Fixed Point:} It implements basic fixed point iterations by computing a best response at each iteration and then computing the induced mean field. It is possible to use softmax policies for the best response computations.
    \item {\bf Fictitious Play:} Besides the standard fictitious play (with uniform average over past policies), the implementation allows to do damped fixed point iteration with a learning rate (which amounts to doing an average over past iterations with exponential smoothing). It is possible to use softmax policies for the best response computations. See e.g.~\citep{elie2020convergencemodelfreemfg,perrin2020fictitious} for more details. 
     \item {\bf Mirror Descent:} It implements Online Mirror Descent. See \citep{perolat2022scaling} for more details on this method. 
    An alternative is {\bf Munchausen Mirror Descent}, also implemented. 
    \item {\bf Boltzmann Policy Iteration:} It corresponds to policy iteration (for MFGs) with softmax policies. See \citep{cui2021approximately} for more explanations on this algorithm.
\end{itemize}
Second, several deep RL algorithms are also implemented in \texttt{OpenSpiel}:
\begin{itemize}
    \item {\bf Munchausen Deep Mirror Descent: } It is a combination of Munchausen OMD algorithm with deep RL.  See \citep{lauriere2022scalable} for more explanations on this algorithm and numerical examples.
    \item {\bf Average Network Fictitious Play: } It is a combination of Fictitious Play with deep RL to learn not only the best response at each iteration, but also to train a neural network learning the average policy. See \citep{lauriere2022scalable} for more explanations on this algorithm and numerical examples.
\end{itemize}

To implement algorithms, some important auxiliary files are:
\begin{itemize}
    \item \texttt{greedy\_policy.py}: allows to compute the greedy policy with respect to a Q-function.
    \item \texttt{distribution.py}: the class \texttt{DistributionPolicy} contains a tabular representation of a distribution associated to a given policy.
    \item \texttt{nash\_conv.py}: computes the exploitability of a policy.
    \item \texttt{policy\_value.py}: the class \texttt{PolicyValue} allows to compute the value of a policy.
\end{itemize}

\subsection{Environments (games)}

Environments in \texttt{OpenSpiel} are implemented either in C++ or in python, and are respectively available at the following adresses:
\begin{itemize}
    \item open\_spiel/tree/master/open\_spiel/games/mfg;
    \item open\_spiel/tree/master/open\_spiel/python/mfg/games.
\end{itemize}
Here is a brief description of the current implemented environments, as of February 2024.

\paragraph{Garnet: } 
A garnet stands for an abstract and randomly generated MDP, that can easily be turned into a single-population MFGs by modifying the reward via a modification of the reward structure. As detailed in \citep[Section 5.1]{perolat2022scaling}, a Garnet is characterized by the set of parameters $(n_x, n_a, n_b, s_f , \eta)$, where $n_x$ and $n_a$ stand respectively for the numbers of states and actions. The term $n_b$ is a branching factor characterizing the structure of the MDP. The transition kernel is built in the following way: $n_b$ transiting states are drawn randomly without replacement, and the associated transition probabilities are
obtained by partitioning the unit interval with $n_x-1$ uniformly sampled random points. The action driven reward $\tilde r(x, a)$ equals 0 for $s_f$ randomly sampled states, and set otherwise for all actions to a random value sampled uniformly in the unit interval. The population driven reward is simply the entropic one: $\bar r(s, \mu) = -\eta \log(\mu(x))$. Such term incentives agents to spread out and ensures the monotone property of the MFG, thus implying uniqueness of Nash. 

In \texttt{OpenSpiel}, a garnet is encoded within the \texttt{MFG\_GARNET} game type, while its set of parameters reads as $(n_x, n_a, n_b, s_f , \eta)=(\texttt{SIZE}, \texttt{NUM\_ACTION}, \texttt{NUM\_CHANCE\_ACTION}, \texttt{SPARSITY\_FACTOR, ETA})$ and its creation requires the additional specification of an horizon.

\paragraph{1D Crowd modelling:} The crowd modeling environment captures congestion effects in a population of agents moving on a 1 dimensional discrete torus, where each agent wishes to be close to a common point of interest, while avoiding overcrowded areas. It is presented in \citep[Section 4.2]{perrin2020fictitious} and described here for sake of completeness. The dynamics of a player follows
$$
 x_{n+1} = x_{n} + b(x_{n},a_{n})  + \epsilon_n\,,
$$
where $b(x_{n},a_{n})$ represents the movement of a player in position $x_n$ taking action $a_n$. Such movement is valued $-1$, $0$ or $1$ depending whether the action $a_n$ of the player implies a move to the left, no movement or a move to the right. The additional symmetric noise $\epsilon_n$ is valued $-1$, $0$ or $1$ with equal probability $1/3$. 

The number $|\mathcal{X}|$ of states is encoded in \texttt{\_SIZE} and set to 10 by default, while the time horizon $T$ is denoted  \texttt{\_HORIZON} with 10 default value. The instantaneous reward of a representative agent playing action $a_n$ in state $x_n$ while the population distribution is $\mu_n$ is given by
$$
 r(x_n,a_n,\mu_n) = |x_n-x^*| - \frac{|a_n|}{|\mathcal{X}|} - \log(\mu_n(x_n))\,,
$$
where the first term measures the proximity to the point of interest $x^*$, the second one captures the moving cost and the last one pictures the crowd aversion. The point of interest $x^*$ is the middle point of the $1$-dimensional state axis.

\paragraph{2D Crowd modelling:} This environment is a bi-dimensional version of the 1D Crowd modelling environment described above. Players move on a 2D torus and their actions are drawn within \{0:down,1: left, 2:neutral, 3:right, 4:up\}. The point of interest is placed at a fixed location in the torus, while an entropic term penalises being in a crowded place. Can also be specified the initial distribution (\texttt{INITIAL\_DISTRIBUTION}), the position of the reward (\texttt{POSITIONAL\_REWARD}), the crowd aversion coefficient (\texttt{CROWD\_AVERSION\_COEF}) or states that can not be visited (\texttt{FORBIDDEN\_STATES}). This model serves as a basis for the four-room exploration MFG and the maze MFG discussed in Section~\ref{sec:numerical-experiments}.

\paragraph{Linear Quadratic environment:} This environment encompasses a discrete MFG with linear dynamics and quadratic reward, whose continuous counterpart is commonly used in the MFG literature a it admits a closed form solution. It is described in Section 4.1 of  \cite{perrin2020fictitious} and reported here for sake of completeness. 

The state space $X$ is composed by points on a uniform symmetric $1$-dimensional grid $X = \{-L, . . . , L\}$ The finite action space is given by $A = \{-M, . . . , M\}$ allowing to move towards left or right directions. The dynamics of a typical player picking action $a_n$ at time $n$ are governed by the following equation:
$$
x_{n+1} = x_{n} + (K(m_n - x_n) + a_n)\Delta_n + \sigma \epsilon_n\;,
$$
where $\Delta_n$ represents the time step interval, $m_n$ is the mean field average position of the population and $\epsilon_n$ follows a discrete noise approximating a centered Gaussian noise with variance $\Delta_n$. The resulting state $x_{n+1}$ is rounded to the closest discrete state. At each time step, the player can move up to $M$ nodes and receives the running reward:
$$
r(x_n, a_n, m_n) = \left[-\frac{1}{2}|a_n|^2 + q a_n(m_n - x_n) - \frac{\kappa}{2}(m_n - x_n)^2\right]\Delta_n,
$$ 
where $q$ and $\kappa$ are given non-negative constants. The first term captures the moving cost and both remaining ones incentivise players to remain close to the average position of the population. At terminal date, the player receaives the quadratic reward $r(x_N , a_N , \mu_N ) = - \frac{c_{term}}{2}(m_N - x_N)^2$.

The parameters of the game are encoded with the following correspondence: $L$ as \texttt{SIZE}, $N$ as \texttt{HORIZON}, $M$ as \texttt{N\_ACTIONS\_PER\_SIDE}, $K$ as \texttt{MEAN\_REVERT}, $\sigma$ as \texttt{VOLATILITY}, $q$ as \texttt{CROSS\_Q}, $\kappa$ as \texttt{KAPPA}, $c_{term}$ as \texttt{TERMINAL\_COST} and $\Delta_n$ as \texttt{DELTA\_T}.

\paragraph{Predator-prey environment:} 
This multi-population game is an environment where several populations are chasing each other in a geometric room, that can either be a torus or a basic square. Its structure is inspired from the Hens-Foxes-Snakes outdoor game for kids, where the populations of hens, snakes and foxes are chasing each other in a cyclic manner. This game corresponds to the predator-prey game described in \citep[Section 5.4]{perolat2022scaling}.
The environment is configurable in the following high-level ways: number of populations (which correspond to ``players'' in \texttt{OpenSpiel}), reward matrix, 
geometry (torus, basic square).

\paragraph{Dynamic Routing game: } This is an environment modeling the evolution of drivers moving along roads organized in a graph structure. The state is the position on the edges of the graph, as well as the waiting time until moving to the one of the next edges. The goal for a representative player is to move from an initial node (or rather edge) to a terminal one. Players can have different origins and destinations. This model was proposed by \cite{cabannes2021solving}, who solved it using Online Mirror Descent using the version implemented in \texttt{OpenSpiel}. We refer to \citep{cabannes2021solving} and the \texttt{OpenSpiel} documentation for more details.

\end{document}